\documentclass[10pt,journal,compsoc]{IEEEtran}
\ifCLASSOPTIONcompsoc
  \usepackage[nocompress]{cite}
\else
  \usepackage{cite}
\fi
\ifCLASSINFOpdf
\else
\fi
\usepackage{amsmath}
\usepackage{amssymb}
\usepackage{ntheorem}
\usepackage{wrapfig,graphicx}
\usepackage{thmtools}
\usepackage{booktabs}       
\usepackage{multirow}
\usepackage{bm}
\usepackage{algorithm,algorithmic}
\usepackage{ulem}
\usepackage{enumitem}
\usepackage{ragged2e}  
\usepackage[colorlinks,anchorcolor=blue]{hyperref}
\newcommand{\MC}{\mathcal}
\newcommand{\MBB}{\mathbb}
\newcommand{\MBF}{\bm}
\newcommand{\eg}{\textit{e.g.}}
\newcommand{\ie}{\textit{i.e.}}
\newcommand{\etal}{\textit{et al.}}
\newcommand{\wrt}{\textit{w.r.t.}}

\newtheorem{theorem}{Theorem}
\newtheorem{lemma}[theorem]{Lemma}
\newtheorem{proposition}{Proposition}
\newtheorem{corollary}{Corollary}

\graphicspath{{figures/}}

\begin{document}
%
\title{Generalized Label Shift Correction via Minimum Uncertainty Principle: Theory and Algorithm}
%
%
%
%

\author{You-Wei Luo and Chuan-Xian Ren
\IEEEcompsocitemizethanks{\IEEEcompsocthanksitem Y.W. Luo and C.X. Ren are with School of Mathematics, Sun Yat-Sen University, Guangzhou, 510275, China. C.X. Ren is the corresponding author (rchuanx@mail.sysu.edu.cn).\IEEEcompsocthanksitem This work is supported in part by the National Natural Science Foundation of China under Grants 61976229, 12026601, and also supported in part by the Open Research Projects of Zhejiang Lab (No. 2021KH0AB08).}
}

%
%

\markboth{Journal of \LaTeX\ Class Files,~Vol.~14, No.~8, August~2015}%
{Shell \MakeLowercase{\textit{et al.}}: Bare Demo of IEEEtran.cls for Computer Society Journals}
%

\IEEEtitleabstractindextext{%
\begin{abstract}
   \justifying
   As a fundamental problem in machine learning, dataset shift induces a paradigm to learn and transfer knowledge under changing environment. Previous methods assume the changes are induced by covariate, which is less practical for complex real-world data. We consider the Generalized Label Shift (GLS), which provides an interpretable insight into the learning and transfer of desirable knowledge. Current GLS methods: 1) are not well-connected with the statistical learning theory; 2) usually assume the shifting conditional distributions will be matched with an implicit transformation, but its explicit modeling is unexplored. In this paper, we propose a conditional adaptation framework to deal with these challenges. From the perspective of learning theory, we prove that the generalization error of conditional adaptation is lower than previous covariate adaptation. Following the theoretical results, we propose the minimum uncertainty principle to learn conditional invariant transformation via discrepancy optimization. Specifically, we propose the \textit{conditional metric operator} on Hilbert space to characterize the distinctness of conditional distributions. For finite observations, we prove that the empirical estimation is always well-defined and will converge to underlying truth as sample size increases. The results of extensive experiments demonstrate that the proposed model achieves competitive performance under different GLS scenarios.
\end{abstract}

}

\maketitle

\IEEEdisplaynontitleabstractindextext

\IEEEpeerreviewmaketitle

\IEEEraisesectionheading{\section{Introduction}\label{sec:introduction}}

\begin{figure}
  \begin{center}
      \includegraphics[width=0.485\textwidth,trim=18 20 20 15,clip]{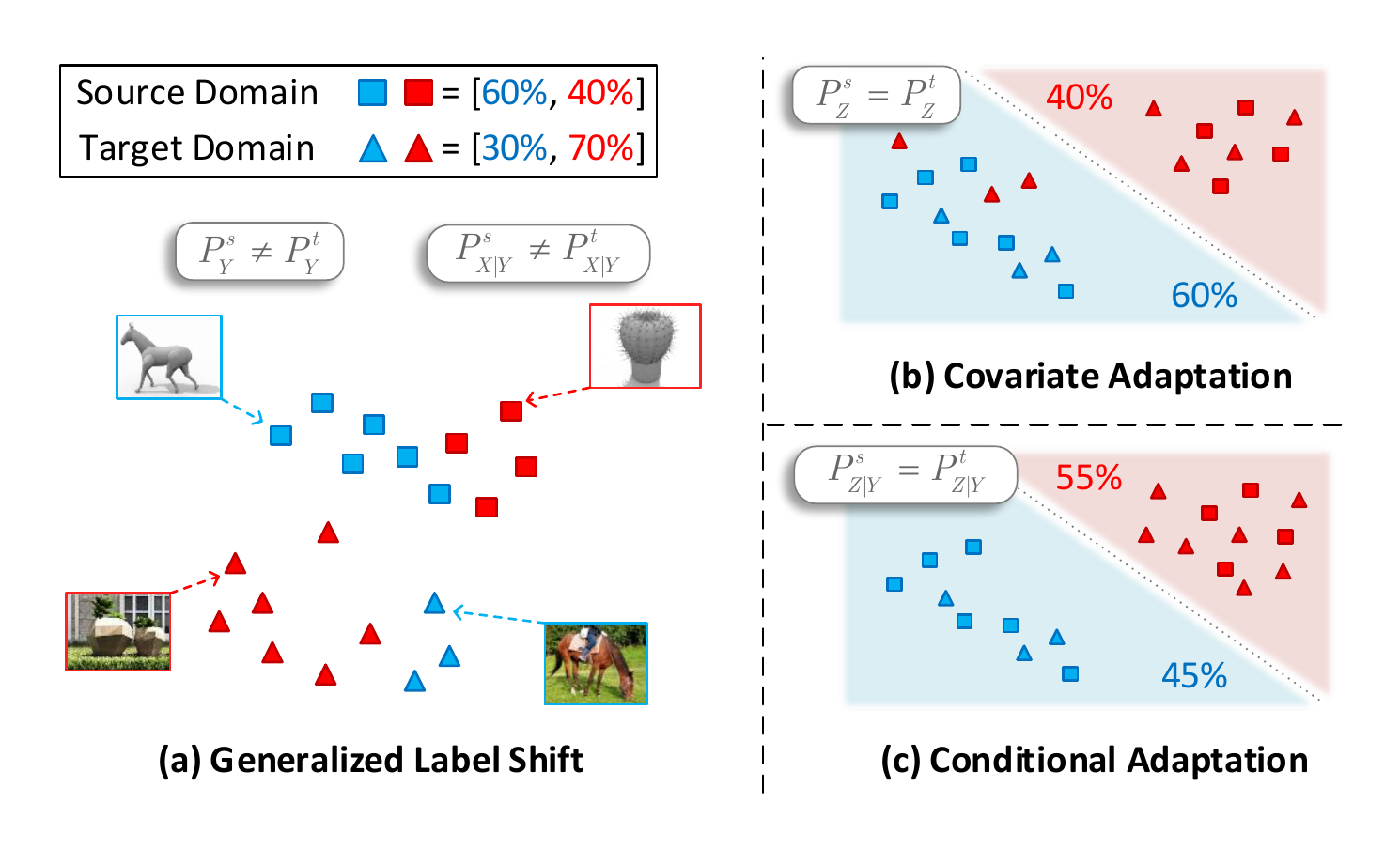}
      \caption{(a) Illustration of generalized label shift, where both label distribution $P_Y$ and conditional distribution $P_{X|Y}$ change across domains. (b) covariate adaptation matches the marginal distribution $P_Z$ via transformation $G(\cdot): X\mapsto Z$, while distorting the conditional distributions $P_{Z|Y}$, \eg, the target samples ``\textcolor{red}{$\blacktriangle$}'' (negative transfer). (c) conditional adaptation matches the cluster structures correctly. Best viewed in color.}
      \label{fig:conditional_shift_problem}
  \end{center}
  \vskip -0.2in
\end{figure}

\IEEEPARstart{A}{s} a fundamental problem in machine learning, dataset shift induces a paradigm to study the essential connection between changing environment and adaptive model \cite{quionero2009dataset}. Since real-world data are usually correlated with changing environment, the learned model will be biased and experience degraded performance in the new environment. Mathematically, dataset shift is a situation that the joint distribution of input (covariate) and output (label) changes across the domains (environments). The basic problems in dataset shift are to detect and correct the shifting distributions under different causality assumptions \cite{scholkopf2012causal}. Due to its theoretical background, dataset shift is closely related to another vital problem called Unsupervised Domain Adaptation (UDA), which aims to deal with the shortage of prior knowledge (\eg, labels) in the new environment. The primary goal of UDA is to leverage the task-related knowledge in the source domain with sufficient labeled data to help the target domain without labels, while removing undesirable information in knowledge transfer. As shown in Figure \ref{fig:conditional_shift_problem} (a), the source and target domains follow different distributions under dataset shift. Based on the dataset shift theory, considerable efforts have been made to understand the adaptation mechanism \cite{zhang2013domain,combes2020domain}, and explore knowledge transfer models \cite{ganin2015unsupervised,courty2016optimal,long2018transferable,zhang2020optimal,luo2020unsupervised,li2020deep,tang2021towards}. These advancements are generally applicable in many real-world scenarios: pedestrian re-identification \cite{ren2020domain}, autonomous driving \cite{wulfmeier2017addressing,tang2021towards}, medical image \cite{mahmood2018unsupervised}, cross-modal learning \cite{xuan2020cross,jing2020cross}.

Modeling UDA problem requires studying the particular attributions of  shift. Let $X$, $Y$ denote the covariate and label variable, respectively. For simplicity, let $P$ be the distribution where the subscript and superscript on it represent the corresponding variable and domain, respectively. From the perspective of dataset shift, there are three common assumptions for characterizing the cause of shifting distributions.

\textbf{(\uppercase\expandafter{\romannumeral1})} Covariate shift decomposes the joint distribution as $P_{XY}=P_{Y|X}P_X$. It assumes that the distribution of covariate $P_X$ changes across domains while the labeling rule $P_{Y|X}$ is invariant. A basic theory for UDA was developed by Ben-David {\etal} \cite{ben2010theory}, which provided an informative generalization error bound on the target domain. By defining a distance on domains called $\MC{H}\Delta\MC{H}$-divergence, they proved that small divergence is crucial for successful UDA. Following this theory, covariate adaptation models learn a transformation $Z=G(X)$ such that the discrepancy between marginal distributions $P_Z$ are minimized as Figure \ref{fig:conditional_shift_problem} (b). Covariate adaptation methods can be roughly concluded as moment/distribution matching \cite{hu2015deep,long2018transferable}, adversarial confusion \cite{ganin2015unsupervised,long2018conditional}, generative model \cite{ren2020domain}, manifold learning \cite{luo2020unsupervised,ren2019heterogeneous}, optimal transport \cite{courty2016optimal,li2020Enhanced,zhang2020optimal,luo2021conditional} and so on.

\textbf{(\uppercase\expandafter{\romannumeral2})} Label shift \cite{zhang2013domain} decomposes the joint distribution as $P_{XY}=P_{X|Y}P_Y$. It assumes the label distributions $P_Y$ of the domains are different while the conditional distributions $P_{X|Y}$ are identical. The label shift problem is partially described in Figure \ref{fig:conditional_shift_problem} (a), where the source and target domains have different class proportions. Zhao {\etal} \cite{zhao2019learning} discussed the limitation of covariate adaptation, and improved the generalization error bound of Ben-David {\etal} by introducing the discrepancy between labeling rules. Specifically, under the label shift, matching the marginal distributions $P_Z$ is error-prone as the conditional distributions $P_{Z|Y}$ will be distorted according to the law of total probability \cite{yan2017mind}. Unfortunately, the misaligned $P_{Z|Y}$ will increase the uncertainty and error of the transferred classifier on the target domain, since the cluster structures are distorted as shown in Figure \ref{fig:conditional_shift_problem} (b). A common solution to the label shift is learn the adaptation model under importance weighted Empirical Risk Minimization (ERM) framework \cite{zhang2013domain,lipton2018detecting,combes2020domain}.

\textbf{(\uppercase\expandafter{\romannumeral3})} By relaxing the identical constraint of $P_{X|Y}$, Generalized Label Shift (GLS) \cite{combes2020domain,ren2018generalized,zhang2013domain} is explored as Figure \ref{fig:conditional_shift_problem} (a). An extreme scenario of GLS is the so-called Partial Domain Adaptation (PDA) problem \cite{cao2018partial,zhang2018importance,cao2019learning,li2020deep}, where the label space of the target domain is a subset of the source label space, \ie, the prior probabilities of the target domain equal to zero for some classes. The primary goal of GLS is to alleviate the negative transfer by matching the conditional distributions as Figure \ref{fig:conditional_shift_problem} (c). In fact, the GLS precondition is more practical for recognition problem, since the class proportions (\ie, label distributions) are not necessarily the same for different domains \cite{combes2020domain}. Meanwhile, $P_{Z|Y}$ is closely related to cluster structures and discriminability on the target domain. However, current GLS methods mainly focus on the shift of prior distribution while assuming that the conditional shift will be automatically mitigated during the learning of domain invariant representations. Such an assumption is implicit and intractable, so there is no guarantee for conditional shift.

Currently, there are two major challenges in GLS. First, GLS is not well-connected with the statistical learning theory. So there is a lack of theoretical justification as to how GLS correction can help knowledge transfer. Second, explicit modeling of conditional invariant transformation $G(\cdot)$ is unexplored, and the discriminability of $G(\cdot)$ is not sufficiently learned with empirical/structure risk minimization.

In this work, we deal with the challenges by proposing the Minimum Uncertainty Learning (MUL) principle, which will be guaranteed by statistical learning theory. We first define the \textit{transfer uncertainty} and \textit{decision uncertainty} in transfer process. As shown in Figure \ref{fig:uncertainty}, conditional invariant learning is reformulated as the minimization of cross-domain conditional discrepancy, \ie, \textit{transfer uncertainty}. The discriminability is interpreted as the overlapping region of the conditional distributions of different classes, and further quantified by the \textit{decision uncertainty}. With these two uncertainty terms, we derive a tighter generalization error bound for GLS. The empirical model of MUL is built on the Reproducing Kernel Hilbert Space (RKHS) to characterize the conditional distribution discrepancy. Generally, MUL can deal with the GLS, UDA and PDA problems. Competitive experiment results are achieved by MUL model. Our contributions can be summarized as follows.
\begin{itemize}
  \item We connect the statistical learning theory with GLS by introducing the uncertainty terms in knowledge transfer. Following the principle of MUL, we theoretically prove that the generalization error of MUL is smaller than previous covariate adaptation.
  \item We propose the MUL model for GLS, which unifies the two mainstream schemes in current domain adaptation, \ie, discriminability and transferability. MUL model can be generally applied to different GLS scenarios, \eg, UDA and PDA.
  \item To quantify the conditional discrepancy in MUL efficiently, we propose the \textit{conditional metric operator} based on the conditional mean embedding theory in RKHS. A computable formulation for MUL is developed by deriving its empirical estimation.
  \item For empirical MUL, theoretical results including identifiability, statistical consistency and fast computation formula are proved. These properties ensure MUL is well-defined in application. Extensive comparison and analysis are conducted under various GLS settings, where MUL achieves superior results.
\end{itemize}

\section{Related Works}\label{sec:related_work}
In this section, we first review the covariate adaptation and conditional adaptation models for UDA in Section \ref{subsec:dataset_shift}. Then we review the mainstream schemes in current domain adaptation (\ie, discriminability and transferability) in Section \ref{subsec:transfer_discriminant}. In the following, $Y$ is the discrete label variable with $c$ classes which takes its value from $\MC{Y}=\{y_1,y_2,\ldots,y_c\}$.

\subsection{Dataset Shift Correction}\label{subsec:dataset_shift}
\textbf{Covariate Adaptation. }
A basic guarantee for successful UDA was proved by Ben-David \etal \cite{ben2010theory,david2010impossibility}, which introduced the $\MC{H}\Delta\MC{H}$-divergence and joint optimal risk $\lambda^*$ to characterize the generalization error on the target domain. Most covariate adaptation methods try to learn the domain invariant representations with $P^s_{Z}=P^t_{Z}$, which essentially rely on the fact that aligned marginal distributions are sufficient for small $\MC{H}\Delta\MC{H}$-divergence. To align the marginal distributions, various models are proposed based on different statistical distance. Deep Adaptation Network (DAN) \cite{long2018transferable} employed the well-known two-sample test called Maximum Mean Discrepancy (MMD) \cite{gretton2012kernel} to learn domain invariant representations. Domain Adversarial Neural Network (DANN) \cite{ganin2015unsupervised} was the first method that introduces the adversarial two-player game to UDA. It trained the feature generator and domain discriminator alternatively, which minimized the Jensen-Shannon divergence. Optimal Transport (OT) \cite{courty2016optimal} sought the optimal plan with minimum transport cost across domains, which equals to the Wasserstein distance. Zhang \etal \cite{zhang2020optimal} extended OT to RKHS, and provided a closed-form solution for the transport problem in RKHS. Though covariate adaptation enhances the transferability, the discriminability {\wrt} the conditional distributions $P_{Z|Y}$ will be degraded \cite{zhao2019learning}.

\noindent
\textbf{Conditional Adaptation. }
The conditional adaptation for UDA was first proposed by Zhang \etal \cite{zhang2013domain} based on the GLS precondition. Following them, Gong \etal \cite{gong2016domain} tried to correct GLS by aligning the constructed marginal distributions. Combes \etal \cite{combes2020domain} applied the importance weights to covariate adaptation methods, which successfully mitigated the negative transfer problem caused by shifting $P_Y$. However, these works mainly focused on $P_Y$ while making some intractable assumptions on the conditional distributions or cluster structures. Other methods tried to improve covariate adaptation by exploiting the label information. Conditional Domain Adversarial Network (CDAN) \cite{long2018conditional} proposed the adversarial learning over the multilinear map between variables $Z$ and $Y$. Adversarial Tight Match (ATM) \cite{li2020maximum} minimized the intra-class density to form the compact class cluster. Discriminative Manifold Propagation (DMP) \cite{luo2020unsupervised} built a manifold learning framework to align the local manifold structures across domains, which connected the alignment error with manifold theory. Zhang \etal \cite{zhang2021deep} extended the manifold assumption of DMP by considering the spherical manifold with Gaussian kernel. Xia \etal \cite{xia2020structure} aligned the local structure by exploiting the Gromov-Wasserstein distance on graphs. The methods above either ignore the label shift or change the prior $P_Y$ when aligning the joint distributions. In our work, we present a conditional adaptation framework, which learns conditional invariant transformation and corrects label shift simultaneously.

\vskip -0.4in
\subsection{Transferability and Discriminability}\label{subsec:transfer_discriminant}

\noindent
\textbf{Transferability. }
For task-related knowledge transfer, it is natural to consider the discrepancy between the conditional distributions $P^s_{Z|Y}$ and $P^t_{Z|Y}$. Different from the marginal discrepancy, the conditional discrepancy characterizes the class-specific transferability and negative transfer problem. Thus, the first problem in current UDA is the matching of knowledge at class-level. As the labels on the target domain are unknown, some methods tried to match the domains at class-level with hard-assigned pseudo labels \cite{ding2018graph,li2020maximum,luo2021conditional} or integrate the label information into covariate adaptation via soft predictions \cite{long2018conditional,tang2021towards,li2020Enhanced,jing2020cross}. However, the conditional distributions $P_{Z|Y}$ are still not modeled explicitly. Some recent progresses \cite{zhang2013domain,ren2018generalized,combes2020domain} tried to reduce the conditional discrepancy under the GLS precondition. These works assumed the transformed distributions $P_{Z|Y}$ are identical across domains. Then the label shift can be tackled via importance sampling strategy. Specifically, the transformation $G(\cdot)$ was parameterized linearly in literatures \cite{zhang2013domain,ren2018generalized}, \ie, $Z = \MBF{W} \odot X + \MBF{b}$ where the parameters $(\MBF{W},\MBF{b})$ depend on $Y$. Combes \etal \cite{combes2020domain} parameterized $G(\cdot)$ by the Deep Neural Networks (DNNs). Though the cluster structure condition in \cite{combes2020domain} is somehow weaker than the identical assumption of $P_{Z|Y}$, it is still intractable. Thus, the existence of $G(\cdot)$ and its explicit modeling remain open.

\begin{figure}
  \begin{center}
      \includegraphics[width=0.46\textwidth,trim=16 16 21 18,clip]{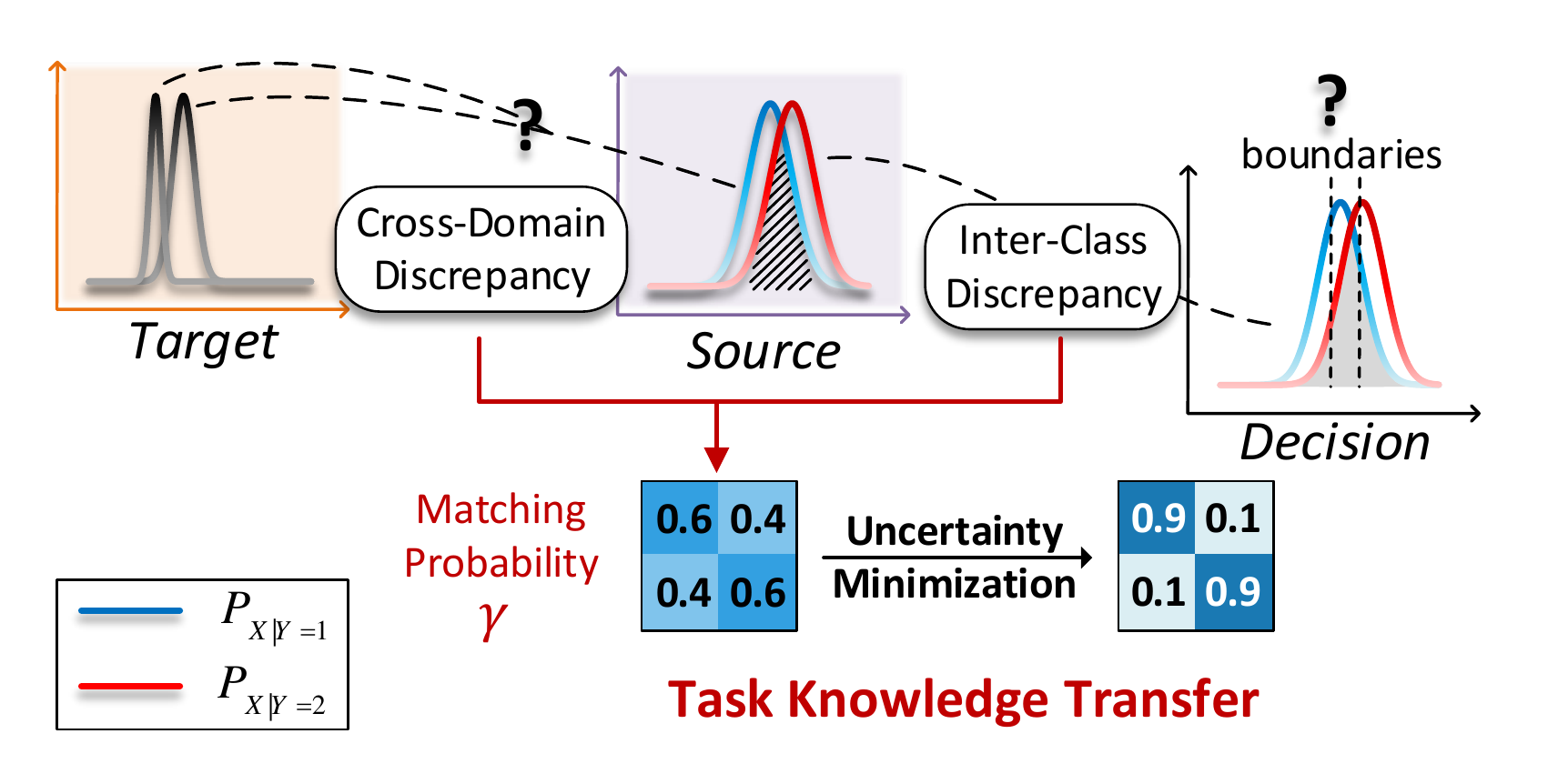}
      \caption{Illustration of the uncertainty. The transfer uncertainty appears in the matching of task-related knowledge across domains, \ie, the matching of $P^s_{X|y_i}$ and $P^t_{X|y_i}$, which is characterized by the cross-domain discrepancy. The decision uncertainty appears in the learning of task-related knowledge, \ie, the identification of $P_{X|y_i}$ and $P_{X|y_j}$, which is formulated as the inter-class discrepancy. It can be visualized as the gray overlapping region of distributions. Best viewed in color.}
      \label{fig:uncertainty}
  \end{center}
  \vskip -0.2in
\end{figure}

\noindent
\textbf{Discriminability. }
Chen \etal \cite{chen2019transferability} showed that existing transfer methods, which only focus on the transferability, may degrade the discriminability. Therefore, in addition to studying the conditional adaptation for transferability, it is also important to consider the structure of $P_{Z|Y}$ which helps to enhance the discriminability. Specifically, the overlapping conditional distributions form a gray uncertain region which serves as the lower bound for error of any predictor. Some models \cite{zhang2013domain,gong2016domain} considered GLS from the perspective of causal relation $Y \rightarrow X$. The key assumption is that the transformed conditional distributions $P_{Z|y_i}$ are linearly independent, which implies the transformed representations are supposed to be discriminative enough. Adaptive Feature Norm (AFN) \cite{xu2019larger} and Batch Spectral Penalization (BSP) \cite{chen2019transferability} studied the discriminability and transferability trade-off via the spectrum of representations. They reached a conclusion that the representations with larger spectral radius are more transferable but less discriminative. Enhance Transport Distance (ETD) \cite{li2020Enhanced} improved the discriminability of OT by weighting the transport distances with label information. Recent advancements \cite{ding2018graph,chen2019transferability,li2020Enhanced,luo2020unsupervised} enhanced the discriminability but ignored conditional shift (transferability).

\section{Theoretical Insight}\label{sec:theoretical_insight}
In this section, we clarify the theoretical motivation of MUL. We first provide some intuitions for uncertainty in knowledge transfer, and mathematically characterize the uncertainty via discrepancy. Then we derive an informative generalization bound and connect it with uncertainty. Finally, we define the principle of minimum uncertainty learning.

\subsection{A Unified View via Uncertainty}
To introduce a unified view of transferability and discriminability, we define the uncertainty in knowledge transfer. Generally, we consider the matching problem between cross-domain conditional distributions as
\begin{equation*}
\gamma^{st} = \mathop{\arg\min}_{\gamma \in \Omega} \sum_{i,j=1}^c \gamma_{ij} D(P^s_{Z|y_i} \| P^t_{Z|y_j}),
\end{equation*}
where $D(\cdot \| \cdot)$ is statistical distance and $\Omega$ is the set of doubly stochastic matrices. Similarly, let $\gamma^{ss}$ be the matching probability between $P^s_{Z|y_i}$ and $P^s_{Z|y_j}$. Intuitively, $\gamma^{st}$ and $\gamma^{ss}$ are supposed to be diagonal matrices, \ie, one-to-one matching, for mitigating the negative transfer and misclassification. It implies both transferability and discriminability can be characterized by the uncertainty shown as Figure \ref{fig:uncertainty}.

The matching problem provide a basic motivation for successful adaptation, \ie, minimum uncertainty in decision and transfer. Based on it, we present following definitions.
\begin{itemize}[leftmargin=*]
\item \textbf{Decision uncertainty} means uncertainty in $\gamma^{ss}$ and $\gamma^{tt}$. A low uncertainty ensures a small risk in decision-making (\ie, the overlapping region in Figure \ref{fig:uncertainty}).
\item \textbf{Transfer uncertainty} means uncertainty in $\gamma^{st}$. A low uncertainty ensures a lower possibility of misalignment and the effectiveness of decision on the target domain.
\end{itemize}
Note that minimizing decision and transfer uncertainties in matching problem is equivalent to maximize the inter-class discrepancy $D(P^s_{Z|y_i} \| P^s_{Z|y_j})$ and minimize the cross-domain discrepancy $D(P^s_{Z|y_i} \| P^t_{Z|y_i})$, respectively. Based on the conditional discrepancy, we will connect the intuitive uncertainty minimization with learning theory in the next.

\subsection{Learning Theory for Domain Adaptation}
A basic theory for UDA was established by Ben-David \etal \cite{ben2010theory} under binary classification scenario. For any hypothesis $h(\cdot):~X \mapsto Y$ in hypothesis class $\MC{H}$, its expected generalization error $\varepsilon(h)$ on the target domain is bounded as
\begin{equation}\label{eq:Ben-David_bound}
   \varepsilon_t(h) \leq  \varepsilon_s(h) +  d_{\MC{H}\Delta \MC{H}}(\MC{D}_s,\MC{D}_t)/2 + \lambda^*,
\end{equation}
where $\lambda^* = \min_{h\in \MC{H}} [\varepsilon_s(h)+\varepsilon_t(h)]$ is the optimal joint risk on both the source and target domains. As $\lambda^*$ serves as a constant \wrt~$h$ in Eq.~\eqref{eq:Ben-David_bound}, it is intractable to evaluate. Many adaptation methods usually ignore this term, which leads to a loose and inexact estimation of generalization error. To mitigate this problem, Zhao \etal \cite{zhao2019learning} provided a more general bound which is free of the intractable $\lambda^*$:
\begin{equation*}\label{eq:Zhao_bound}
   \varepsilon_t(h) \leq ~  \varepsilon_s(h) + d_{\MC{H}\Delta \MC{H}}(\MC{D}_s,\MC{D}_t)/2 + \beta^*,
\end{equation*}
where $\beta^*=\min\{\MBB{E}_{X^s}[|f_s - f_t |],\MBB{E}_{X^t}[|f_s - f_t |]\}$, $f(x)= p_{y_2|x}$ is the labeling rule and $p$ the probability density/mass function (depending on the variables).

Though the generalization upper bound above is generally tighter \cite{zhao2019learning}, the discrepancy of labeling rules in $\beta^*$ is still hard to evaluate and cannot connect to GLS. To better understand $\beta^*$, we build a probabilistic adaptation model on conditional distribution $P_{Z|Y}$. Consequently, we connect the generalization error $\varepsilon_t$ with GLS informatively, and deduce a principle for successful knowledge transfer.
\begin{theorem}\label{thm:our_bound}
   Denote the adjusted source distribution as
      $P^{\tilde{s}}_{X} = \sum_{i=1}^c P^s_{X|y_i} p^t_{y_i}$.
   Then $\beta^*$ is bounded as
   \begin{equation*}\label{eq:our_bound}
      \begin{aligned}
         &~\min\{\MBB{E}_{X^s}[|f_s - f_t |],\MBB{E}_{X^t}[|f_s - f_t |]\}\\
         \leq &~ \underbrace{ \left[ \eta D_1(P^s_{X|y_2} \| P^t_{X|y_2}) + D_2(P^{\tilde{s}}_{X} \| P^t_X) \right]}_{\emph{cross-domain discrepancy}} + \delta D_3(P^s_Y\| P^t_Y),
      \end{aligned}
   \end{equation*}
   where $\eta=\min\{p^s_{y_2},p^t_{y_2}\}$, $\delta$ depends on the conditional distributions $P_{X|Y}$, and $D_{1}, D_{2}, D_{3}$ are statistical distances.
\end{theorem}

Theorem \ref{thm:our_bound} suggests that $\beta^*$ is bounded by the transfer uncertainty, and the conditional matching is a necessary condition for successful adaptation. Now we consider the generalization error under the conditional invariant transformation $Z=G(X)$, where hypothesis $h$ and labeling rule $f$ are functions of $Z$. Denote the term $\beta^*$ of covariate adaptation model with $P^s_{Z}=P^t_{Z}$ as $\beta^*_{\text{cov}}$; for conditional invariant model with $P^s_{Z|Y}=P^t_{Z|Y}$, denote as $\beta^*_{\text{con}}$. We now introduce the relation between $\beta^*_{\text{cov}}$ and $\beta^*_{\text{con}}$.
\begin{corollary}\label{cor:beta_con_cov_comparison}
   If $P^s_{Z|Y}=P^t_{Z|Y}$, then $\delta = \delta(P^s_{Z|Y},P^t_{Z|Y}) < 1$ and the cross-domain conditional discrepancy will be mitigated:
   \begin{equation*}\label{eq:transfer_uncertainty_zero}
      D_1(P^s_{Z|y_2} \| P^t_{Z|y_2}) = D_2(P^{\tilde{s}}_{Z} \| P^t_Z) = 0.
   \end{equation*}
   Further, the following inequality holds:
   \begin{small}
   \begin{equation*}\label{eq:beta_con_cov_comparison}
      \beta^*_{\text{con}} < D_3(P^s_Y\| P^t_Y) \leq \beta^*_{\text{cov}} \leq D_1(P^s_{Z|y_2} \| P^t_{Z|y_2}) + D_3(P^s_Y\| P^t_Y).
   \end{equation*}
   \end{small}
\end{corollary}
Corollary \ref{cor:beta_con_cov_comparison} shows that $\beta^*_{\text{con}}$ is strictly smaller than the lower bound of $\beta^*_{\text{cov}}$. It implies that the generalization error of conditional invariant model is always smaller than covariate adaptation model, since the lower bound of $\beta^*_{\text{cov}}$ is usually hard to achieve. Besides, the conditional distributions are distorted by covariate adaptation (see proof), which implies $D_1(P^s_{Z|y_2} \| P^t_{Z|y_2})>0$ and negative transfer. Thus, the upper bound of $\beta^*_{\text{cov}}$ is significantly larger than $\beta^*_{\text{con}}$.

For decision uncertainty, note that the classification error is a function of both hypothesis $h$ and transformation $G$ now. It means that we can consider the hypothesis $h$ on the transformed variable $Z$ as $Y=h(Z)$, and then optimize the lower bound of original error $\varepsilon_s(h)$:
\begin{equation*}
   \varepsilon_s(h,G) \triangleq \min_G~\varepsilon^G_s(h).
\end{equation*}
As shown in Figure \ref{fig:uncertainty}, the key to minimize the optimal classification error (\ie, lower bound) is to learn a transformation $G$ with lower decision uncertainty. This problem is mathematically equivalent to maximize $D(P^s_{Z|y_i} \| P^s_{Z|y_j})$.

\begin{theorem}\label{thm:source_error_and_task_uncertainty}
   If $G^*$ is the solver of decision uncertainty problem
   \begin{equation*}
      \mathop{\arg\max}_G \sum_{i\neq j} D(P_{Z|y_i} \| P_{Z|y_j}),
   \end{equation*}
   then $\varepsilon^{G^*}(h^*)=\underset{G}{\min}~ \varepsilon^G(h^*)$, where $h^*$ is the Bayes classifier.
\end{theorem}

Theorem \ref{thm:source_error_and_task_uncertainty} implies that minimization of the \textit{decision uncertainty}, \ie, maximization of conditional discrepancies, is sufficient to minimize the transformed source error $\varepsilon^G_s(h^*)$. Note that $\varepsilon^G_s(h^*)$ is generally smaller than the source error $\varepsilon_s(h^*)$ in original covariate space.

In conclusion, the theoretical results above imply that the uncertainty minimization via discrepancy optimization is sufficient for successful knowledge transfer under the more challenging GLS scenario. Besides, combining these results, it is straightforward to obtain a tighter generalization bound under the optimal transformation $G^*$:
\begin{equation}\label{eq:transformed_upper_bound}
   \varepsilon^{G^*}_t(h) \leq ~  \varepsilon^{G^*}_s(h) + d_{\MC{H}\Delta \MC{H}}(\MC{D}_s,\MC{D}_t)/2 + D_3(P^s_Y\| P^t_Y).
\end{equation}
Usually, the hypothesis $h$ is learned from finite samples with a certain loss function. Then $h$ can be arbitrarily close to the Bayes classifier $h^*$ with sufficient data, and $\varepsilon^{G^*}_s(h)$ is reasonably small. Finally, the transformation $G^*$ will achieve knowledge transfer successfully.

\subsection{Principle of Minimum Uncertainty Learning}

Above theoretical results suggest that the conditional invariant transformation $Z=G(X)$ with minimum uncertainty is sufficient to achieve a smaller generalization error for successful adaptation. Generally, we formulate the principles of minimizing \textbf{t}ransfer \textbf{u}ncertainty and \textbf{d}ecision \textbf{u}ncertainty as the following discrepancy optimization problem:
\begin{align*}
   &\min_{G}~ \MC{J}_{\text{TU}} (G) =  \sum_{i} p^t_{y_i} D(P^s_{Z|y_i} \| P^t_{Z|y_i}), \\
   &\max_{G}~ \MC{J}_{\text{DU}} (G) =  \sum_{i\neq j} D(P^s_{Z|y_i} \| P^s_{Z|y_j}).
\end{align*}
The criterion $\MC{J}_{\text{TU}}$ follows the cross-domain discrepancy in Theorem \ref{thm:our_bound} and $\MC{J}_{\text{DU}}$ follows the results in Theorem \ref{thm:source_error_and_task_uncertainty}, where the weight $p^t_{y_i}$ is deduced from the adjusted matching, \ie, $|P^{\tilde{s}}_{X} - P^t_{X}|=\sum_{i} p^t_{y_i} |P^s_{X|y_i}-P^t_{X|y_i}|$. Note that statistical moment-based methods are usually built on the Euclidean space, so they are insufficient to achieve discrepancy optimization. To overcome this limitation, we will build the model in RKHS with the proposed metric operator. Consequently, our model not only explicitly optimizes the discrepancy-based criteria shown above, but also shows some appealing properties in empirical estimation.

Besides the conditional shift, the correction of label shift is also crucial for successful domain adaptation \cite{yan2017mind,combes2020domain}. Under label shift, the ERM on the source domain is biased \cite{zhang2013domain}. Note that the prior discrepancy $D_3(P^s_Y\| P^t_Y)$ in error bound also implies the label shift. Usually, the shifting $P_Y$ can be corrected by importance weighted ERM \cite{zhang2013domain,gong2016domain,lipton2018detecting,combes2020domain}. Let $l(\cdot)$ be the loss function (\eg, cross-entropy and MSE), the importance weighted ERM is motivated by
\begin{align}
   \varepsilon_t &= \MBB{E}_{(X^t,Y^t)}[l(x,y;h)] \nonumber \\
   &= \iint l(x,y;h) \frac{p^t_{x|y} p^t_{y}}{p^s_{x|y} p^s_{y}} p^s_{xy} ~{\mathrm{d}x} {\mathrm{d}y} \nonumber \\
   & = \MBB{E}_{(X^s,Y^s)}[w(x,y)l(x,y;h)], \nonumber
\end{align}
where $w(x,y)=\frac{p^t_{x|y} p^t_{y}}{p^s_{x|y} p^s_{y}}$ is the so-called importance weight. Now the target risk can be equivalently reformulated with the weighting source risk $\varepsilon^G_{\tilde{s}}\triangleq\MBB{E}_{(X^s,Y^s)}[w(x,y)l(x,y;h)]$. Label shift models always rely on the identical conditional distribution assumption \cite{lipton2018detecting,combes2020domain}, \ie, $P^s_{X|Y}=P^t_{X|Y}$, which no longer holds under GLS. For MUL, this assumption reasonably holds for the transformed distributions $P_{Z|Y}$ with minimum transfer uncertainty. Once the assumption is satisfied, the importance weight and weighting empirical risk can be simplified as $w(y) = p^t_{y}/ p^s_{y}$ and $\varepsilon_{\tilde{s}} \triangleq \MBB{E}_{(X^s,Y^s)}[w(y)l(x,y;h)]$, respectively. With above analysis in mind, we formulate the principle of conditional invariant transformation $G(\cdot)$ for GLS:
\begin{equation}\label{eq:MAL_framework}
   \min_{G} ~ \varepsilon^G_{\tilde{s}} + \MC{J}_{\text{TU}}(G) - \MC{J}_{\text{DU}}(G).
\end{equation}

Note that $\MC{J}_{\text{TU}}$ and $\MC{J}_{\text{DU}}$ require to measure the statistical distance between conditional distributions. Classical divergences (\eg, $f$-divergence) always require to estimate the probability distributions $P_{Z|y}$, which is difficult and not straightforward for the continuous variables. Covariate adaptation methods usually use MMD \cite{gretton2012kernel} to measure the discrepancy between the continuous marginal distributions. The class-wise estimation manner is a simple extension for measuring the conditional discrepancy, which splits the dataset $\MC{D}$ into multiple subsets $\MC{D}_{y_i}$ according to the label variable $Y$. However, this strategy makes the sample size small for each MMD estimation, and will yield $|\mathcal{Y}|\times$ larger estimation error where $|\mathcal{Y}|$ is the number of classes. Besides, the class-wise computation is inefficient for large $|\mathcal{Y}|$ and inapplicable when $Y$ is continuous. In the next section, we will propose the \textit{conditional metric operator} based on the conditional embedding theory \cite{song2009hilbert} in RKHS, which characterizes the conditional discrepancy with whole dataset.

\begin{figure*}
   \begin{center}
       \includegraphics[width=0.98\textwidth,trim=20 21 18 18,clip]{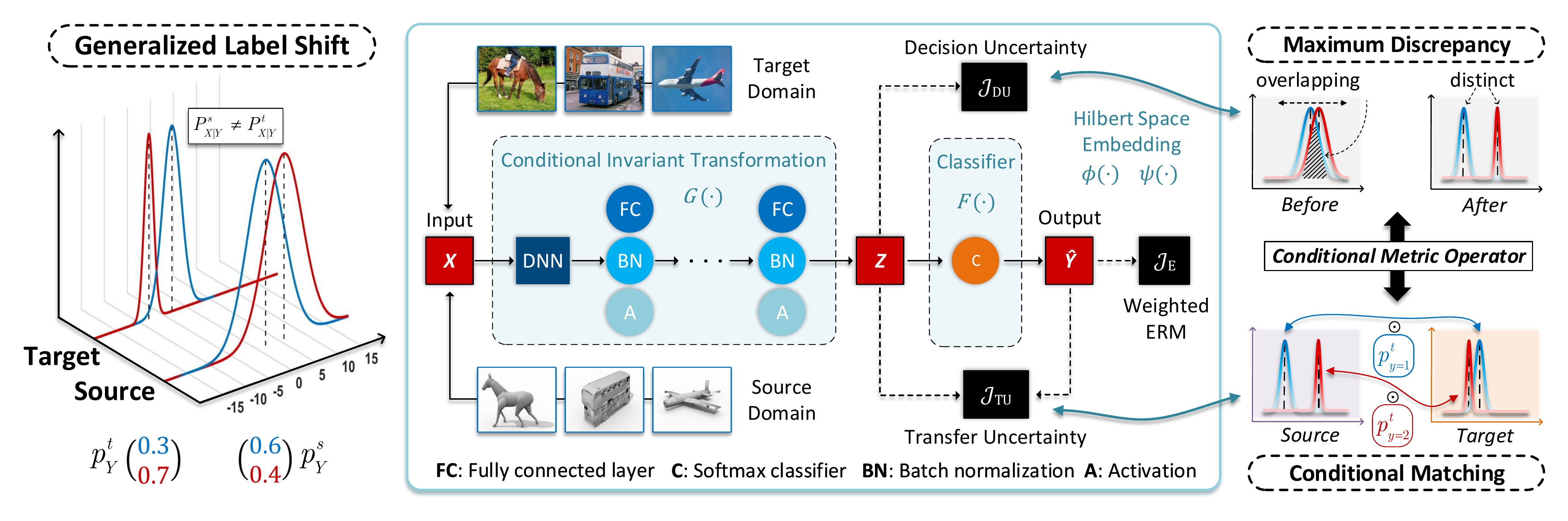}
       \caption{Flowchart of MUL for GLS. A simple GLS is shown in the left, where different colors represent different conditional distributions $P_{X|Y=y}$ and $p_Y$ is the probability mass (class proportions). In the middle, MUL consists of a transformation $G(\cdot)$ and a classifier $F(\cdot)$. To learn and transfer the task-related knowledge under GLS, $F(\cdot)$ and $G(\cdot)$ are incentivized to be unbiased on the target domain and conditional invariant across domains, respectively. As shown in the right, MUL characterizes the conditional distributions via the Hilbert space embeddings. The ``conditional matching'' for transfer uncertainty and ``maximum discrepancy'' for decision uncertainty are achieved by $\MC{J}_{\text{TU}}$ and $\MC{J}_{\text{DU}}$, respectively. Best viewed in color.}
       \label{fig:flowchart_MAL}
   \end{center}
   \vskip -0.2in
\end{figure*}

\section{MUL via Discrepancy Optimization}\label{sec:minimum_uncertainty_learning}
Now we introduce MUL for GLS correction via discrepancy optimization. We first briefly review the fundamental theory of Hilbert space embedding of distributions in Section \ref{subsec:kernel_embedding_review}. Then we propose the MUL model for GLS in Section \ref{subsec:MAL_model}, which is generally applicable to UDA and PDA. Finally, we propose the \textit{conditional metric operator} and empirical estimation for discrepancy optimization in Section \ref{subsec:conditional_metric_operator}, which show some appealing theoretical properties.

\subsection{Kernel Embedding Theory}\label{subsec:kernel_embedding_review}
Recently, the distribution embedding theory of RKHS has been successfully applied in various fields \cite{muandet2017kernel}, \eg, generative model, distribution testing and statistical models. Besides, many solid foundations of the embedding theory also have been explored for theoretical guarantees, \eg, the characteristic properties for marginal distributions via kernel mean embedding \cite{smola2007hilbert,gretton2012kernel} and conditional distributions via Conditional Mean Embedding (CME) \cite{song2009hilbert,klebanov2020rigorous}.

In the following, we denote the measure space, which includes a $\sigma$-algebra $\MC{B}$ on $\MC{X}$ and a probability measure $P_X$ on $\MC{B}$, by $(\MC{X}, \MC{B}, P_X)$. Let the set of probability measures with respect to $\MC{B}$ on $\MC{X}$ be $\text{Pr}(\MC{X})$. A RKHS on $\MC{X}$, which is uniquely defined by a kernel function $k_\MC{X}(\cdot,\cdot)$, is denoted as $(\MC{H}_\MC{X},k_{\MC{X}})$. The kernel feature map $\phi(\cdot): \MC{X} \rightarrow \MC{H}_{\MC{X}}$ induced by $k_{\MC{X}}$ is $\phi (x) = k_{\MC{X}} (x,\cdot)$. For any $\phi(x)$ and $\phi(x')$ in $\MC{H}_\MC{X}$, their inner product is defined by $\left< \phi(x),\phi(x') \right>_{\MC{H}_\MC{X}}=k_\MC{X}(x,x')$. Further, the feature map satisfies the well-known reproducing property: $\left< f,\phi(x) \right>_{\MC{H}_\MC{X}}=f(x), ~\forall f \in \MC{H}_{\MC{X}}$.

With the map $\phi$, we can consider the mean element $\mu_X$ in RKHS $\MC{H}_\MC{X}$ with law $P_X$, which is known as \textit{kernel mean embedding} \cite{smola2007hilbert}. It has been proved that the embedding
\begin{equation}\label{eq:KME-definition}
   \mu_X:~ \text{Pr}(\MC{X}) \rightarrow \MC{H}_{\MC{X}}, \quad P_X \longmapsto \MBB{E}_X \left[ \phi (X) \right]
\end{equation}
is injective if $k_\MC{X}$ is universal. Formally, this embedding is written as $\mu_X =\MBB{E}_X \left[ \phi (X) \right]$. If the expectation $\MBB{E}_X [ \sqrt{k_{\MC{X}}(X,X)} ] \leq \infty$, the embedding $\mu_X$ will be still an element in $\MC{H}_\MC{X}$ and $\left< \mu_X, f \right>_{\MC{H}_\MC{X}} = \MBB{E}_X \left[ f(X) \right]$. With the ability of characterizing the distributions, kernel mean embedding is exploited to induce the popular two sample test, \ie, MMD \cite{gretton2012kernel}.

Beyond the marginal distribution embedding of single variable $X$, it is natural to consider whether there also exists an embedding for describing the interaction between random variables $X$ and $Y$. Let $\psi(\cdot)$ be the feature map of RKHS $\MC{H}_{\MC{Y}}$ on $\MC{Y}$ with kernel $k_{\MC{Y}}$. The definition of CME was firstly introduced by Song \etal \cite{song2009hilbert}. They define the conditional embedding operator $\MC{C}_{X|Y}$ which allows to deduce the conditional mean in $\MC{H}_\MC{X}$ as
\begin{equation}\label{eq:CME-early-definition}
   \mu_{X|y}=\MC{C}_{X|Y}\phi (y),~~ \forall y\in\MC{Y}.
\end{equation}
Unfortunately, Eq.~\eqref{eq:CME-early-definition} is not globally well-defined. For example, when $X$ and $Y$ are independent, the conditional mean is supposed to be $\mu_{X|y}=\mu_{X}$ while the result obtained from Eq.~\eqref{eq:CME-early-definition} will be $\mu_{X|y}=0$ as $\MC{C}_{X|Y}=0$. To overcome this pathology, a rigorous theory on CME was recently proposed by Klebanov \etal \cite{klebanov2020rigorous}. They present a more general definition of CME based on the \textit{uncentered} operators which relax the assumption in centered operators. The \textit{uncentered} cross-covariance operator $^{u}\MBF{R}_{XY}: \MC{H}_{\MC{Y}} \rightarrow \MC{H}_{\MC{X}}$ is defined as $^{u}\MBF{R}_{XY} = \MBB{E}_{(X,Y)} \left[ \phi (X) \otimes \psi (Y) \right].$
Then the conditional embedding operator and CME in Eq.~\eqref{eq:CME-early-definition} can be redefined as
\begin{equation*}\label{eq:CME-rigorous-definition}
   ^{u}\MC{C}_{X|Y} = {^{u}\MBF{R}_{XY}} {^{u}\MBF{R}^{-1}_{YY}}, \quad \mu_{X|y}= {^{u}\MC{C}_{X|Y}} \psi (y).
\end{equation*}
This embedding rule implies that CME can be equivalently formulated as the image of its condition $\psi (y)$ under the mapping ${^{u}\MC{C}_{X|Y}}$. Note that in conditional embedding theory, the estimation of conditional mean $\mu_{X|y}$ is reformulated as the estimation of operator ${^{u}\MC{C}_{X|Y}}$ based on all conditions (whole dataset). For simplicity, we denote the uncentered cross-covariance operator as $\MBF{R}_{XY}$ and the uncentered conditional embedding operator as ${\MC{C}_{X|Y}}$ hereinafter.


\subsection{Conditional Invariant Transformation via MUL}\label{subsec:MAL_model}
Let the superscripts of variables imply the domains, \eg, ($X^s,Y^s$) for the source domain. We consider the one-hot encoding for $Y$, then $\MC{Y}=\{\MBF{e}_1,\MBF{e}_2,\cdots,\MBF{e}_c\}$, where $\MBF{e}_i$ is the standard basis in $\MBB{R}^c$. Though many marginal distribution matching methods have been designed based on the generalization theory of Ben-David \etal \cite{ben2010theory}, recent advancement \cite{zhao2019learning} and our theoretical results in Section \ref{sec:theoretical_insight} show that the marginal matching models will yield a larger generalization error with the misaligned conditional distributions $P^s_{Z|y_i} \neq P^t_{Z|y_i}$. Specifically, since the marginal distribution obeys the law of total probability $P_Z= \sum_{i} P_{Z|y_i}p_{y_i} $, it is error-prone to align the marginal distributions $P^s_Z$ and $P^t_Z$ when the \textit{label shift} problem exists, \ie, $P^s_Y \neq P^t_Y$. In this section, we present the MUL to deal with these problems.

Let $\MC{D}^s = \{(\MBF{x}_i^s,\MBF{y}_i^s)\}_{i=1}^{{n_s}}$ and $\MC{D}^t = \{\MBF{x}_i^t\}_{i=1}^{{n_t}}$ be i.i.d. observations drawn according to source distribution $P^s_{XY}$ and target distribution $P^t_{X}$, respectively. Here we abuse the notation for simplicity of expression and let $\MBF{y}_i$ be the $i$-th observation. To learn the conditional invariant representations, we design a transformation $G(\cdot)$ for covariate $X$ and consider the conditional distributions on $Z=G(X)$, \ie, $P_{Z|Y}$. Then task classifier $F(\cdot)$ is trained to ensure the prediction $\hat{Y}=F(Z)$ is close to the ground-truth label $Y$, where $\hat{Y}\in \MBB{R}^c$ lying in the probability simplex s.t. $\sum_i \hat{Y}_i = 1$. The flowchart of MUL model is presented in Figure \ref{fig:flowchart_MAL}. It attempts to identify and match the conditional distributions across domains while seeking the transformed distributions with higher discriminability. Specifically, MUL deals with the uncertain region during knowledge transfer by maximizing the discrepancies between different conditional distributions $P_{Z|y_i} ~ (y_i\in \MC{Y})$. As the result in Theorem \ref{thm:source_error_and_task_uncertainty}, the error rate of optimal decision rule $P_{Y|Z}$ will be lower than $P_{Y|X}$ built on the raw space $\MC{X}$. For the cross-domain conditional distributions, MUL minimizes the weighted conditional discrepancy to correct GLS.

As discussed before, a well-defined measure for conditional discrepancy is crucial for MUL. Based on CME in Section \ref{subsec:kernel_embedding_review}, Maximum Conditional Mean Discrepancy (MCMD) between $P^s_{Z|y_i}$ and $P^t_{Z|y_i}$ was recently introduced by Park \etal \cite{park2020measure}. The original definition of MCMD cannot straightforwardly measure the conditional discrepancy between $P^s_{Z|y_i}$ and $P^t_{Z|y_j}$ ($i\neq j$), which limits the application to inter-class discrepancy measure. To overcome this limitation and further deal with GLS, we generalize the MCMD by: 1) giving a more general definition and empirical estimation for case $y_i\neq y_j$; 2) proposing the \text{conditional metric operator} which guarantees some nice properties for empirical MCMD; 3) alleviating the cubic computational complexity \wrt~sample-size. Generally, ref. \cite{park2020measure} focuses on removing the stringent assumptions in CME, while our work focuses on a generally applicable conditional discrepancy with appealing empirical property and efficiency. Therefore, we redefine the MCMD as the discrepancy between $P^s_{Z|y_i}$ and $P^t_{Z|y_j}$, i.e.,
\begin{equation*}
   \text{MCMD}(P^s_{Z|y_i},P^t_{Z|y_j}) = \left\| \mu^s_{Z|y_i} - \mu^t_{Z|y_j}  \right\|_{\MC{H}_{\MC{Z}}}.
\end{equation*}

According to the mean embedding property in Eq.~\eqref{eq:KME-definition}, if the reproducing kernel $k_\MC{Z}$ is universal, then the conditional mean embedding $\mu_{Z|y}:~ P_{Z|y} \longmapsto \MBB{E}_{Z|y} \left[ \phi (Z) | Y=y\right]$
will also be injective. This means that each conditional distribution $P_{Z|y}$ has a unique embedding $\mu_{Z|y}$ in $\MC{H}_{\MC{Z}}$. With the injective property, we conclude that the redefined MCMD is indeed a metric on conditional distributions.
\begin{theorem}\label{thm:MCMD-metric-property}
If the kernels $k_{\MC{Z}}$ and $k_{\MC{Y}}$ are universal, then
\begin{equation*}
   \emph{MCMD}(P^s_{Z|y_i},P^t_{Z|y_j}) = \left\| \MC{C}^s_{Z|Y} \psi(y_i) - \MC{C}^t_{Z|Y} \psi(y_j) \right\|_{\MC{H}_{\MC{Z}}}
\end{equation*}
is a metric on $P_{Z|Y}$, \ie, $\emph{MCMD}(P^s_{Z|y_i},P^t_{Z|y_j})=0 $ if and only if $ P^s_{Z|y_i} = P^t_{Z|y_j}$.
\end{theorem}

\begin{figure}
   \begin{center}
       \includegraphics[width=0.46\textwidth,trim=25 18 15 20,clip]{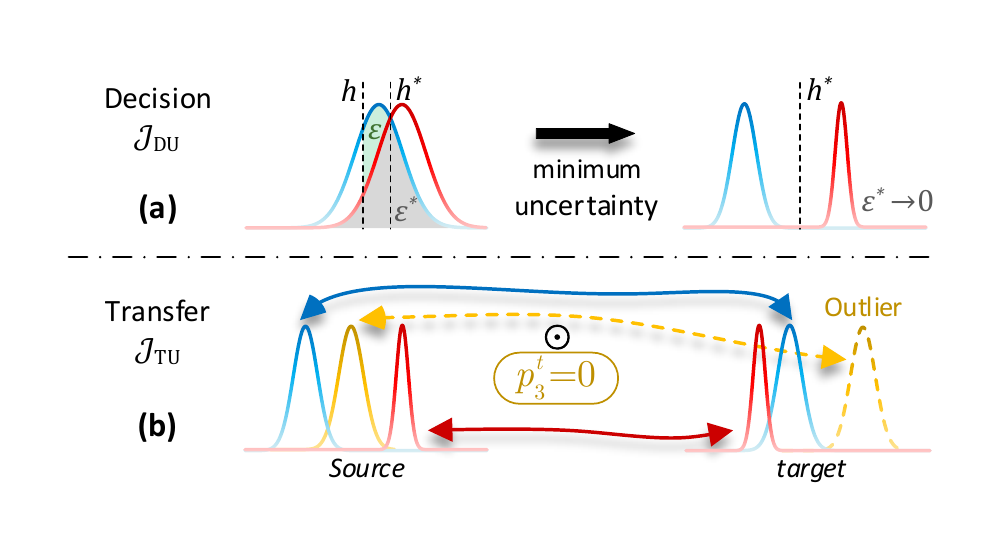}
       \caption{Illustration of $\MC{J}_{\text{DU}}$ and $\MC{J}_{\text{TU}}$. (a) the error $\varepsilon^*$ of optimal hypothesis $h^*$ is the (gray) overlapping region of the conditional distributions, and other hypothesis $h$ induces an additional error $\varepsilon$ (green region). The larger discrepancy between conditional distributions guarantee a smaller $\varepsilon^*$. (b) under GLS, some classes may be rare or even absent on the target domain; these classes are suppressed by weights $p_{y_i}^t$ during matching. Best viewed in color.}
       \label{fig:objective_MUL}
   \end{center}
   \vskip -0.2in
\end{figure}

Based on the MCMD metric, we consider the maximum discrepancy model for decision uncertainty minimization on all $P_{Z|Y=y_i}$. With this goal in mind, we formulate the following discrepancy-based objective:
   \begin{equation*}\label{eq:objective_discriminative}
   \MC{J}_{\text{DU}}(G) = \sum_{i \neq j} \text{MCMD}^2(P^s_{Z|y_i},P^s_{Z|y_j}).
   \end{equation*}
As shown in Figure \ref{fig:objective_MUL} (a), the error $\varepsilon^*$ of optimal hypothesis $h^*$ is lower than any other hypothesis $h$ whose error is $\varepsilon^*+\varepsilon$. The term $\MC{J}_{\text{DU}}$ tries to maximize the divergences between the transformed distributions under different conditions, which means that the distributions of different classes are supposed to be significantly distinct. Consequently, $\MC{J}_{\text{DU}}$ will alleviate the overlap of conditional distributions, which also improves the accuracy of optimal classifier by minimizing the area of uncertain region ($\varepsilon^*$). Finally, the transfer model will be encouraged to explore the discriminative structure and reduce the uncertainty and risk on decision-making. 

Similarly, the conditional matching objective for transfer uncertainty can be formulated based on the MCMD metric:
\begin{equation}\label{eq:objective_conditional-matching}
   \MC{J}_{\text{TU}}(G) = \sum_{i} p_{y_i}^t \cdot \text{MCMD}^2\left( P^s_{Z|y_i}, P^t_{Z|y_i} \right).
\end{equation}
As shown in Figure \ref{fig:objective_MUL} (b), by minimizing the transfer uncertainty $\MC{J}_{\text{TU}}$, the model is encouraged to transfer the learned discriminative structure to the target domain correctly. Note the weights $p_{y_i}^t$ is crucial for conditional adaptation in the presence of label shift. This figure shows an extreme scenario for GLS (\ie, PDA), where the outlier class (\ie, the classes that not appear on the target domain) of the source domain is masked by $p_{y_i}^t=0$. Therefore, the weighted conditional matching in Eq. \eqref{eq:objective_conditional-matching} will keep the consistency between the prior probabilities of the target and the adjusted source domain, and mitigate the negative transfer by suppressing the impact of outlier classes. Since $\MC{J}_{\text{TU}}$ incentivizes the matching of the cluster structures (\ie, $P_{Z|Y}$), the accuracy of the hypothesis (classifier) trained on the source domain will be improved on the target domain.

To learn the basic transformation $G$ and classifier $F$ for classification task, we employ weighting ERM with cross-entropy loss. We optimize $G$ and $F$ as
\begin{equation}\label{eq:objective_entropy_functions}
   \MC{J}_{\text{E}}(G,F) = \hat{\varepsilon}^G_{\tilde{s}} = \sum_{i,j} - w(\MBF{y}_i)\cdot \MBF{y}^s_{ji} \log \hat{\MBF{y}}^s_{ji}.
\end{equation}
As weighting ERM is an unbiased estimator of target true risk, $\MC{J}_{\text{E}}$ will encourage $F\circ G$ to approximate the posterior $p^t_{Y|X}$. Following the principle in Eq.~\eqref{eq:MAL_framework}, we apply the importance weighted ERM on the source domain as $\MC{J}_{\text{E}}$, which will mitigate the bias induced by label shift.
By integrating the above learning criteria, the final objective of MUL is
\begin{equation}\label{eq:objective_MAL-final}
  \min_{G,F} ~ \MC{J}_{\text{MUL}} = \MC{J}_{\text{E}}(G,F) + \lambda_{\text{TU}} \MC{J}_{\text{TU}}(G) - \lambda_{\text{DU}} \MC{J}_{\text{DU}}(G).
\end{equation}
In the unsupervised setting for domain adaptation, we have no access to the labels of the target domain, which means that the posterior $p^t_{Y|X}$ and prior $p^t_{Y}$ are unknown. A common solution for the posterior estimation is to approximate $p^t_{Y|X}$ with predictor $F\circ G(X)$. For prior estimation, we employ the Black Box Shift Estimation (BBSE) \cite{lipton2018detecting,combes2020domain} which works even the predictor is biased and inaccurate. The algorithm of BBSE will be detailed in Section \ref{subsec:algorithm}. Note the conditional discrepancy MCMD is built on the implicit RKHS. We will present the explicit formulation in the next.

\subsection{Conditional Metric Operator}\label{subsec:conditional_metric_operator}

In this section, we propose the conditional metric operator to bridge the gap between MCMD in RKHS and its explicit formulation in Euclidean space. Different from previous methods estimating the class-wise discrepancies with the pseudo label separately, we embed the conditional discrepancy into the RKHS with the conditional metric operator. The conditional metric operator integrates the conditional relations under all conditions ${y_i}\in\MC{Y}$, which guarantees that the discrepancy is determined only by the operator itself. As the distributions are processed in spaces $\MC{Z}$ and $\MC{Y}$, we denote their mapped features in RKHSs by $\MBF{\Phi}^{s/t}$ and $\MBF{\Psi}^{s/t}$, respectively. The empirical estimation of the uncentered cross-covariance operator is $\hat{\MBF{R}}_{ZY} = \frac{1}{n} \MBF{\Phi}\MBF{\Psi}^T$. Note the kernel feature maps $\phi(\cdot)$ and $\psi(\cdot)$ are implicit, thus we will process the kernel-based model with the well-known \textit{kernel trick}, \ie, $\MBF{\Phi}^T\MBF{\Phi}=\MBF{K}$ and $\MBF{\Psi}^T\MBF{\Psi}=\MBF{L}$, where $ \MBF{K}_{ij} = k_{\MC{Z}}(\MBF{z}_i,\MBF{z}_j) $ and $ \MBF{L}_{ij} = k_{\MC{Y}}(\MBF{y}_i,\MBF{y}_j) $. Note the MCMD between $P^s_{Z|y_i}$ and $P^t_{Z|y_j}$ can be written as 
\begin{eqnarray}
    & & \text{MCMD}^2(P^s_{Z|y_i},P^t_{Z|y_j}) \nonumber \\
    \!\!&\!\!=\!\!&\!\! \big< \MC{C}^s_{Z|Y}\psi(y_i)\!-\!\MC{C}^t_{Z|Y}\psi(y_j),~\MC{C}^s_{Z|Y}\psi(y_i)\!-\!\MC{C}^t_{Z|Y}\psi(y_j) \big> \label{eq:MAL-Mahalanobis-distance-Hilbert},  \\
    \!\!&\!\!=\!\!&\!\! \big<\psi(y_i), \MBF{M}^{ss}_{\MC{H}} \psi(y_i) \big>\!+\!\left<\psi(y_j), \MBF{M}^{tt}_{\MC{H}}\psi(y_j) \right>
   \!-\!2\left<\psi(y_i), \MBF{M}^{st}_{\MC{H}}\psi(y_j)\right> \nonumber 
\end{eqnarray}
\noindent where $\MBF{M}^{st}_{\MC{H}} = (\MC{C}^s_{Z|Y})^* \MC{C}^t_{Z|Y}$ is an operator on $\MC{H}_{\MC{Y}}$. Specially, when MCMD is considered only for different conditions $y_i$, \eg, decision uncertainty $\MC{J}_{\text{DU}}$, Eq.~\eqref{eq:MAL-Mahalanobis-distance-Hilbert} is simplified as
\begin{align}
   \text{MCMD}^2(&P_{Z|y_i},P_{Z|y_j}) \nonumber \\
   =& \left< \left[ \psi(y_i) - \psi(y_j) \right], \MBF{M}_{\MC{H}} \left[ \psi(y_i) - \psi(y_j) \right] \right>, \label{eq:MAL-Mahalanobis-distance-Hilbert-simple}
\end{align}
where $\MBF{M}_{\MC{H}} = \MC{C}^*_{Z|Y} \MC{C}_{Z|Y}$. We call $\MBF{M}_{\MC{H}}$ the \textit{conditional metric operator} because it serves as the invariant coefficient of conditional discrepancy measure Eq.~\eqref{eq:MAL-Mahalanobis-distance-Hilbert} or Eq.~\eqref{eq:MAL-Mahalanobis-distance-Hilbert-simple} for any input condition pair $(y_i,y_j)$. Since $\psi(y_i)$ is predetermined by the label variable $Y$, the conditional discrepancy $\text{MCMD}(P^s_{Z|y_i},P^t_{Z|y_j})$ is dominated by the coefficient $\MBF{M}_{\MC{H}}$ of the quadratic form in Eq.~\eqref{eq:MAL-Mahalanobis-distance-Hilbert} or Eq.~\eqref{eq:MAL-Mahalanobis-distance-Hilbert-simple}. Note $\MBF{M}_{\MC{H}}$ is parameterized by the learnable variable $Z$, which means that the conditional distributions are optimized by the parameters \wrt~transformation $G(\cdot)$.

As the operators in Eq.~\eqref{eq:MAL-Mahalanobis-distance-Hilbert} and Eq.~\eqref{eq:MAL-Mahalanobis-distance-Hilbert-simple} are still built on infinite-dimensional Hilbert space, now we reformulate the implicit discrepancy on the explicit Euclidean space with the observed data. Since the operator $\MBF{R}_{YY}$ is generally singular in finite-sample case, we regularize it as $ \MBF{R}_{YY} + \varepsilon \MBF{I}$. The empirical estimations of the uncentered conditional embedding operator and element are given by
\begin{equation*}\label{eq:conditional-embeddings_empirical}
   \begin{array}{c}
      \hat{\MC{C}}_{Z|Y} = \hat{\MBF{R}}_{ZY} ( \hat{\MBF{R}}_{YY} + \varepsilon \MBF{I})^{-1} =
      \MBF{\Phi} \tilde{\MBF{L}}^{-1} \MBF{\Psi}^T, \\
      \hat{\mu}_{Z|y} = \hat{\MC{C}}_{Z|Y}\psi(y) = \MBF{\Phi} \tilde{\MBF{L}}^{-1} \MBF{\Psi}^T \psi(y) = \MBF{\Phi} \tilde{\MBF{L}}^{-1} \MBF{L}_{y},
   \end{array}
\end{equation*}
where $\tilde{\MBF{L}} = \varepsilon n \MBF{I}_n + \MBF{L}$ and $\MBF{L}_{y}$ equals to the $i$-th column of $\MBF{L}$ such that $\MBF{y}_i = y$. Now we  derive the empirical estimation of $\MC{J}_{\text{DU}}$ based on source data with ground-truth label.
\begin{align}
   \MC{J}_{\text{DU}} &= \sum_{i \neq j} \text{MCMD}^2(\hat{P}^s_{Z|y_i},\hat{P}^s_{Z|y_j}) \nonumber \\
   &= \sum_{i \neq j} \left( \MBF{L}^s_{y_i} - \MBF{L}^s_{y_j} \right)^T \MBF{M}^{ss}_{\MBB{R}} \left( \MBF{L}^s_{y_i} - \MBF{L}^s_{y_j} \right) \label{eq:MAL-Mahalanobis-distance-Euclid},
\end{align}
where $\MBF{M}^{ss}_{\MBB{R}} = \tilde{\MBF{L}}^{s^{-1}} \MBF{K}^{ss} \tilde{\MBF{L}}^{s^{-1}}$ is the conditional matrix in Euclidean space $\MBB{R}^{n}$. Eq.~\eqref{eq:MAL-Mahalanobis-distance-Euclid} can also be taken as the Mahalanobis distance with parameter $\MBF{M}_{\MBB{R}}$. Similarly, the conditional discrepancy in $\MC{J}_{\text{TU}}$ Eq.~\eqref{eq:objective_conditional-matching} is estimated as
\begin{align}
   \MC{J}_{\text{TU}}
   =& \sum_{i} \text{MCMD}^2(\hat{P}^s_{Z|y_i},\hat{P}^t_{Z|y_i}) \nonumber \\
   =& \sum_{i} p^t_{y_i} \left\| \left[ \hat{\MC{C}}^s_{Z|Y} - \hat{\MC{C}}^t_{Z|Y} \right] \psi(y_i)  \right\|^2_{\MC{H}_{\MC{Z}}} \nonumber \\
   =& \sum_{i} p^t_{y_i}
   {\underbrace{\left( \setlength{\arraycolsep}{0.7pt} \begin{array}{c}\MBF{L}^s_{y_i}\\ -\MBF{L}^t_{y_i} \end{array}\right)}_{\MBF{L}^a_{y_i}}} ^T
   \underbrace{\left(\begin{array}{cc} \MBF{M}^{ss}_{\MBB{R}} & \MBF{M}^{st}_{\MBB{R}} \\
   \MBF{M}^{ts}_{\MBB{R}} & \MBF{M}^{tt}_{\MBB{R}} \end{array}\right)}_{\MBF{M}^{aa}_{\MBB{R}}}
   \left( \setlength{\arraycolsep}{0.7pt} \begin{array}{c}\MBF{L}^s_{y_i}\\ -\MBF{L}^t_{y_i} \end{array}\right),
    \label{eq:Conditional-Matching-esitimation}
\end{align}
where $\MBF{M}^{st}_{\MBB{R}} = \tilde{\MBF{L}}^{s^{-1}} \MBF{K}^{st} \tilde{\MBF{L}}^{t^{-1}}$ and $(\MBF{M}^{st}_{\MBB{R}})^T = \MBF{M}^{ts}_{\MBB{R}}$. Note that Theorem \ref{thm:MCMD-metric-property} proves that MCMD is a metric on conditional distributions while we can only estimate the empirical MCMD with finite observations in real-world applications. For example, the empirical discrepancy $D(\hat{P}_{X_1}, \hat{P}_{X_2})$ between two different distributions ($P_{X_1} \neq P_{X_2}$) may be zero since the metric property may no longer hold with finite observations. To study the property of empirical MCMD, we present some theoretical results which guarantee the identifiability of empirical estimation in finite-sample case.
\begin{lemma}\label{lem:conditional-norm-metric}
   If the kernel $k_{\MC{Z}}$ is strictly positive definite, then norm $\| \cdot \|_{\MBF{M}_{\MBB{R}}}$ induced from $\| \MBF{a} \|^2_{\MBF{M}_{\MBB{R}}} = \left< \MBF{a}, \MBF{M}_{\MBB{R}} \MBF{a} \right> $ is a metric on $\MBB{R}^n$.
\end{lemma}

Lemma \ref{lem:conditional-norm-metric} suggests that $\| \cdot \|_{\MBF{M}_{\MBB{R}}}$, which is deduced from the finite-sample estimation of $\MBF{M}_{\MBB{R}}$, is well-defined on $\MBB{R}^n$. Note that $\MBF{M}_{\MBB{R}}$ could be the conditional matrix $\MBF{M}^{ss}_{\MBB{R}}$ in Eq.~\eqref{eq:MAL-Mahalanobis-distance-Euclid} or the $2\times 2$ block conditional matrix $\MBF{M}^{aa}_{\MBB{R}}$ in Eq.~\eqref{eq:Conditional-Matching-esitimation}. The key to prove this claim is to show that $\MBF{M}_{\MBB{R}}$ is positive definite. Now the discrepancy optimization model on $\MC{H}_{\MC{Z}}$ is formulated as a metric learning problem on $\MBB{R}^n$. The transformation $G(\cdot)$ will optimize the conditional matrices $\MBF{M}_{\MBB{R}}$, which characterizes the discriminability and transferability with the distribution embeddings in Hilbert space.

Generally, when the conditional distributions of the source and target domains are not sufficiently aligned, the block matrix $\MBF{M}^{aa}_{\MBB{R}}$ will always positive definite. As the vectors $\MBF{L}^a_{y}$ are non-zero, the objective $\MC{J}_{\text{TU}}$ will be strictly positive. Thus, the formulation in Eq.~\eqref{eq:Conditional-Matching-esitimation} and Lemma \ref{lem:conditional-norm-metric} implies that the conditional discrepancy in $\MC{J}_{\text{TU}}$ will be non-zeros unless the conditional embedding operators $\hat{\MC{C}}_{Z|Y}$ are matched. For the decision uncertainty $\MC{J}_{\text{DU}}$, we can further extend the metric to the space of conditional variable $Y$.
\begin{theorem}[Identifiability]\label{thm:metric-on-y}
   If $k_{\MC{Z}}$ and $k_{\MC{Y}}$ are strictly positive definite, $\hat{D}(y_i,y_j)=\emph{MCMD}(\hat{P}_{Z|y_i},\hat{P}_{Z|y_j})$ is a metric on $\MC{Y}$.
\end{theorem}

Theorem \ref{thm:metric-on-y} ensures that the empirical estimation of $\MC{J}_{\text{DU}}$ is always well-defined, \ie, $\text{MCMD}(\hat{P}_{Z|y_i},\hat{P}_{Z|y_j})$ is non-zero if and only if $i \neq j$. The identifiability of MCMD are crucial for gradient-based optimization. It ensures that objectives is always non-zeros and the gradients won't vanish unless the conditional distributions are the same. Note the conditions (universal, strictly positive definite) are satisfied by some common kernels on $\MBB{R}^{n}$, \eg, Gaussian and Laplacian \cite{sriperumbudur2011universality}.

\section{Model Analysis and Algorithm}\label{sec:theory_analysis_implementation}

We first prove the statistical consistency of the empirical MUL in Section \ref{subsec:statistical_consistency_MAL}. Then the complexity analysis is presented in Section \ref{subsec:complexity_analysis_MAL}. Finally, the schematic algorithm is presented in Section \ref{subsec:algorithm}.

\subsection{Statistical Consistency}\label{subsec:statistical_consistency_MAL}

Now we focus on the statistical consistency of empirical estimations in Eq.~\eqref{eq:MAL-Mahalanobis-distance-Euclid} and Eq.~\eqref{eq:Conditional-Matching-esitimation}. Our results are mainly based on the consistency of the (cross-)covariance operators which converge in the Hilbert-Schmidt norm at $O_{P}(m^{-\frac{1}{2}})$ \cite{fukumizu2007statistical}. From the definition of operator norm $\| \cdot \|$ and the inequality $\| \cdot \| \leq \| \cdot \|_{HS}$, we have
\begin{align*}
   &\big\| \mu_{Z|y_i} - \mu_{Z|y_j}  \big\|_{\MC{H}_{\MC{Z}}} \leq \big\| \MC{C}_{Z|Y} \big\|_{HS} \big\| \psi(y_i)-\psi(y_j) \big\|_{\MC{H}_{\MC{Z}}}, \\
   &\big\| \mu^s_{Z|y_i} - \mu^t_{Z|y_i}  \big\|_{\MC{H}_{\MC{Z}}} \leq \big\| \MC{C}^s_{Z|Y} - \MC{C}^t_{Z|Y} \big\|_{HS} \big\| \psi(y_i)\big\|_{\MC{H}_{\MC{Z}}}.
\end{align*}
These inequalities show that the empirical estimations are bounded by the embedding operators. Let $\hat{\MC{C}}^{(m)}_{Z|Y}$ be the embedding operator estimated from $m$ i.i.d samples. As the singularity problem of $\hat{\MBF{R}}_{YY}$ is associated with the sample size, we define the regularization parameter $\varepsilon$ of $\hat{\MBF{R}}_{YY}$ as a series depend on $m$, \ie, $\varepsilon_m$. Denote the conditional divergence and domain discrepancy by $D_Y=\| \mu_{Z|y_i} - \mu_{Z|y_j}  \|^2_{\MC{H}_{\MC{Z}}}$ and $D_D=\| \mu^s_{Z|y_i} - \mu^t_{Z|y_i}  \|^2_{\MC{H}_{\MC{Z}}}$, respectively. Now we present the asymptotic properties of the uncentered conditional embedding operator and estimated discrepancies.

\begin{theorem}[Consistency]\label{thm:consistency-theorem}
   Assuming $\varepsilon_m$ satisfies that $\varepsilon_{m} \to 0$ and $\varepsilon_{m}^{\frac{3}{2}} m^{\frac{1}{2}} \to \infty$ ($m \to \infty$), then
   \begin{equation*}
      \left\| \MC{C}_{Z|Y} - \hat{\MC{C}}^{(m)}_{Z|Y} \right\|_{HS} \rightarrow 0 \quad (m \to \infty)
   \end{equation*}
   in probability with rate $O_{P}(\varepsilon_{m}^{-\frac{3}{2}} m^{-\frac{1}{2}})$. Further we have the empirical estimation of MUL is consistent, \ie, $\hat{D}^{(m)}_Y \to D_Y$ and $\hat{D}^{(m)}_D \to D_D$ in probability with rate $O_{P}(\varepsilon_{m}^{-\frac{3}{2}} m^{-\frac{1}{2}})$.
\end{theorem}

Note the converge rate $O_{P}(\varepsilon_{m}^{-\frac{3}{2}} m^{-\frac{1}{2}})$ depends on the decay rate of $\varepsilon_{m}$. Specifically, if we set $\varepsilon_{m}=m^{-\alpha}~(\alpha>0)$, which satisfies the assumptions in Theorem \ref{thm:consistency-theorem}, then the converge rate becomes $O_{P}(m^{\frac{-1+3\alpha}{2}})$ which is $O_{P}(m^{-\frac{1}{2}})$ for $\alpha \to 0$. This result shows that the empirical estimations are consistent with rate nearly $O_{P}(m^{-\frac{1}{2}})$ if $\varepsilon_{m} \to 0$ slowly.

\subsection{Complexity Analysis}\label{subsec:complexity_analysis_MAL}

Kernel methods \cite{song2009hilbert,luo2021conditional} usually suffer from the cubic time complexity \wrt~sample-size. We introduce an efficient computation strategy to implement MUL in quadratic time. The computational complexity mainly consists of the uncertainty terms. To simplify the notations, we denote the dimension of variable $Z$ as $d$, number of classes as $c$ and $m=\max \{ {n_s},{n_t} \}$. For $\MC{J}_{\text{DU}}$ and $\MC{J}_{\text{TU}}$, the complexities of kernel matrices $\MBF{K}$ and $\MBF{L}$ of variables $Z$ and $Y$ are $\MC{O}(dm^2)$ and $\MC{O}(cm^2)$, respectively. The cubic time complexity $\MC{O}(m^3)$ is then induced by the inverse of kernel matrices $\tilde{\MBF{L}}$. Note that for ground-truth source labels and target pseudo-labels, the ranks of their corresponding kernel matrices $\MBF{L}^s$ and $\MBF{L}^t$ are $c$ since they are computed from data with $c$ distinct points (labels). Then there exists a rank-$c$ truncated eigendecomposition $\MBF{L}=\MBF{UD}\MBF{U}^T$ where $\MBF{U}\in \MBB{R}^{m\times c}, \MBF{D}\in \MBB{R}^{c\times c}$. Hopefully, the complexity of regularized inverse can be reduced to $\MC{O} (cm^2)$ via the Woodbury formula, \ie,
\vspace{-1pt}
\begin{equation*}\label{eq:truncated-evd}
    \left( \MBF{L} + \MBF{I}_n \right)^{-1} =
    \MBF{I}_n - \MBF{U} \MBF{D} \left( \MBF{D} + \MBF{I}_c \right)^{-1} \MBF{U}^T.
\end{equation*}
The quadratic form in Eq.~\eqref{eq:MAL-Mahalanobis-distance-Euclid} and Eq.~\eqref{eq:Conditional-Matching-esitimation} can be finished in $\MC{O}(cm^2)$. Specifically, the conditional discrepancies are computed by the matrix-vector product, while the conditional matrices will not be explicitly formulated in numerical computation. We conclude the overall complexity as follows and provide the details in the supplementary file.
\begin{proposition}\label{pro:quadratic-time-complexity}
   The complexity of uncertainty terms $\MC{J}_{\text{DU}}$ and $\MC{J}_{\text{TU}}$ is $\MC{O}((d+c)m^2)$. Then MUL can be implemented in quadratic time with complexity $\MC{O}(m^2)$.
\end{proposition}

The quadratic time complexity is usually required to obtain the exact solution for kernel methods. Here we also provide some insight into the approximate solution with linear time complexity. A common approximate method is the random projection \cite{rahimi2007random}. The key is to project the input $\MBF{Z}\in \MBB{R}^{d\times m}$ into a random feature space as $\MBF{S}\in \MBB{R}^{r\times m}$, where $r$ is the dimension of random projection. With the random feature, the kernel matrix is approximated as $\MBF{K} \approx \MBF{S}^T \MBF{S}$. Then the low-rank decomposition and explicit formulation of kernel matrix will be avoided, which have quadratic time complexity. Note that the approximation performance depends on the input dimension $d$ \cite[Claim 1]{rahimi2007random}, thus, $r$ does not rely on the sample size $m$. Following this idea, the approximate solution only compute matrix-vector product and its time complexity will be $\MC{O}(mr)$ (linear \wrt~$m$).

\begin{table*}[t]
   \centering
   \setlength{\abovecaptionskip}{0.0cm}
   \setlength{\belowcaptionskip}{-0.01cm}
   \caption{Classification accuracies (\%) on Office-Home, VisDA-2017, Office 31 and ImageCLEF datasets under vanilla setting. ResNet-101 is employed on VisDA-2017 and ResNet-50 on others. The superscripts denote standard deviations hereafter.}
   \label{tab:UDA_4dataset}

   \renewcommand{\tabcolsep}{0.3pc} 
   \renewcommand{\arraystretch}{1.0} 
   \begin{tabular}{c|cccccccccccc|c}
   \toprule[1pt]
   \textbf{Office-Home} & Ar$\to$Cl & Ar$\to$Pr & Ar$\to$Rw & Cl$\to$Ar & Cl$\to$Pr & Cl$\to$Rw &
   Pr$\to$Ar & Pr$\to$Cl & Pr$\to$Rw & Rw$\to$Ar & Rw$\to$Cl & Rw$\to$Pr & Avg. \\
   \hline
   Source \cite{he2016deep} & 34.9 & 50.0 & 58.0 & 37.4 & 41.9 & 46.2 & 38.5 & 31.2 & 60.4 & 53.9 & 41.2 & 59.9 & 46.1 \\
   DAN \cite{long2018transferable} & 43.6 & 57.0 & 67.9 & 45.8 & 56.5 & 60.4 & 44.0 & 43.6 & 67.7 & 63.1 & 51.5 & 74.3 & 56.3 \\
   DANN \cite{ganin2015unsupervised} & 45.6 & 59.3 & 70.1 & 47.0 & 58.5 & 60.9 & 46.1 & 43.7 & 68.5 & 63.2 & 51.8 & 76.8 & 57.6 \\
   CDAN+E \cite{long2018conditional} & 50.7 & 70.6 & 76.0 & 57.6 & 70.0 & 70.0 & 57.4 & 50.9 & 77.3 & 70.9 & 56.7 & 81.6 & 65.8 \\
   SAFN \cite{xu2019larger} & 52.0 & 71.7 & 76.3 & 64.2 & 69.9 & 71.9 & 63.7 & 51.4 & 77.1 & 70.9 & 57.1 & 81.5 & 67.3 \\
   ETD \cite{li2020Enhanced} & 51.3 & 71.9 & \textbf{85.7} & 57.6 & 69.2 & 73.7 &  57.8 & 51.2 & 79.3 & 70.2 & 57.5 & 82.1 & 67.3\\
   ATM \cite{li2020maximum} & 52.4 & 72.6 & 78.0 & 61.1 & 72.0 & 72.6 & 59.5 & 52.0 & 79.1 & \textbf{73.3} & \textbf{58.9} & 83.4 & 67.9 \\
   DMP \cite{luo2020unsupervised} & 52.3 & 73.0 & 77.3 & \textbf{64.3} & 72.0 & 71.8 & 63.6 & 52.7 & 78.5 & 72.0 & 57.7 & 81.6 & 68.1 \\
   \hline
   MUL & \textbf{52.9} & \textbf{75.6} & 78.8 & 62.5 & \textbf{75.4} & \textbf{75.3} & \textbf{64.0} & \textbf{53.3} & \textbf{81.4} & 69.8 & 56.4 & \textbf{83.6} & \textbf{69.1} \\
   \bottomrule[1pt]
   \toprule[1pt]
   \textbf{VisDA-2017} & Plane & bcycl & bus & car & horse & knife & mcyle & person & plant & sktbrd & train & truck & Avg. \\
   \hline
   Source \cite{he2016deep}                   & 55.1 &  53.3 &  61.9 &  59.1 &  80.6 &  17.9 &  79.7 &  31.2 &  81.0 &  26.5 &  73.5 &  8.5 &  52.4 \\
   DAN \cite{long2018transferable}                    & 87.1 &  63.0 &  76.5 &  42.0 &  90.3 &  42.9 &  85.9 &  53.1 &  49.7 &  36.3 &  85.8 &  20.7 &  61.1 \\
   DANN \cite{ganin2015unsupervised}                    & 81.9 &  \textbf{77.7} &  82.8 &  44.3 &  81.2 &  29.5 &  65.1 &  28.6 &  51.9 &  54.6 &  82.8 &  7.8 &  57.4 \\
   CDAN+E \cite{long2018conditional}                & 85.2 &  66.9 &  83.0 &  50.8 &  84.2 &  74.9 &  88.1 &  74.5 &  83.4 &  76.0 &  81.9 &  38.0 &  73.7 \\
   SAFN \cite{xu2019larger}                       & 93.6 & 61.3 & \textbf{84.1} & 70.6 & 94.1 & 79.0 & 91.8 & \textbf{79.6} & 89.9 & 55.6 & 89.0 & 24.4 & 76.1 \\
   DMP  \cite{luo2020unsupervised}              & 92.1 & 75.0 & 78.9 & \textbf{75.5} & 91.2 & 81.9 & 89.0 & 77.2 & \textbf{93.3} & 77.4 & 84.8 & 35.1 & 79.3 \\
   \hline
   MUL & \textbf{94.5} & 76.5 & 81.1 & 70.7 & \textbf{94.9} & \textbf{94.8} & \textbf{92.9} & 76.4 & 91.7 & \textbf{91.7} & \textbf{90.0} & \textbf{37.8} & \textbf{82.8} \\
   \bottomrule[1pt]
   \end{tabular}
   \\[2pt]
   \renewcommand{\tabcolsep}{0.294pc} 
   \renewcommand{\arraystretch}{1.0} 
   \begin{tabular}{c|ccccccc|ccccccc}
   \toprule[1pt]
   \multirow{2}{*}{\textbf{Methods}} & \multicolumn{7}{c}{\textbf{Office-31}} & \multicolumn{7}{|c}{\textbf{ImageCLEF}} \\
   ~& A$\rightarrow$W & D$\rightarrow$W & W$\rightarrow$D & A$\rightarrow$D & D$\rightarrow$A & W$\rightarrow$A & Avg.  & I$\rightarrow$P & P$\rightarrow$I & I$\rightarrow$C & C$\rightarrow$I & C$\rightarrow$P & P$\rightarrow$C & Avg. \\
   \hline
   Source \cite{he2016deep}             & $68.4^{0.2}$ & $96.7^{0.1}$ & $99.3^{0.1}$ & $ 68.9^{0.2}$ & $62.5^{0.3}$ & $60.7^{ 0.3}$ & 76.1     & $74.8^{0.3}$ & $83.9^{0.1}$ & $91.5^{0.3}$ & $78.0^{0.2}$ & $65.5^{0.3}$ & $91.2^{0.3}$ & 80.7 \\
   DAN \cite{long2018transferable}             & $80.5^{0.4}$ & $97.1^{0.2}$ & $99.6^{0.1}$ & $ 78.6^{0.2}$ & $63.6^{0.3}$ & $62.8^{ 0.2}$ & 80.4 &     $74.5^{0.4}$ & $82.2^{0.2}$ & $92.8^{0.2}$ & $86.3^{0.4}$ & $69.2^{0.4}$ & $89.8^{0.4}$ & 82.5 \\
   DANN \cite{ganin2015unsupervised}             & $82.0^{0.4}$ & $96.9^{0.2}$ & $99.1^{0.1}$ & $ 79.7^{0.4}$ & $68.2^{0.4}$ & $67.4^{ 0.5}$ & 82.2     & $75.0^{0.3}$ & $86.0^{0.3}$ & $96.2^{0.4}$ & $87.0^{0.5}$ & $74.3^{0.5}$ & $91.5^{0.6}$ & 85.0 \\
   CDAN+E \cite{long2018conditional}       & $94.1^{0.1}$ & $98.6^{0.1}$ & $ \textbf{100.0}^{0.0}$ & $92.9^{0.2}$ & $71.0^{0.3}$ & $69.3^{0.3}$ & 87.7     & $77.7^{0.3}$ & $90.7^{0.2}$ & $97.7^{0.3}$ & $91.3^{0.3}$ & $74.2^{0.2}$ & $94.3^{0.3}$ & 87.7 \\
   SAFN \cite{xu2019larger}                & $88.8^{0.4}$ & $98.4^{0.0}$ & $99.8^{0.0}$ & $ 87.7^{1.3}$ & $69.8^{0.4}$ & $69.7^{ 0.2}$ & 85.7     & $78.0^{0.4}$ & $91.7^{0.5}$ & $96.2^{0.1}$ & $91.1^{0.3}$ & $77.0^{0.5}$ & $94.7^{0.3}$ & 88.1 \\
   ETD \cite{li2020Enhanced} & 92.1$  $ & \textbf{100.0}$  $ & \textbf{100.0}$  $ & 88.0$  $ & 71.0$  $ & 67.8$  $ & 86.2     & 81.0$  $ & 91.7$  $ & 97.9$  $ & 93.3$  $ & 79.5$  $ & 95.0$  $ & 89.7 \\
   ATM \cite{li2020maximum} & $\textbf{95.7}^{0.3}$ & $99.3^{0.1}$ & $ \textbf{100.0}^{0.0}$ & $\textbf{96.4}^{0.2}$ & $74.1^{0.2}$ & $73.5^{0.3}$ & 89.8     & $80.3^{0.3}$ & $\textbf{92.9}^{0.4}$ & $\textbf{98.6}^{0.4}$ & $93.5^{0.1}$ & $77.8^{0.3}$ & $\textbf{96.7}^{0.2}$ & 90.0 \\
   DMP \cite{luo2020unsupervised} & $93.0^{0.3}$ & $99.0^{0.1}$ & $ \textbf{100.0}^{0.0}$ & $91.0^{0.4}$ & $71.4^{0.2}$ & $70.2^{0.2}$ & 87.4     & $80.7^{0.1}$ & $92.5^{0.1}$ & $97.2^{0.1}$ & $90.5^{0.1}$ & $77.7^{0.2}$ & $96.2^{0.2}$ & 89.1 \\
   \hline
   MUL & $93.4^{0.3}$ & $99.0^{0.3}$ & $ \textbf{100.0}^{0.0}$ & $94.2^{0.3}$ & $\textbf{78.3}^{0.2}$ & $\textbf{77.2}^{0.3}$ & \textbf{90.4}     & $\textbf{81.4}^{0.1}$ & $\textbf{92.9}^{0.1}$ & $97.4^{0.1}$ & $\textbf{95.7}^{0.1}$ & $\textbf{81.7}^{0.1}$ & $95.9^{0.1}$ & \textbf{90.8} \\
   \bottomrule[1pt]
   \end{tabular}

   \vskip -0.15in
\end{table*}


\subsection{Algorithm}\label{subsec:algorithm}

\hspace{1.4em}\textbf{Optimization. } We employ DNNs to implement the MUL model as shown in Figure \ref{fig:flowchart_MAL}. The MUL can be optimized either in an end-to-end fashion with mini-batch Gradient Decent (GD), or in a stacked fashion with batch GD. We employ the latter fashion. Specifically, we froze the backbone DNNs and train the relatively shallow conditional transfer network, \ie, a conditional transfer network with pre-trained deep features as input. Compared to the end-to-end fashion, the advantages of used implementation can be summarized by: 1) the batch GD gathering the conditional information of entire dataset within single iteration which guarantees a more reliable estimation of MUL by Theorem \ref{thm:consistency-theorem}; 2) mini-batch data cannot guarantee to cover all classes which makes the model sensitive to the sample selection; 3) the training of the relative shallow networks is efficient.

\textbf{Estimating Importance Weight. } As MUL requires the importance weight $w(\MBF{y})$ in Eq.~\eqref{eq:objective_entropy_functions} and prior probability $p^t_{Y}$ in Eq.~\eqref{eq:objective_conditional-matching} to correct the label shift, we employ the Quadratic-Programming (QP) variant \cite{combes2020domain} of BBSE \cite{lipton2018detecting} to estimate them. Since $w(\MBF{y})$ only has $c$ values, we rewrite it as $\MBF{w} = \left(w(\MBF{e}_1),w(\MBF{e}_2),\cdots, w(\MBF{e}_c)\right)^T\in \MBB{R}^c$. The vector of probability mass functions $p^{s/t}_{Y}$ is denoted as $\MBF{p}^{s/t}\in \MBB{R}^c$. The main idea of BBSE is to detect and measure the label shift via the predictor $F(\cdot)$. Based on the predictor's output $\hat{Y}=F(Z)$, BBSE considers the distributions on $\hat{Y}$ and $(Y,\hat{Y})$. Similarly, $p^t_{\hat{Y}}$ and $p^s_{Y\hat{Y}}$ are rewritten as $\MBF{q}^t\in \MBB{R}^c$ and $\MBF{C}\in\MBB{R}^{c\times c}$, respectively. Then the QP problem of BBSE is:
\begin{equation}\label{eq:BBSE_QP_estimation}
      \min_{\hat{\MBF{w}}} ~ \| \hat{\MBF{q}}^t - \hat{\MBF{C}} \hat{\MBF{w}}\|_2^2 \quad \quad \text{s.t.} ~~ \hat{\MBF{w}} \geq 0,~ \hat{\MBF{w}}^T \hat{\MBF{p}}^{s}=1,
\end{equation}
where ``$\geq$'' is element-wise operation, and $\hat{\MBF{p}}$, $\hat{\MBF{q}}$, $\hat{\MBF{C}}$ are the plug-in estimation of $\MBF{p}$, $\MBF{q}$, $\MBF{C}$ \cite{lipton2018detecting}. Once $\hat{\MBF{w}}$ is computed, the prior probability can be estimated as $\hat{\MBF{p}}^t=\hat{\MBF{w}}\odot \hat{\MBF{p}}^s$. The convergence of BBSE was proved by Lipton \etal \cite{lipton2018detecting}.

{
\begin{algorithm}[!t]
   \caption {MUL for GLS Correction}\label{alg:MAL}
   \begin{algorithmic}[1]
   \REQUIRE {Source data $\MC{D}^s = \{(\MBF{x}_i^s,\MBF{y}_i^s)\}_{i=1}^{{n_s}}$, Target data $\MC{D}^t = \{\MBF{x}_i^t\}_{i=1}^{{n_t}}$, Pre-training epochs $T_{\text{pre}}$, Adaptation epochs $T_{\text{adapt}}$, Learning rate $\lambda$;}
   \ENSURE {Transformation $G(\cdot)$, Predictor $F(\cdot)$;}\\
   \STATE Initialize the network parameters $\Theta = \{\Theta_G,\Theta_F\}$; \\
   \STATE Initialize $\MBF{w}=\MBF{1}_c$; \\
   \% \textit{Pre-training Stage} \\
   \FOR {$t=1,2,\ldots,T_{\text{pre}}$}
   \STATE Forward propagate $\{\MBF{x}_i^s\}_{i=1}^{{n_s}}$ according to $Z=G(X)$ and $\hat{Y}=F(Z)$; \\
   \STATE compute the classification objective $\MC{J}_{\text{E}}$ in Eq.~\eqref{eq:objective_entropy_functions}; \\
   \STATE Update: $\Theta \leftarrow \Theta - \lambda \nabla\MC{J}^s_{\text{E}} (\Theta)$;\\
   \ENDFOR
   \\ \% \textit{Adaptation Stage} \\
   \FOR {$t=1,2,\ldots,T_{\text{adapt}}$}
   \STATE Forward propagate $\{\MBF{x}_i^s\}_{i=1}^{{n_s}}$ and $\{\MBF{x}_i^t\}_{i=1}^{{n_t}}$; compute the plug-in estimations $\hat{\MBF{p}}^s$, $\hat{\MBF{q}}^t$ and $\hat{\MBF{C}}$ in Eq.~\eqref{eq:BBSE_QP_estimation}; \\
   \STATE Estimate the important weight $\hat{\MBF{w}}$ via Eq.~\eqref{eq:BBSE_QP_estimation}; \\
   \STATE Estimate the prior probability $\hat{\MBF{p}}^t$ via $\hat{\MBF{p}}^t=\hat{\MBF{w}}\odot \hat{\MBF{p}}^s$; \\
   \STATE Compute the overall MUL objective $\MC{J}_{\text{MUL}}$ in Eq.~\eqref{eq:objective_MAL-final}; \\
   \STATE Update: $\Theta \leftarrow \Theta - \lambda \nabla\MC{J}_{\text{MUL}} (\Theta)$;\\
   \ENDFOR
   \end{algorithmic}
\end{algorithm}
}

\textbf{Algorithm. } As the neural networks are employed to learn transformation $G(\cdot)$ and predictor $F(\cdot)$, we denote the parameters refer to $G$ and $F$ as $\Theta_G$ and $\Theta_F$, respectively. The pseudo code of MUL is presented in Algorithm \ref{alg:MAL}. We first initialize $\Theta_G$ and $\Theta_F$ by pre-training them on the source domain with the basic classification objective $\MC{J}_{\text{E}}$. Since the parameters are trained with ground-truth labels, it only takes few epochs for pre-training. In the adaptation step, the decision uncertainty objective $\MC{J}_{\text{DU}}$ is firstly computed from the source data. When the objective value is stable, we further incorporate the target data with pseudo labels into $\MC{J}_{\text{DU}}$. To reduce the risk of pseudo labels, we only select the samples $(\MBF{z}^t_{i},\hat{\MBF{y}}^t_{i})$ whose confidences are larger than the threshold $\tau$, \ie, $\max_j~ \hat{\MBF{y}}^t_{ji}>\tau$. More details for implementation are provided in the supplementary file.

\section{Experiments}
MUL model is evaluated and then compared with the State-of-the-Art (SOTA) methods on four UDA benchmarks.

\begin{figure*}[t]
   \begin{minipage}{0.245\linewidth}
       \centering{\includegraphics[width=0.99\linewidth,height=87pt,trim=80 50 110 80,clip]{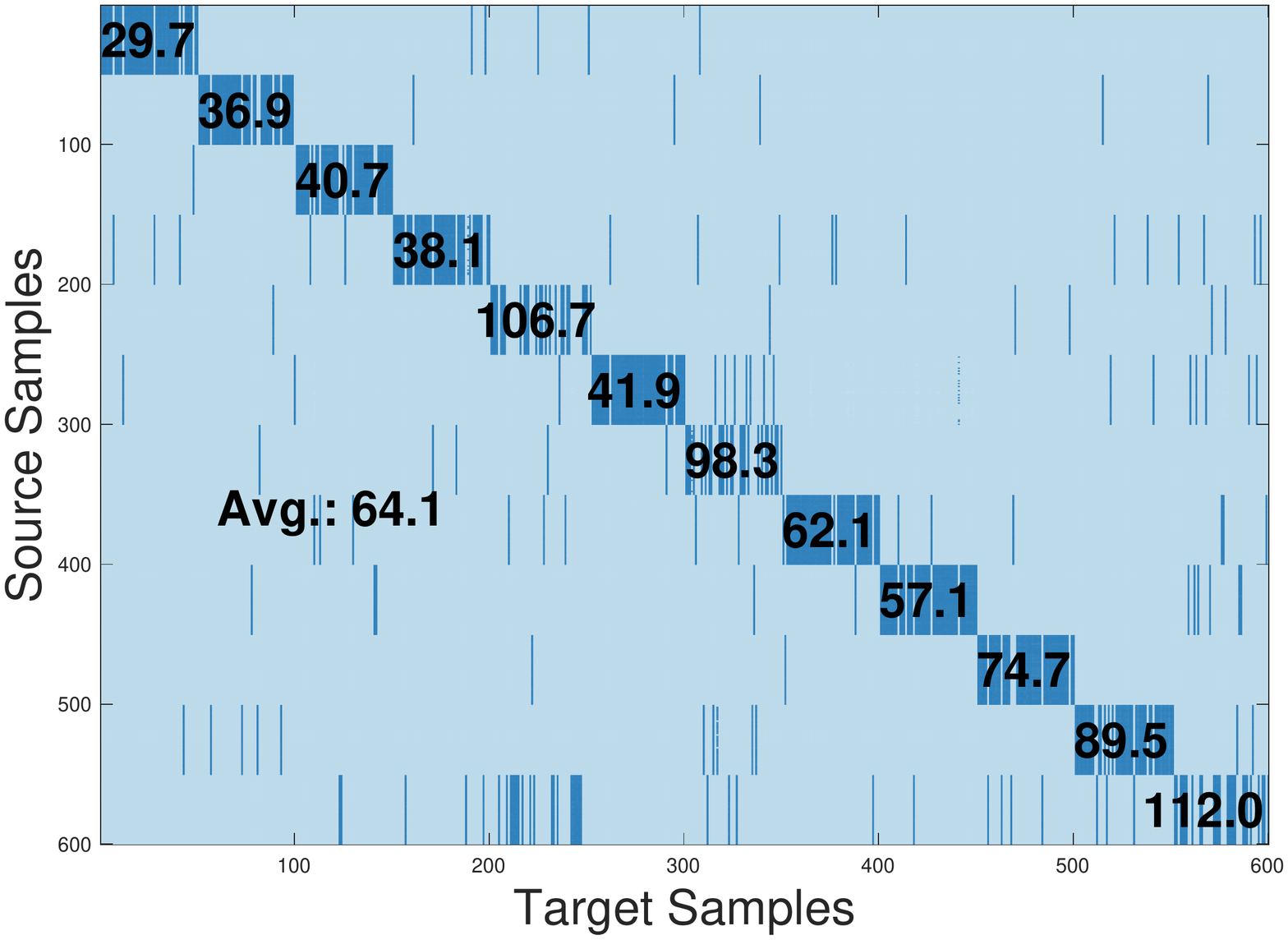}} \\
       (a) $\MBF{D}^{st}$ of DMP
   \end{minipage}
   \hfill
   \begin{minipage}{0.245\linewidth}
       \centering{\includegraphics[width=0.99\linewidth,height=87pt,trim=80 50 110 80,clip]{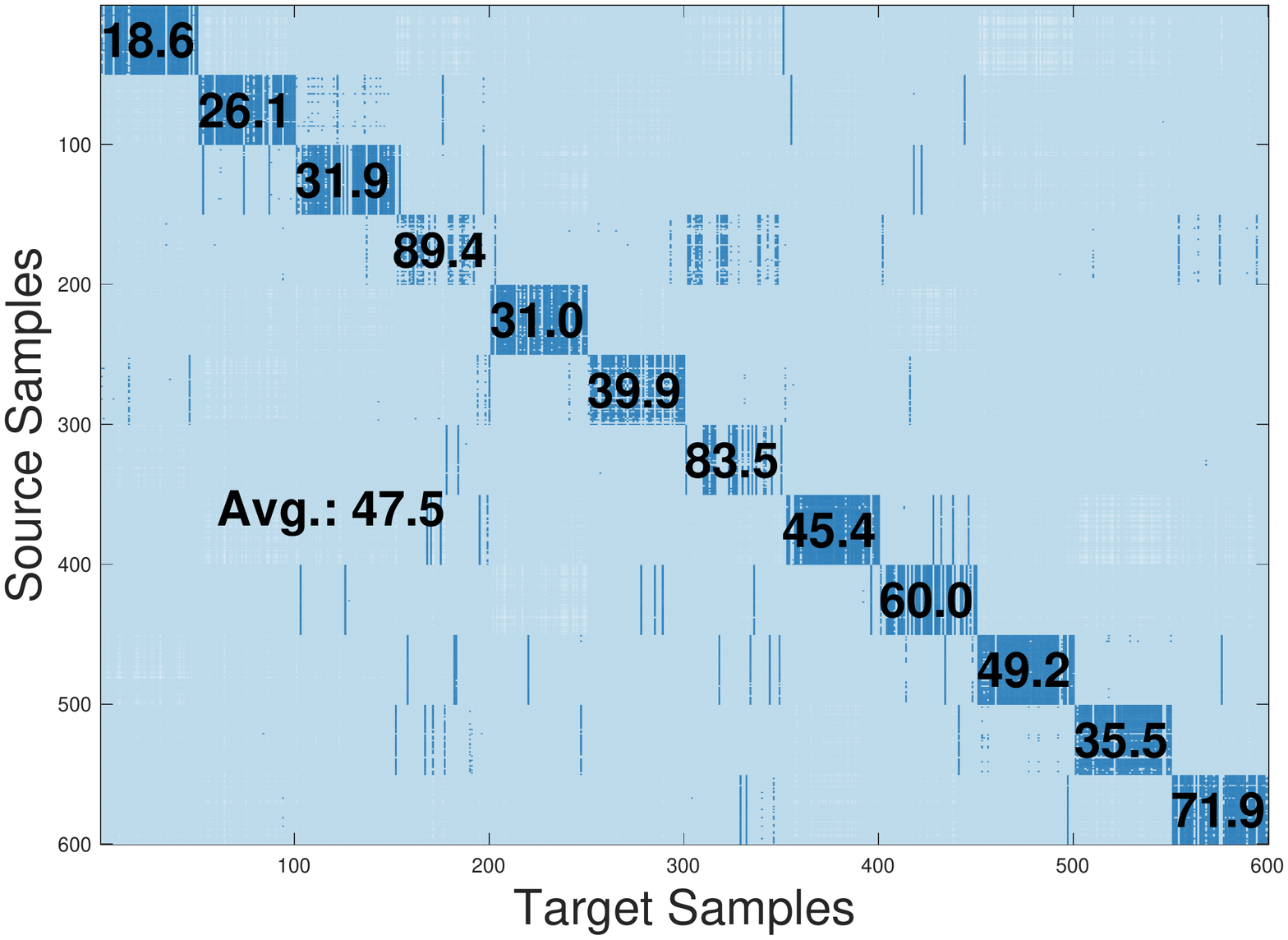}} \\
       (b) $\MBF{D}^{st}$ of MUL
   \end{minipage}
   \hfill
   \begin{minipage}{0.245\linewidth}
       \centering{\includegraphics[width=0.99\linewidth,height=87pt,trim=80 50 110 80,clip]{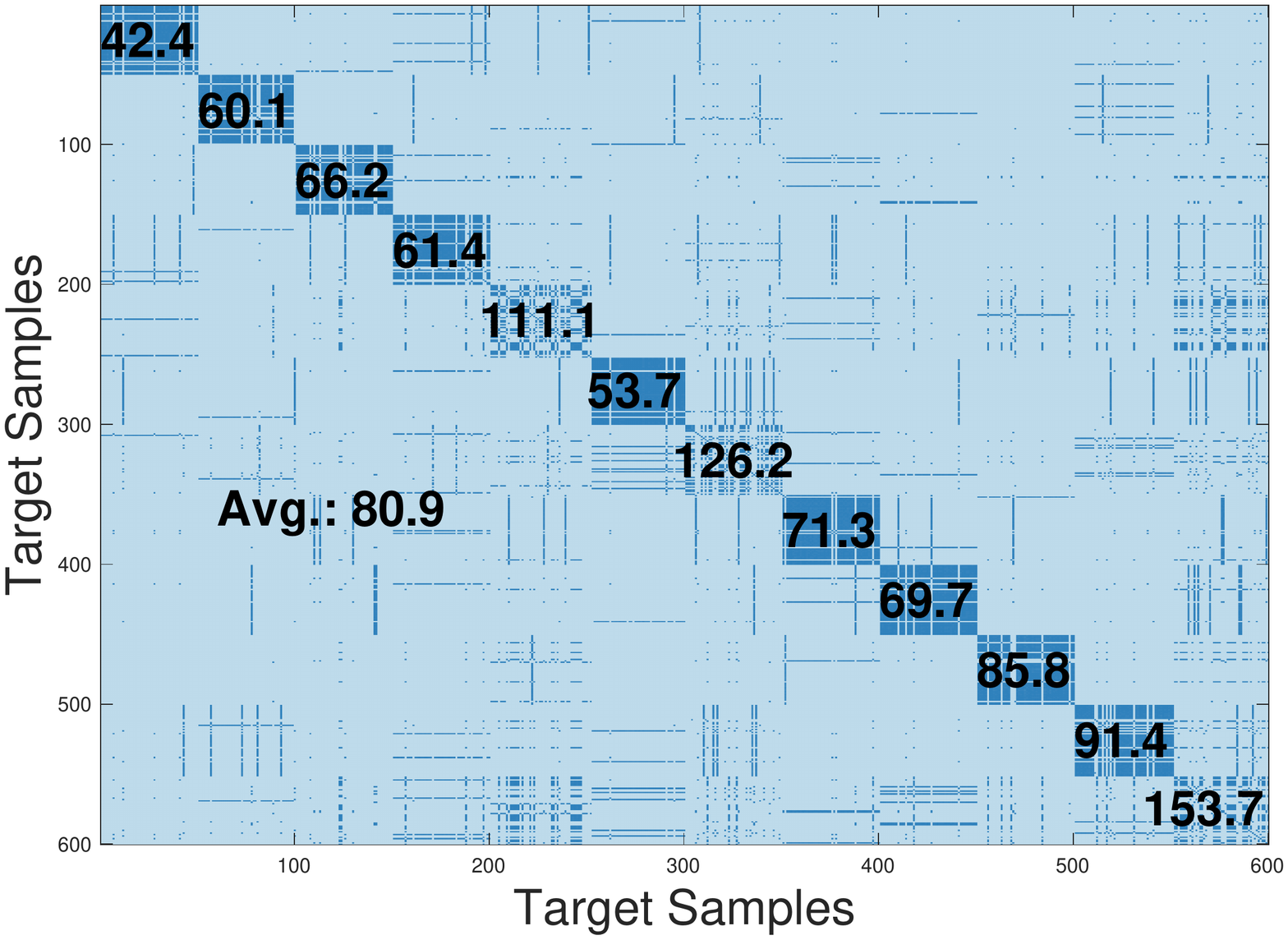}} \\
       (c) $\MBF{D}^{tt}$ of DMP
   \end{minipage}
   \hfill
   \begin{minipage}{0.245\linewidth}
       \centering{\includegraphics[width=0.99\linewidth,height=87pt,trim=80 50 110 80,clip]{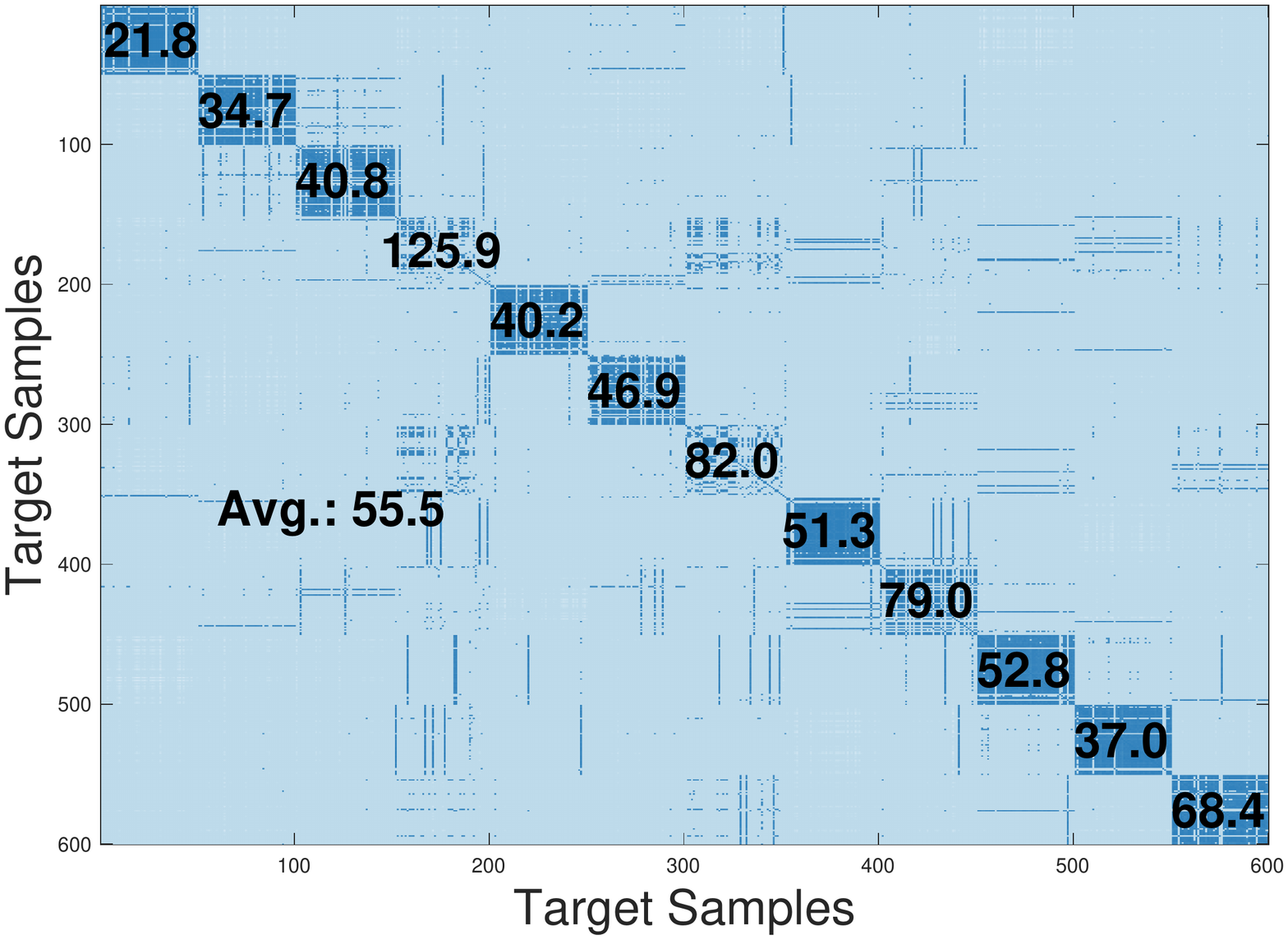}} \\
       (d) $\MBF{D}^{tt}$ of MUL
   \end{minipage}

   \begin{minipage}{0.245\linewidth}
       \centering{\includegraphics[width=0.96\linewidth,height=87pt,trim=151 75 155 105,clip]{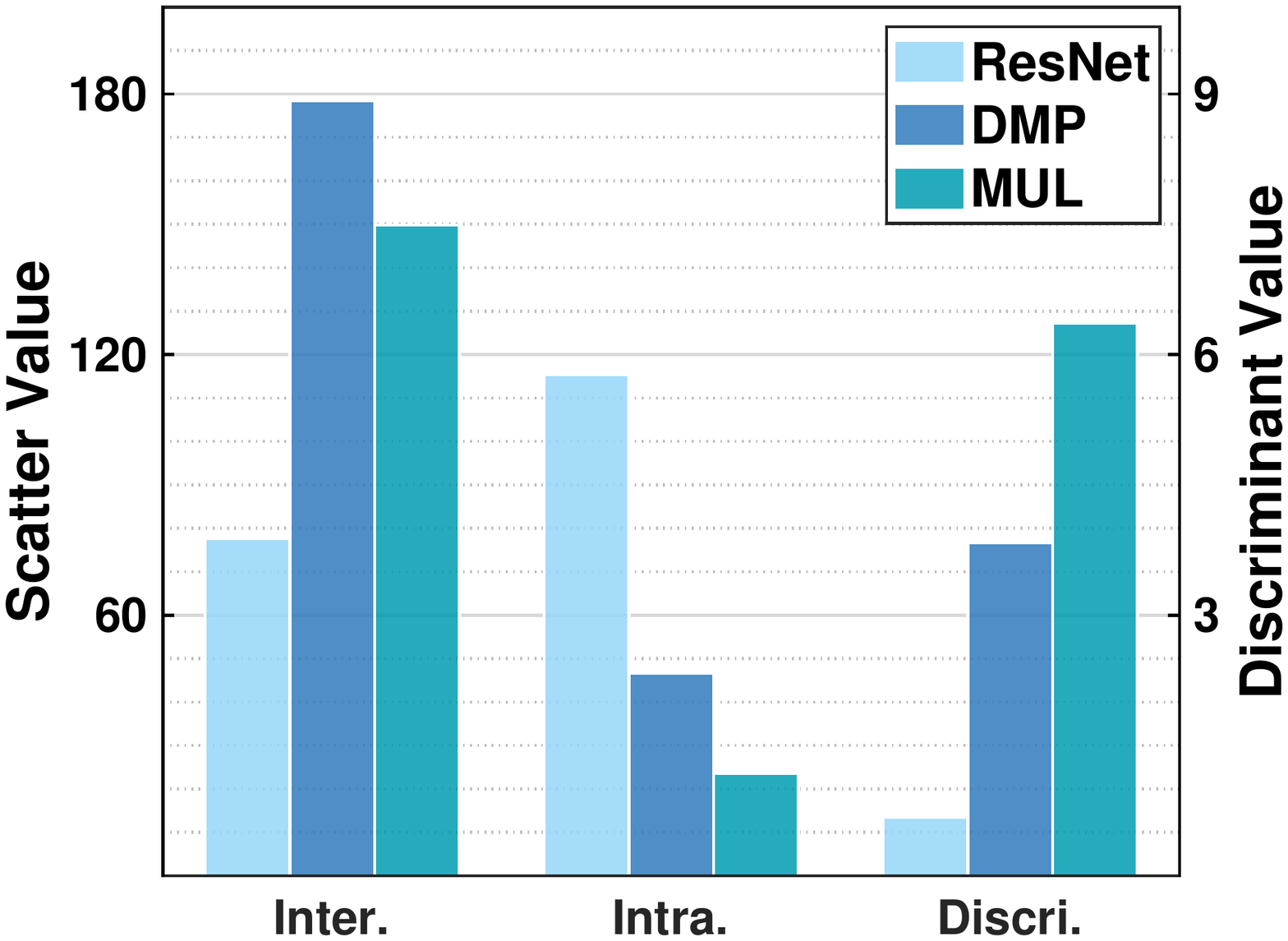}} \\
       (e) Discriminability on $\textbf{I}$
   \end{minipage}
   \hfill
   \begin{minipage}{0.245\linewidth}
      \centering{\includegraphics[width=0.99\linewidth,height=87pt,trim=160 100 190 105,clip]{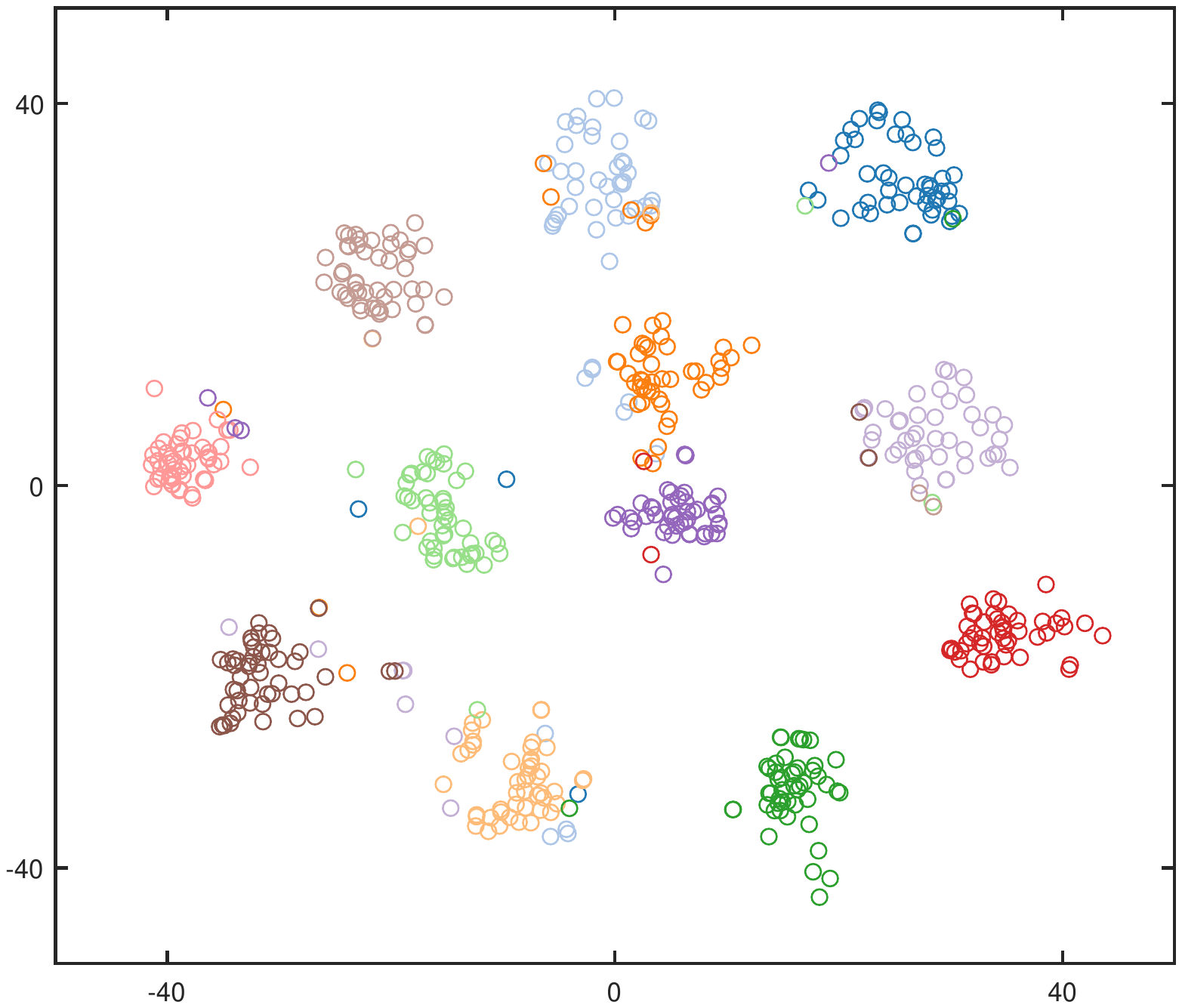}} \\
      (f) ResNet
   \end{minipage}
  \hfill
  \begin{minipage}{0.245\linewidth}
      \centering{\includegraphics[width=0.99\linewidth,height=87pt,trim=160 100 190 105,clip]{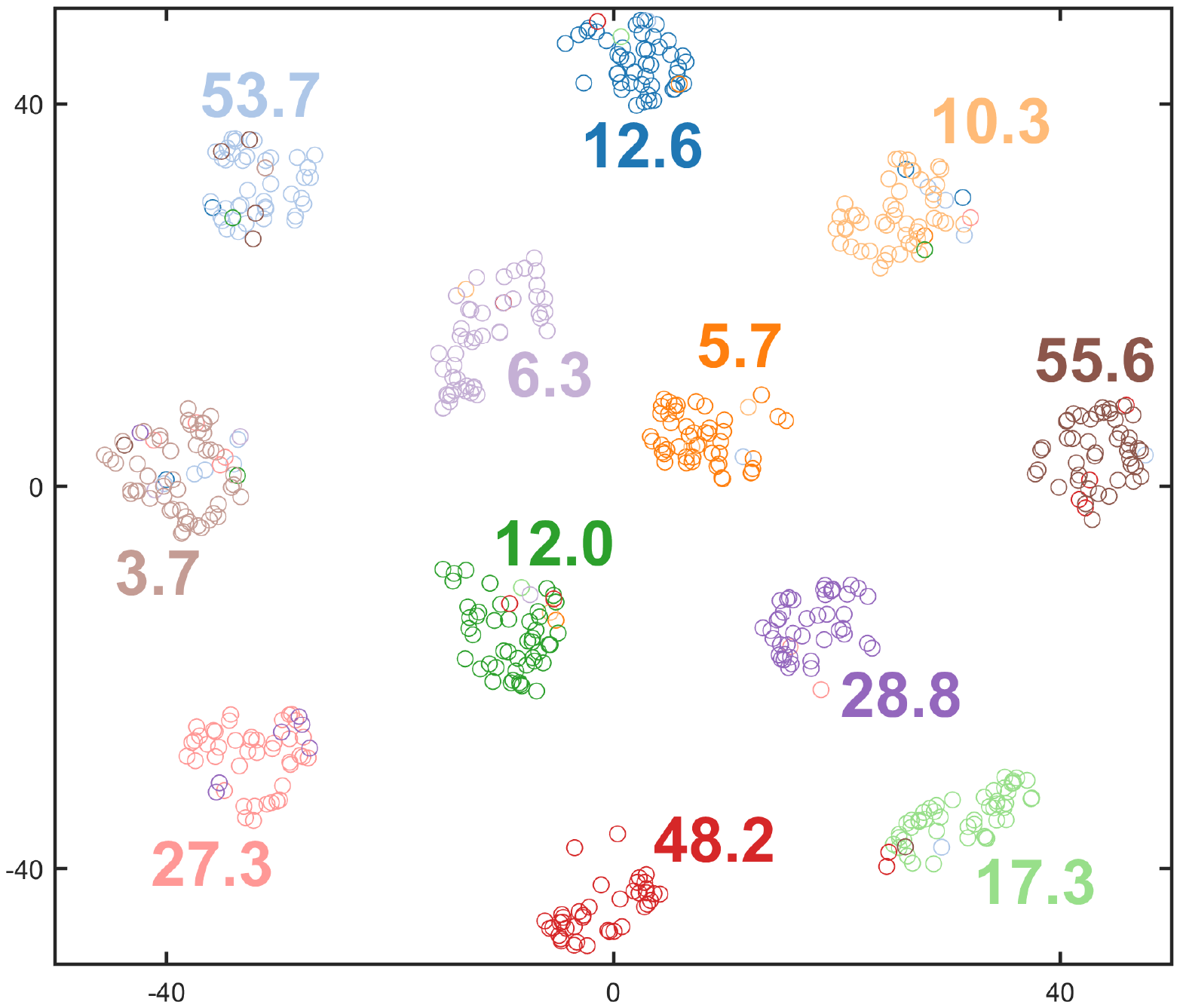}} \\
      (g) DMP
  \end{minipage}
  \hfill
  \begin{minipage}{0.245\linewidth}
      \centering{\includegraphics[width=0.99\linewidth,height=87pt,trim=160 100 190 105,clip]{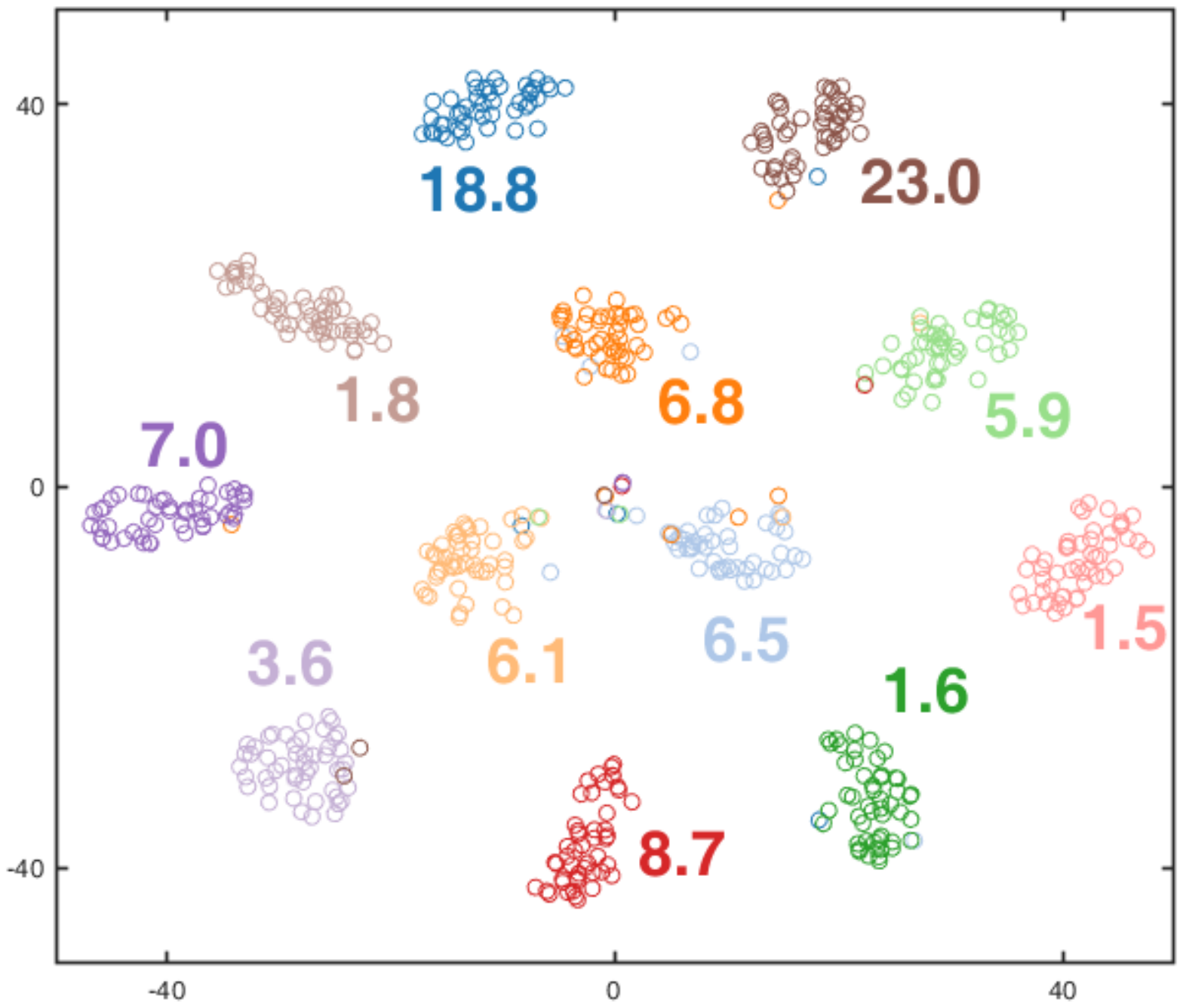}} \\
      (h) MUL
  \end{minipage}

   \caption{Model analysis on ImageCLEF under vanilla setting. (a)-(d): comparison of pairwise distances on \textbf{C}$\rightarrow$\textbf{P} task, where darker colors represent smaller distances. The mean distance of each class is presented on the diagonal block, and the mean distance of diagonal blocks is shown as \textbf{Avg.}. (e): quantitative evaluation of discriminability on target domain for \textbf{C}$\rightarrow$\textbf{I} task; (f)-(h): feature visualization of the target domain for \textbf{C}$\rightarrow$\textbf{I} task, where the values of intra-class scatters are provided beside the clusters. Best viewed in color.}
   \label{fig:Quantitative_Fig_UDA}
   \vskip -0.15in
\end{figure*}

\textbf{Office-Home} \cite{OfficeHome} contains 15500 images from 4 domains with 65 classes, \ie, \textit{Art} (\textbf{Ar}), \textit{Clipart} (\textbf{Cl}), \textit{Product} (\textbf{Pr}) and \textit{Real-World} (\textbf{Rw}). In PDA setting, the first 25 classes (alphabetical order) are selected as the target domain while the source domain consists of the vanilla data.

\textbf{VisDA-2017} \cite{VisDA-2017} is a large-scale visual UDA challenge. The source domain \textbf{S} contains 152397 synthetic images, and the target domain \textbf{R} contains 55388 real images from Microsoft COCO dataset. Following the challenge, we consider the task \textbf{S} $\rightarrow$ \textbf{R}. In PDA setting, the target domain \textbf{R6} contains samples from the first 6 classes (alphabetical order).

\textbf{Office-31} \cite{Office31} is an object recognition dataset. It contains 3 domains with 4110 images, \ie, \textit{Amazon} (\textbf{A}), \textit{Webcam} (\textbf{W}) and \textit{Dslr} (\textbf{D}). In PDA setting, we follow the common protocal \cite{cao2018partial,zhang2018importance} where the target consists of 10 classes.

\textbf{ImageCLEF}\footnote{\url{https://www.imageclef.org/2014/adaptation}} consists of 3 domains with 12 common classes, \ie, \textit{Caltech} (\textbf{C}), \textit{ImageNet} (\textbf{I}), \textit{Pascal} (\textbf{P}), where each domain include 600 images. In PDA setting, the target domain contains the first 6 classes (alphabetical order).

\textbf{Existence of GLS. }
Recall that GLS consists of the label shift ($P_Y$) and conditional shift
($P_{X|Y}$). Those two kinds of shift are common in real-world scenario, and the conditional shift usually appears in the computer vision. In the above four domain adaptation datasets, the images are collected in different environments, so GLS always exists. We quantify the label shift with the $\ell_1$ distance on probability distributions as in Table \ref{tab:L1_distance_label_shift}. Note that $0 \leq \|p_Y^s-p_Y^t\|_1 \leq 2$.

\begin{table}[t]
   \centering
   \vspace{-4pt}
   \setlength{\abovecaptionskip}{0.0cm}
   \setlength{\belowcaptionskip}{-0.01cm}
   \caption{Discrepancy between priors $p^s_Y$ and $p^t_Y$ measured by $\ell_1$ distance.}
   \label{tab:L1_distance_label_shift}
   \begin{small}
   \renewcommand{\tabcolsep}{0.18pc} 
   \renewcommand{\arraystretch}{1.0} 
   \begin{tabular}{c|cccc}
   \toprule[1pt]
   Settings & Office-Home & VisDA-2017 & Office-31 & ImageCLEF \\
   \hline
   UDA & 0.39 & 0.38 & 0.30 & 0.00 \\
   PDA & 1.18 & 1.06 & 1.32 & 1.00 \\
   \bottomrule[1pt]
   \end{tabular}
\end{small}
\vskip -0.2in
\end{table}

\vspace{-6pt}
\subsection{Vanilla Domain Adaptation}
In this section, we consider the vanilla UDA setting.

\textbf{Comparison. }
We compare MUL with several advanced UDA methods. The results on Office-Home and VisDA-2017 datasets are presented in the top Table \ref{tab:UDA_4dataset}. The classical DAN method exploits marginal metric MMD, which yields a larger error than the conditional adaptation. By mining the label information, many methods (\eg, CDAN, ETD and ATM) achieve the superior performance than the methods based on covariate shift assumption, and improve the accuracy by about 10-20\%. MUL characterizes the conditional relation between representations $Z$ and labels $Y$, which makes it more effective than previous methods. Thus, MUL achieves the highest accuracies on most adaptation tasks and improves the mean accuracy to 69.1\% and 82.8\% on Office-Home and VisDA-2017, respectively.

The results on Office-31 and ImageCLEF datasets are shown in the bottom of Table \ref{tab:UDA_4dataset}. MUL achieves the highest accuracy on Office-31 dataset. Note that on the challenging tasks \textbf{D}$\rightarrow$\textbf{A} and \textbf{W}$\rightarrow$\textbf{A}, MUL improves the accuracies by about $4\%$ significantly. The improvements over other discriminant learning model (\eg, ATM and DMP) also demonstrates that the global discriminative structure, which is not accessible in mini-batch training manner, is effectively explored by MUL. As the domain discrepancy on ImageCLEF dataset is relatively smaller, the lower decision uncertainty is highly expected to further boost the models' performance. MUL achieves a significant improvement on task \textbf{C}$\rightarrow$\textbf{P} and  increases the mean accuracy to $90.8\%$.

\begin{figure}[t]
   \begin{center}
   \includegraphics[width=0.75\linewidth]{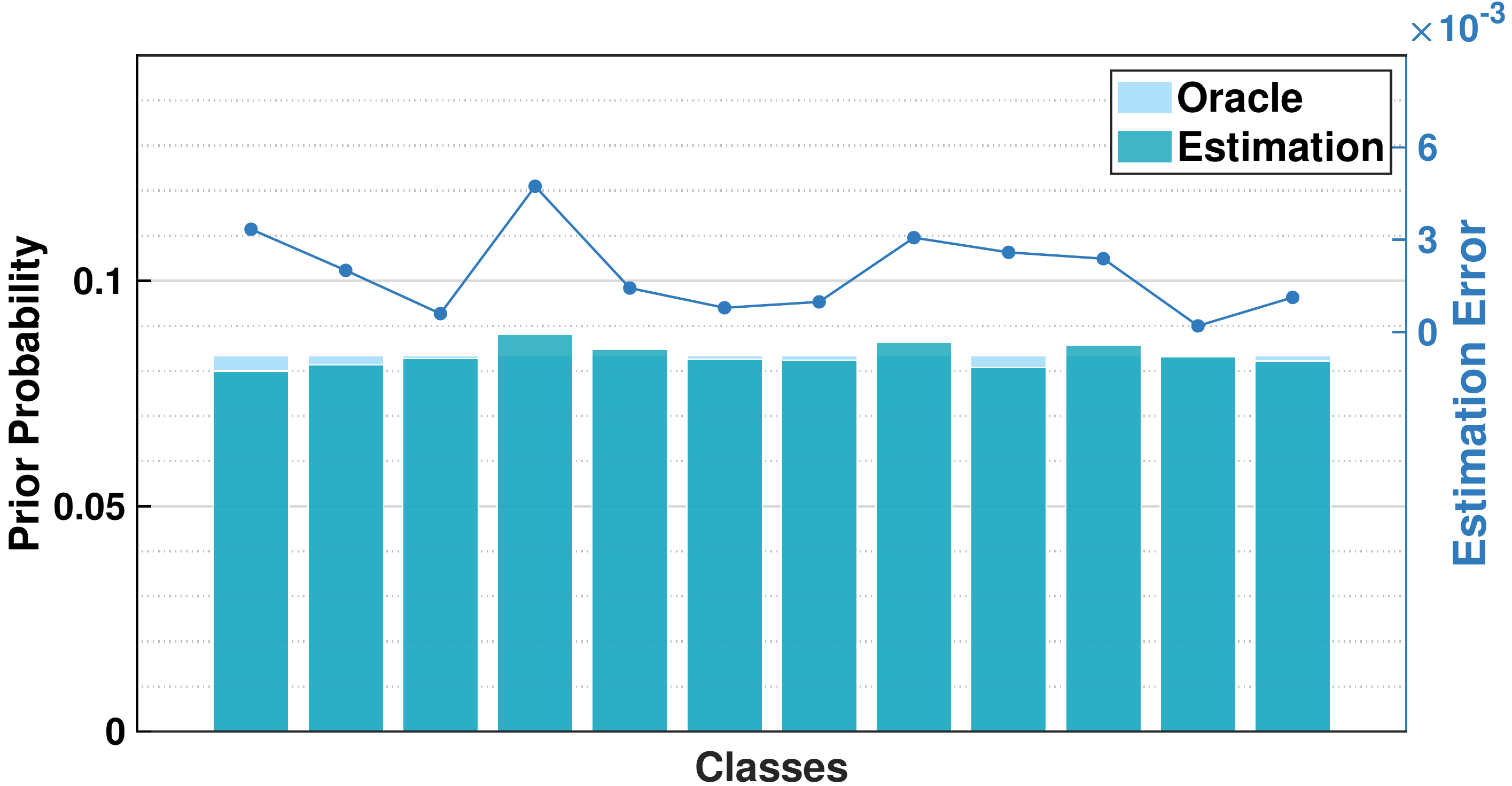}
   \caption{Prior probability  (``Oracle'') $p^t_Y$ and MUL's estimation $\hat{p}^t_Y$ on target domain \textbf{I} for ImageCLEF \textbf{C}$\rightarrow$\textbf{I} task, where the error curve is computed as $|p^t_y-\hat{p}^t_y|$, $\forall y\in \MC{Y}$. Best viewed in color.}
   \label{fig:IW_weight_Fig_UDA}
   \end{center}
   \vskip -0.25in
\end{figure}

\textbf{Transfer uncertainty. }
To evaluate the conditional invariant representations $Z$ learned from the MUL model, we compute the pairwise distance matrix $\left(\MBF{D}^{st} \right)_{ij} = \| \MBF{z}^s_i - \MBF{z}^t_j \|_2^2$ and visualize it in Figure \ref{fig:Quantitative_Fig_UDA} (a)-(d). Note that the samples are sorted by their classes, \ie, $\MBF{Z} = [\MBF{Z}_1,\MBF{Z}_2,\ldots,\MBF{Z}_c]$. For a fair comparison, the $\ell_2$-norms of representations are scaled to be the same for different models. In (a)-(d), the pairwise distance matrices of both DMP and MUL have smaller distances in diagonal blocks, which demonstrate that these methods actually ensure the discriminability and transferability during knowledge transfer. Note that the intra-class distances of MUL are significantly lower than those of DMP, and there are more dark elements in the non-diagonal region of DMP. Those results validate the conditional distribution adaptation via MCMD metric ensures the more preferable local structures and a lower uncertainty for both domains. Besides, we evaluate the estimated prior probability $\hat{p}^t_Y$ for the adjusted matching in Figure \ref{fig:IW_weight_Fig_UDA}. The estimation errors $|\hat{p}^t_Y-p^t_Y|$ of MUL are about $10^{-3}$. Since $p^t_Y=p^s_Y$ on vanilla ImageCLEF dataset, the estimated $\hat{p}^t_Y$ is approximately uniform (``Oracle''). This result demonstrates that MUL is also feasible for mild label shift.

\textbf{Decision uncertainty. }
To analyze the discriminability of the learned representations quantitatively, we employ the LDA-like functions as the evaluation metrics. Specifically, let $\bar{\MBF{z}}$ and $\bar{\MBF{z}}_i$ be the mean and class-wise means of representations. We compute the inter-class separability (\textbf{Inter.}) as $\MC{J}_b = \frac{1}{c} \sum_{i=1}^c \|\bar{\MBF{z}}_i -\bar{\MBF{z}} \|_2^2$, the intra-class scatter (\textbf{Intra.}) as $\MC{J}_w = \frac{1}{n} \sum_{i=1}^{n} \|\MBF{z}_i -\bar{\MBF{z}}_{y_i} \|_2^2$ and the discriminability (\textbf{Discri.}) as $\frac{\MC{J}_b}{\MC{J}_w}$. The results on the target domains are presented in Figure \ref{fig:Quantitative_Fig_UDA} (e). The results validate that the direct application of the model trained on the source domain (\ie, ResNet) cannot ensure the discriminability on the target domain, \ie, the lower \textbf{Discri.} value. Besides, the cross-entropy objective cannot explore the discriminative structure of domains sufficiently. We also observe that MUL with decision uncertainty minimization is effective in learning discriminative representations on both domain.

\begin{table*}[t]
   \centering
   \setlength{\abovecaptionskip}{0.0cm}
   \setlength{\belowcaptionskip}{-0.01cm}
   \caption{Classification accuracies (\%) on Office-Home, VisDA-2017, Office-31 and ImageCLEF datasets (ResNet-50) under partial setting.}
   \label{tab:4Datasets_partial}

   \renewcommand{\tabcolsep}{0.21pc} 
   \renewcommand{\arraystretch}{1.0} 
   \begin{tabular}{c|ccccccccccccc|c}
   \toprule[1pt]
   \multirow{2}{*}{\textbf{Methods}} & \multicolumn{13}{c|}{\textbf{Office-Home}} & \textbf{VisDA-2017} \\
     & Ar$\rightarrow$Cl & Ar$\rightarrow$Pr & Ar$\rightarrow$Rw & Cl$\rightarrow$Ar & Cl$\rightarrow$Pr & Cl$\rightarrow$Rw &
   Pr$\rightarrow$Ar & Pr$\rightarrow$Cl & Pr$\rightarrow$Rw & Rw$\rightarrow$Ar & Rw$\rightarrow$Cl & Rw$\rightarrow$Pr & Mean & S$\rightarrow$R6 \\
   \hline
   Source \cite{he2016deep} & 46.3 & 67.5 & 75.9 & 59.1 & 59.9 & 62.7 & 58.2 & 41.8 & 74.9 & 67.4 & 48.2 & 74.2 & 61.4 & 45.3 \\
   DANN \cite{ganin2015unsupervised} & 43.8 & 67.9 & 77.5 & 63.7 & 59.0 & 67.6 & 56.8 & 37.1 & 76.4 & 69.2 & 44.3 & 77.5 & 61.7 & 51.0 \\
   PADA \cite{cao2018partial}     & 52.0 & 67.0 & 78.7 & 52.2 & 53.8 & 59.0 & 52.6 & 43.2 & 78.8 & 73.7 & 56.6 & 77.1 & 62.1 & 53.5 \\
   ETN \cite{cao2019learning}  & \textbf{59.2} & 77.0 & 79.5 & 62.9 & 65.7 & 75.0 & 68.3 & 55.4 & 84.4 & 75.7 & 57.7 & 84.5 & 70.5 & - \\
   SAFN \cite{xu2019larger}    & 58.9 & 76.3 & 81.4 & 70.4 & 73.0 & 77.8 & 72.4 & 55.3 & 80.4 & 75.8 & 60.4 & 79.9 & 71.8 & 67.7 \\
   DRCN \cite{li2020deep} & 51.6 & 75.8 & 82.0 & 62.9 & 65.1 & 72.9 & 67.4 & 50.0 & 81.0 & 76.4 & 57.7 & 79.3 & 68.5 & 58.2 \\
   DMP \cite{luo2020unsupervised}         & 54.0 & 71.9 & 81.3  & 63.2 & 61.6 & 70.0 & 62.3 & 49.5 & 77.2 & 73.4 & 54.1 & 79.4 & 66.5 & 67.6 \\
   DMP+ent \cite{luo2020unsupervised}       & 59.0 & 81.2 & 86.3 & 68.1 & 72.8 & 78.8 & 71.2 & 57.6 & 84.9 & 77.3 & \textbf{61.5} & 82.9 & 73.5 & 72.7 \\
   \hline
   MUL & 57.4 & \textbf{88.7} & \textbf{90.8} & \textbf{71.0} & \textbf{80.4} & \textbf{82.1} & \textbf{77.9} & \textbf{59.8} & \textbf{91.2} & \textbf{83.5} & 58.1 & \textbf{87.7} & \textbf{77.4} & \textbf{77.5}\\
   \bottomrule[1pt]
   \end{tabular}
   \\[2pt]
   \renewcommand{\tabcolsep}{0.323pc} 
   \renewcommand{\arraystretch}{1.0} 
   \begin{tabular}{c|ccccccc|ccccccc}
   \toprule[1pt]
   \multirow{2}{*}{\textbf{Methods}} & \multicolumn{7}{c|}{\textbf{Office-31}} & \multicolumn{7}{c}{\textbf{ImageCLEF}} \\
    & A$\rightarrow$W & D$\rightarrow$W & W$\rightarrow$D & A$\rightarrow$D & D$\rightarrow$A & W$\rightarrow$A & Mean & I$\rightarrow$P & P$\rightarrow$I & I$\rightarrow$C & C$\rightarrow$I & C$\rightarrow$P & P$\rightarrow$C & Mean \\
   \hline
   Source \cite{he2016deep} & $75.6^{1.1}$ & $96.3^{0.9}$ & $98.1^{0.7}$ & $83.4^{1.1}$ & $83.9^{1.0}$ & $85.0^{0.9}$ & 87.1 & $78.3^{0.2}$ & $86.9^{0.2}$ & $91.0^{0.2}$ & $84.3^{0.4}$ & $72.5^{0.4}$ & $91.5^{0.3}$ & 84.1 \\
   DANN \cite{ganin2015unsupervised} & $73.6^{0.2}$ & $96.3^{0.3}$ & $98.7^{0.2}$ & $81.5^{0.2}$ & $82.8^{0.2}$ & $86.1^{0.2}$ & 86.5 & $78.1^{0.2}$ & $86.3^{0.2}$ & $91.3^{0.4}$ & $84.0^{0.3}$ & $72.1^{0.3}$ & $90.3^{0.2}$ & 83.7 \\
   PADA \cite{cao2018partial}     & $86.5^{0.3}$ & $99.3^{0.5}$ & $\textbf{100.0}^{0.0}$ & $82.2^{0.4}$ & $92.7^{0.3}$ & $95.4^{0.3}$ & 92.7 & $81.7^{0.2}$ & $92.1^{0.2}$ & $94.6^{0.2}$ & $89.8^{0.2}$ & $77.7^{0.3}$ & $94.1^{0.1}$ & 88.3 \\
   SAFN \cite{xu2019larger}    & $87.5^{0.7}$ & $96.6^{0.2}$ & $99.4^{0.7}$ & $89.8^{1.5}$ & $92.6^{0.2}$ & $92.7^{0.1}$ & 93.1 & $79.5^{0.2}$ & $90.7^{0.2}$ & $93.0^{0.1}$ & $90.3^{0.1}$ & $77.8^{0.2}$ & $94.0^{0.2}$ & 87.5 \\
   DRCN \cite{li2020deep} & $90.8{~}$ & $ \textbf{100.0}{~}$ & $ \textbf{100.0}{~}$ & $94.3{~}$ & $95.2{~}$ & $94.8{~}$ & 95.9 & - & - & - & - & - & - & -  \\
   DMP  \cite{luo2020unsupervised}  & $94.5^{0.5}$ & $99.9^{0.1}$ & $\textbf{100.0}^{0.0}$ & $95.0^{1.0}$ & $94.7^{0.3}$ & $95.4^{0.3}$ & 96.6 & $81.5^{0.2}$ & $94.3^{0.1}$ & $96.2^{0.1}$ & $93.0^{0.3}$ & $78.2^{0.2}$ & $96.5^{0.1}$ & 90.0 \\
   DMP+ent  \cite{luo2020unsupervised}  & $\textbf{96.6}^{0.9}$ & $\textbf{100.0}^{0.0}$ & $\textbf{100.0}^{0.0}$ & $96.4^{0.9}$ & $95.1^{0.2}$ & $95.4^{0.1}$ & 97.2 & $82.4^{0.2}$ & $\textbf{94.5}^{0.3}$ & $96.7^{0.2}$ & $94.3^{0.1}$ & $ 78.7^{0.8}$ & $96.4^{0.2}$ & 90.5 \\
   \hline
   MUL & $94.2^{1.1}$ & $\textbf{100.0}^{0.0}$ & $\textbf{100.0}^{0.0}$ & $\textbf{98.5}^{0.3}$ & $\textbf{95.6}^{0.1}$ & $\textbf{96.3}^{0.1}$ & \textbf{97.5} & $\textbf{87.5}^{0.2}$ & $92.2^{0.2}$ & $\textbf{98.1}^{0.1}$ & $\textbf{94.6}^{0.3}$ & $\textbf{87.6}^{0.1}$ & $\textbf{98.5}^{0.1}$ & \textbf{93.1}
   \\
   \bottomrule[1pt]
   \end{tabular}
   \vskip -0.10in
\end{table*}

\begin{figure*}[t]
   \begin{minipage}{0.245\linewidth}
       \centering{\includegraphics[width=0.99\linewidth,height=87pt,trim=80 50 110 80,clip]{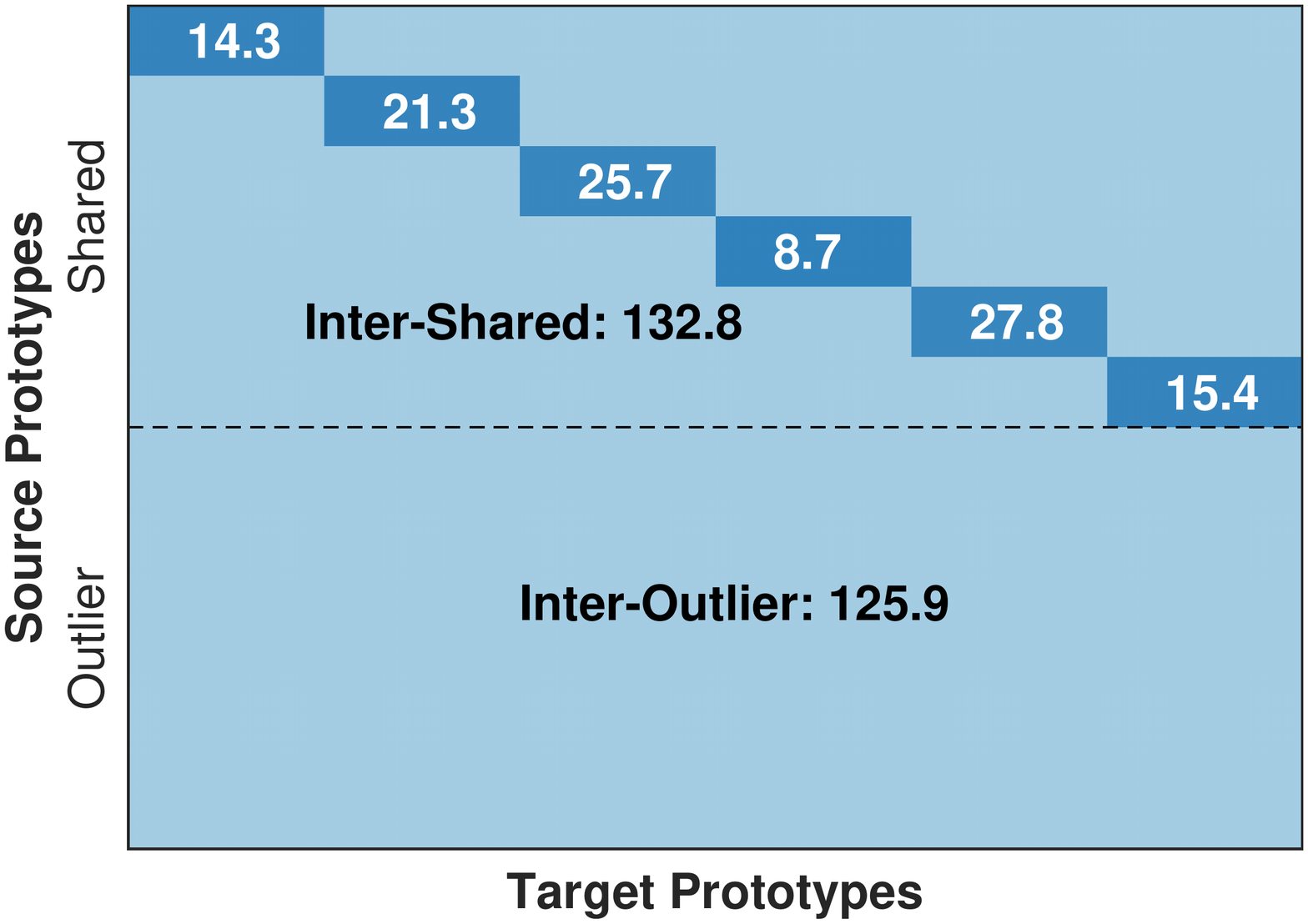}} \\
       (a) $\MBF{D}^{st}$ of ResNet
   \end{minipage}
   \hfill
   \begin{minipage}{0.245\linewidth}
       \centering{\includegraphics[width=0.99\linewidth,height=87pt,trim=80 50 110 80,clip]{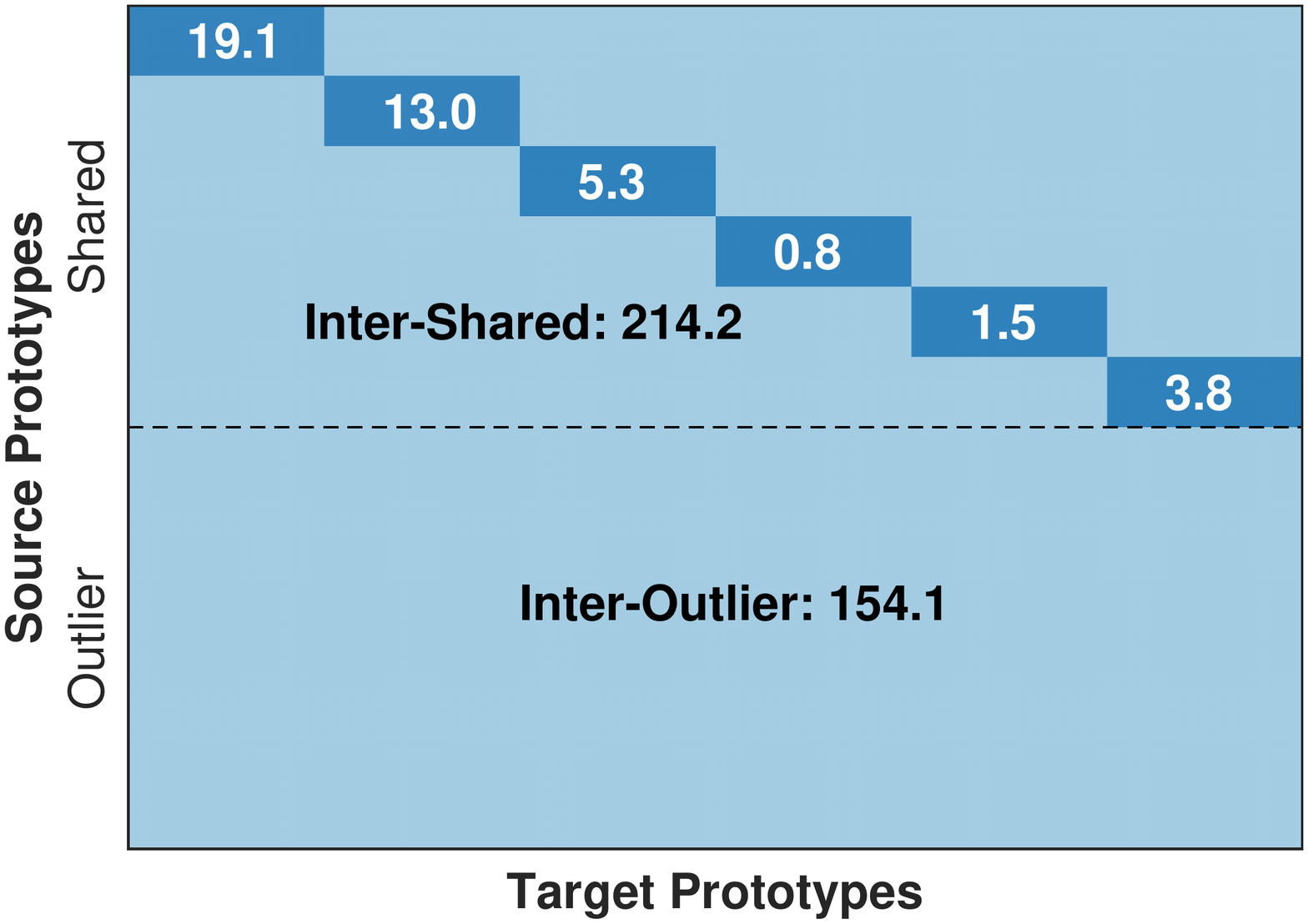}} \\
       (b) $\MBF{D}^{st}$ of DMP
   \end{minipage}
   \hfill
   \begin{minipage}{0.245\linewidth}
       \centering{\includegraphics[width=0.99\linewidth,height=87pt,trim=80 50 110 80,clip]{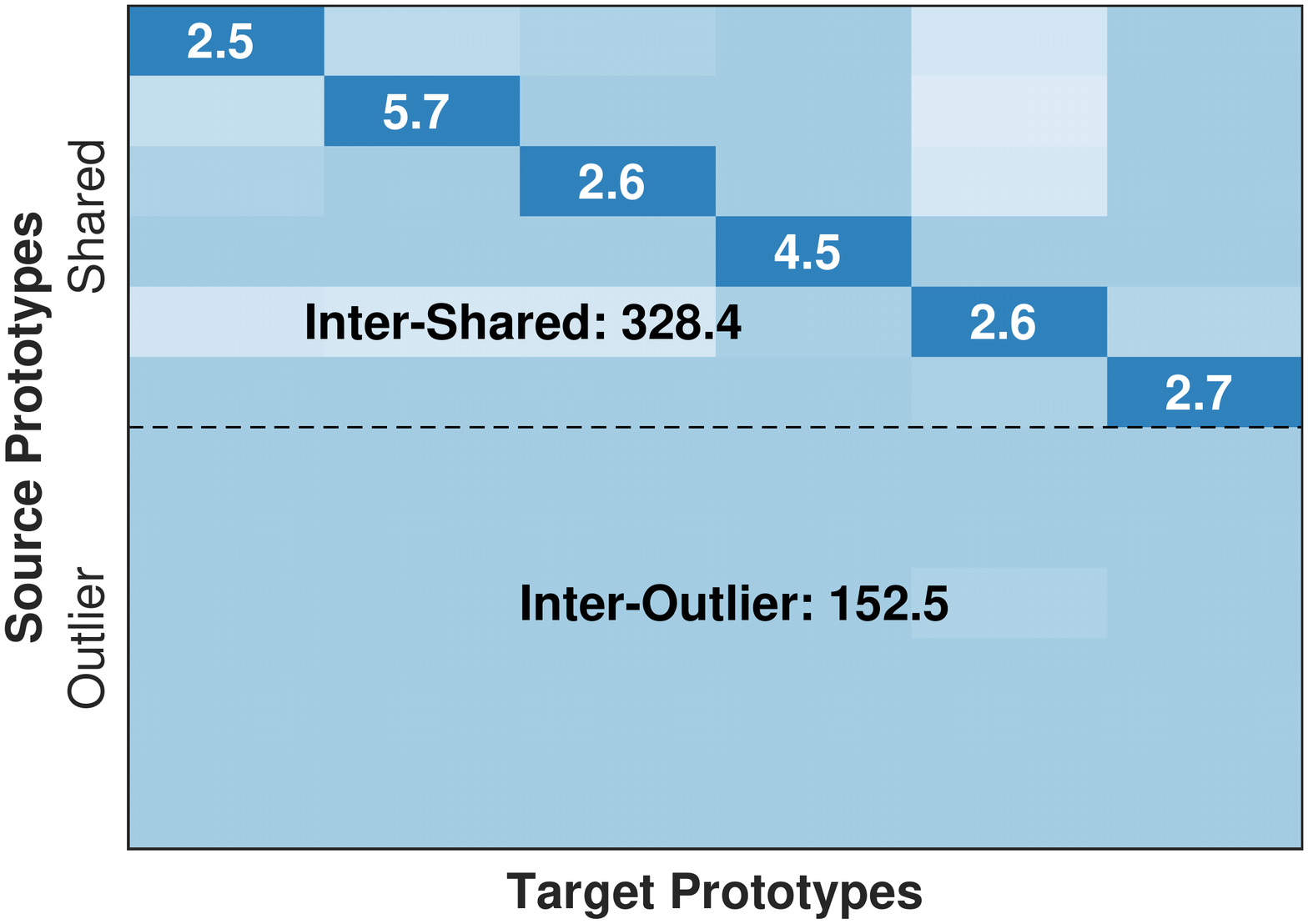}} \\
       (c) $\MBF{D}^{st}$ of MUL
   \end{minipage}
   \hfill
   \begin{minipage}{0.245\linewidth}
       \centering{\includegraphics[width=0.99\linewidth,height=87pt,trim=151 80 145 105,clip]{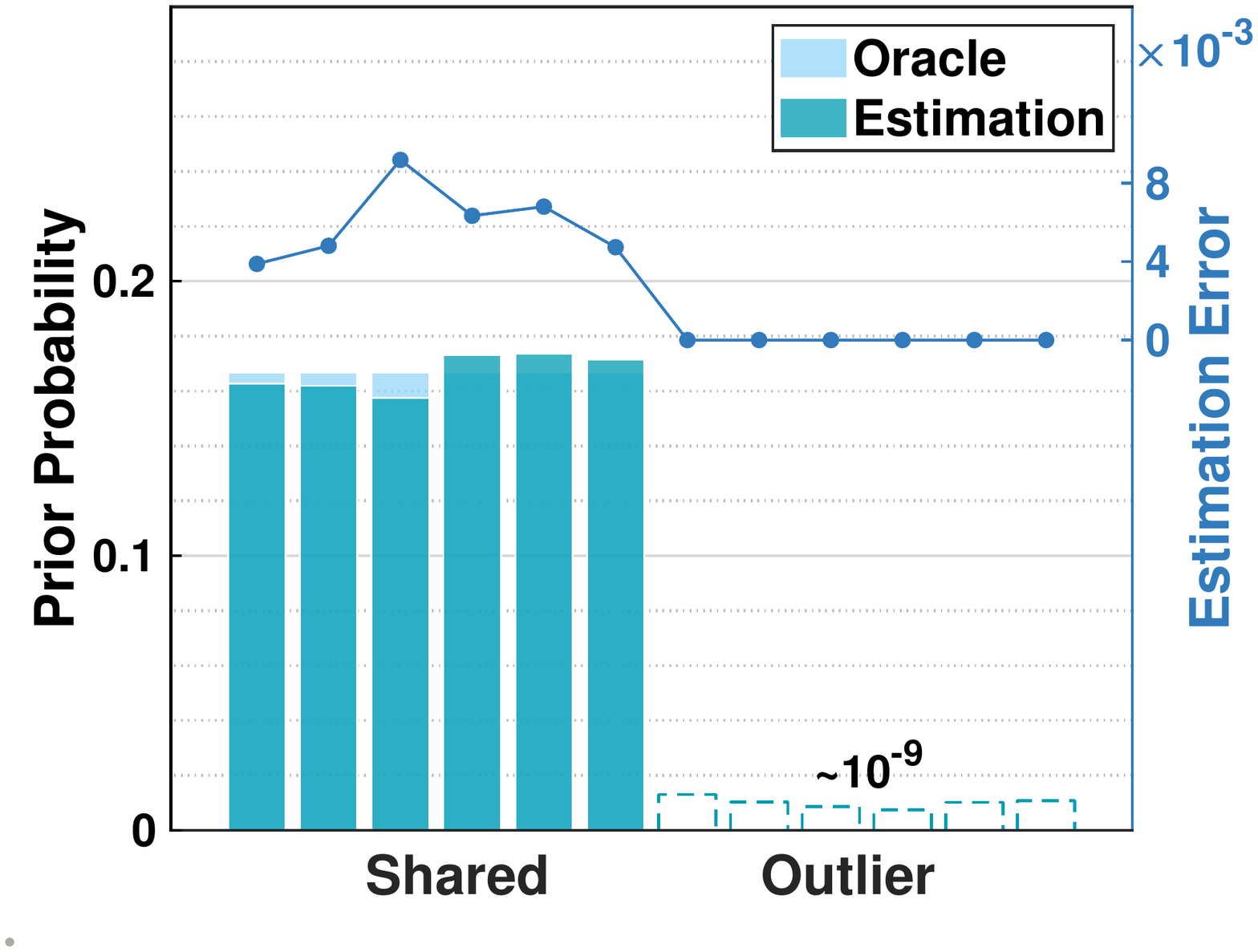}} \\
       (d) Weight Estimation on $\textbf{I}$
   \end{minipage}
   \begin{minipage}{0.245\linewidth}
      \centering{\includegraphics[width=0.99\linewidth,height=87pt,trim=160 100 190 105,clip]{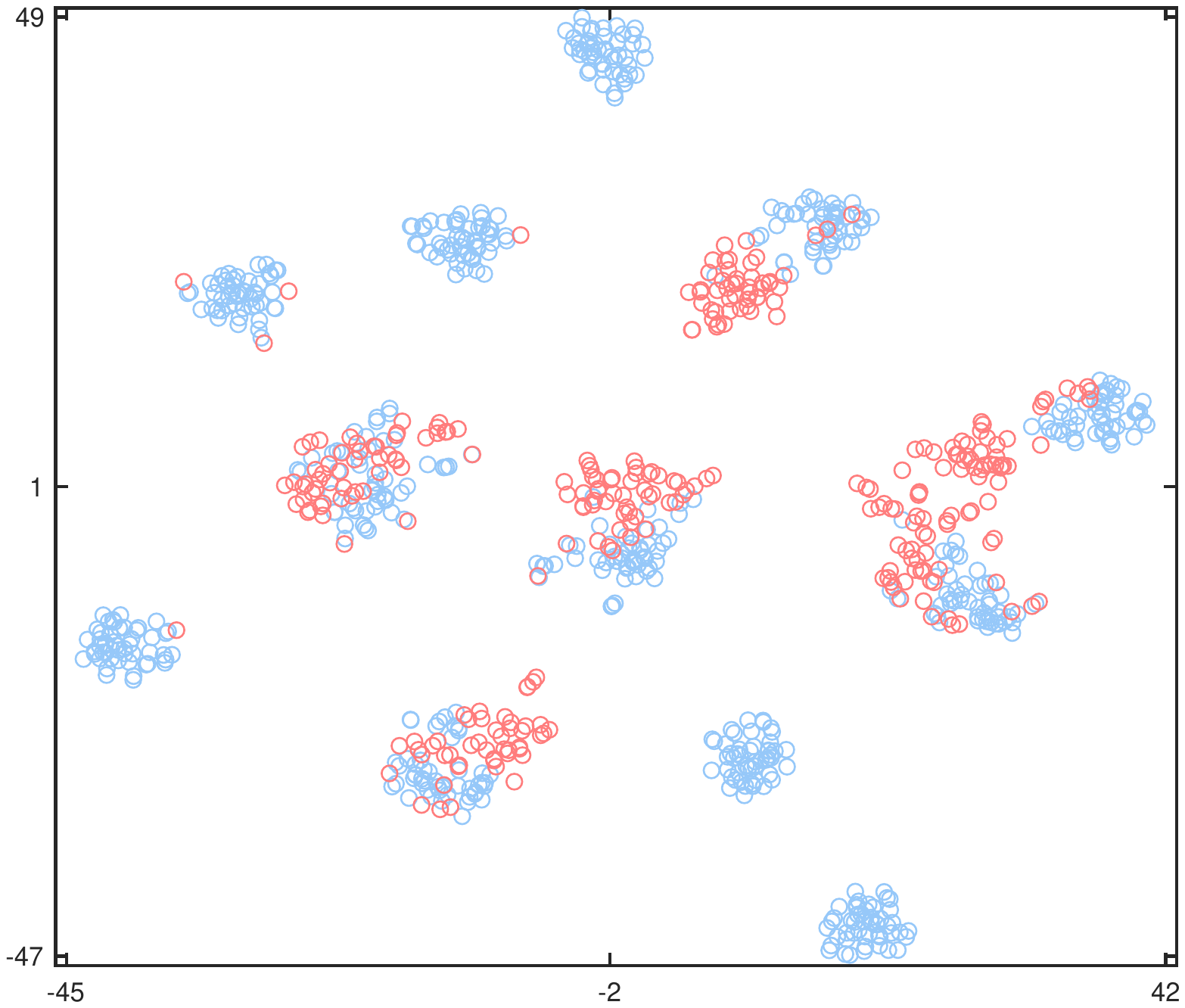}} \\
      (e) ResNet
  \end{minipage}
  \hfill
  \begin{minipage}{0.245\linewidth}
      \centering{\includegraphics[width=0.99\linewidth,height=87pt,trim=160 100 190 105,clip]{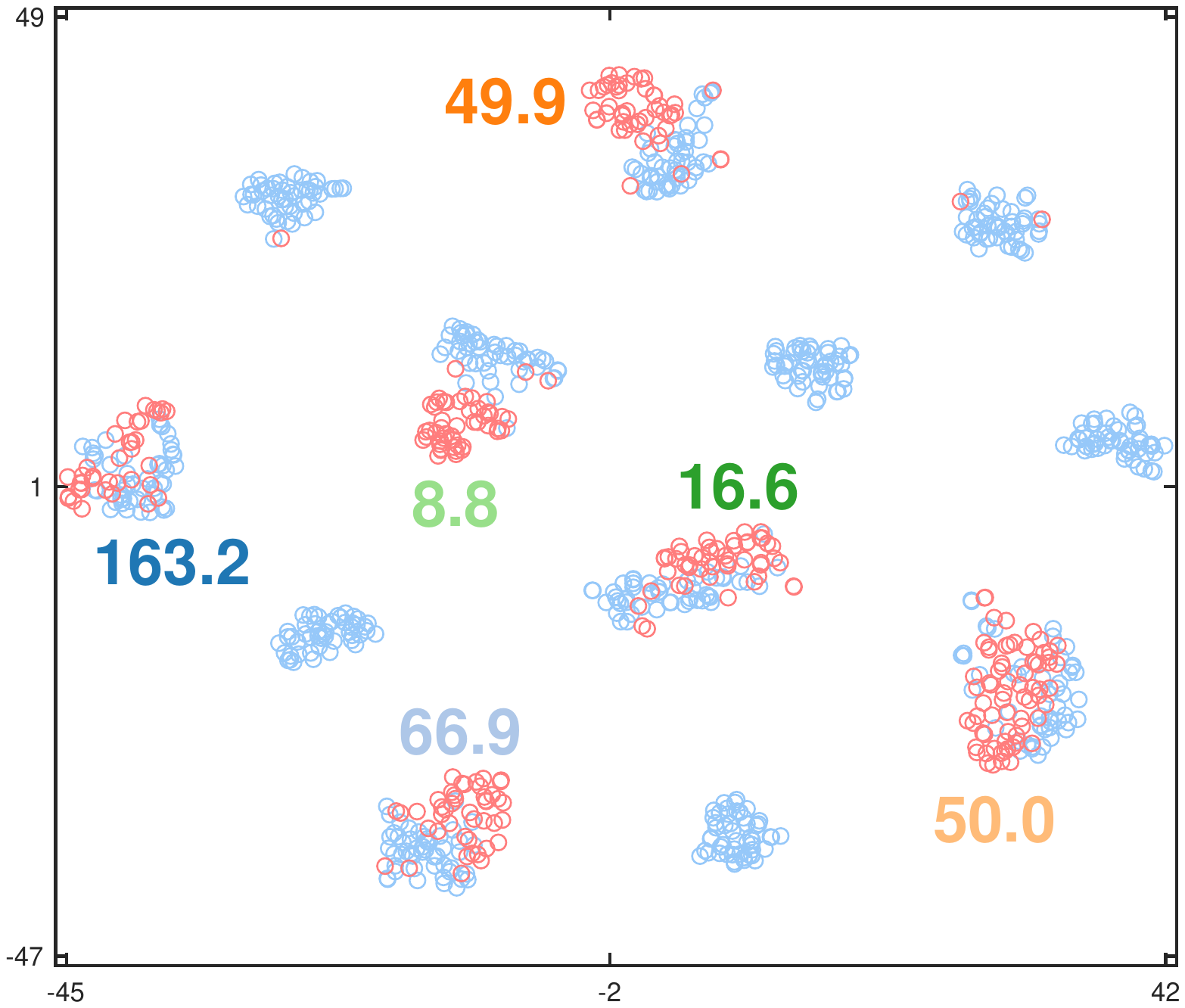}} \\
      (f) DMP
  \end{minipage}
  \hfill
  \begin{minipage}{0.245\linewidth}
      \centering{\includegraphics[width=0.99\linewidth,height=87pt,trim=160 100 190 105,clip]{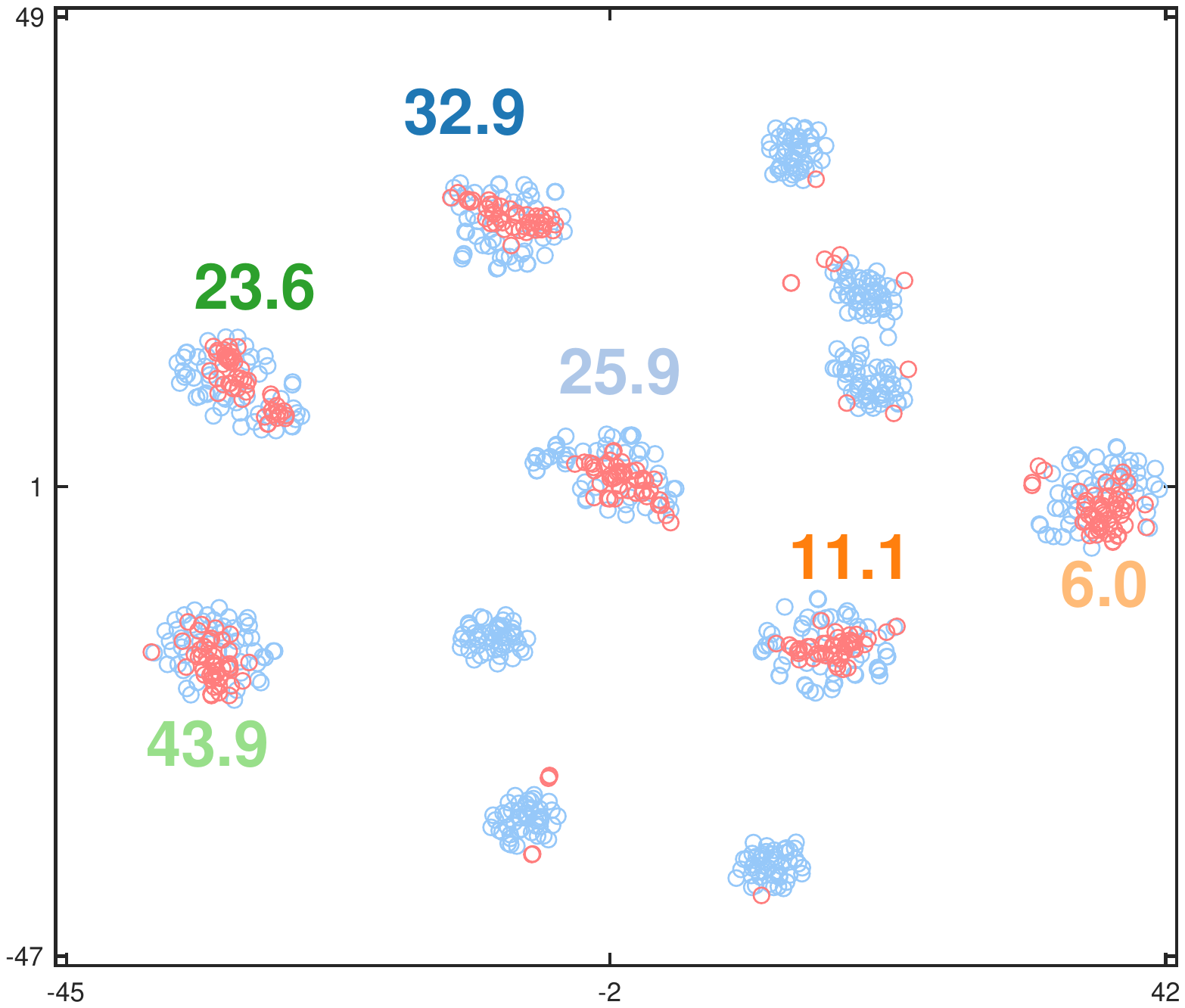}} \\
      (g) MUL
  \end{minipage}
  \hfill
  \begin{minipage}{0.245\linewidth}
      \centering{\includegraphics[width=0.99\linewidth,height=87pt,trim=151 80 145 105,clip]{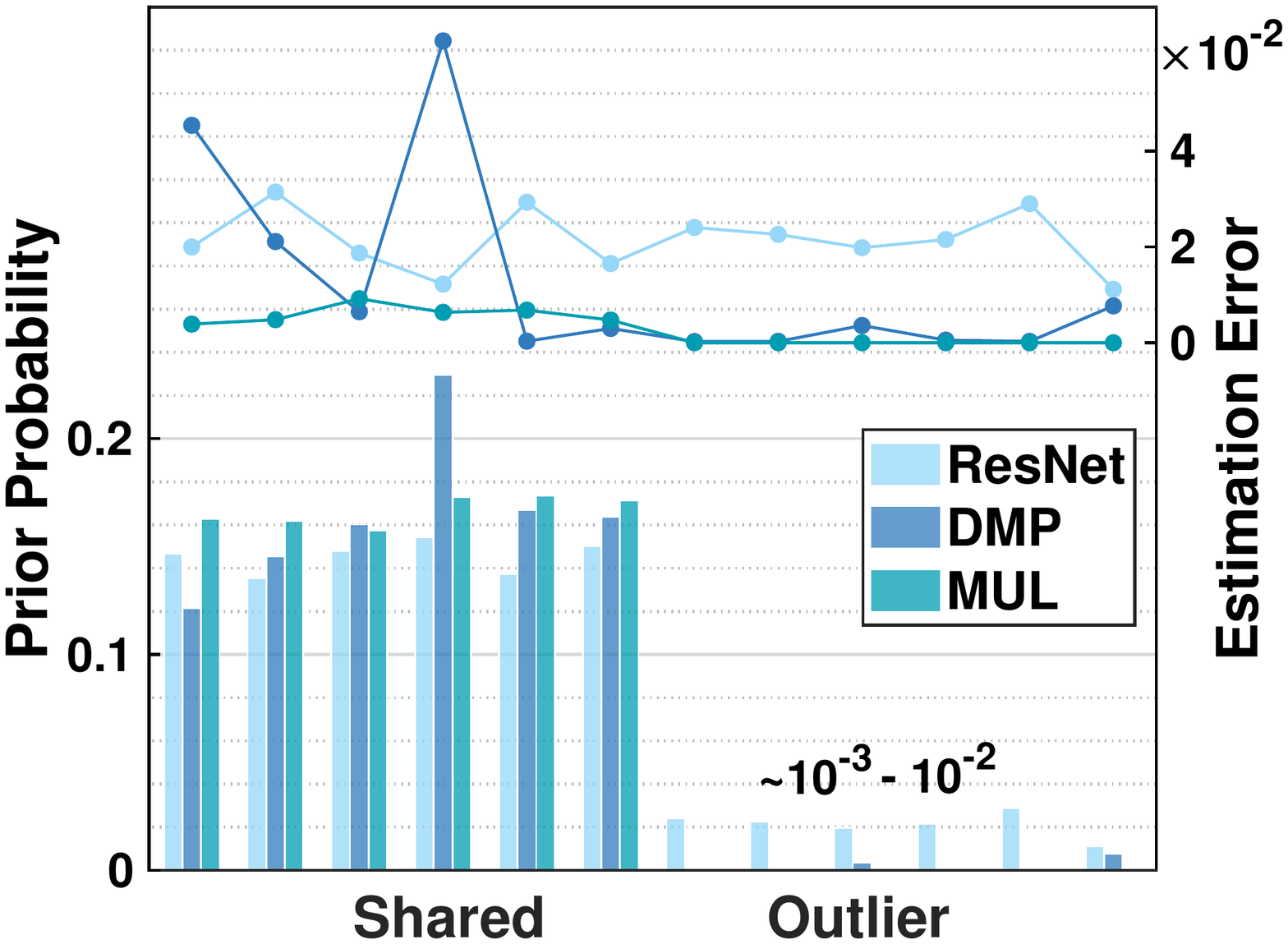}} \\
      (h) Weight Comparison on $\textbf{I}$
  \end{minipage}

   \caption{Model analysis on ImageCLEF \textbf{C}$\rightarrow$\textbf{I} task under partial setting. (a)-(c): pairwise distance matrices between class prototypes, where darker colors represent smaller distances and values are distances; (e)-(g): feature visualization of the source ``\textcolor[RGB]{150,200,249}{$\bigcirc$}'' and target ``\textcolor[RGB]{255,125,125}{$\bigcirc$}'' domains with different methods, where the values are intra-class scatters; (d) and (h): the estimated prior probabilities of different methods. Best viewed in color.}
   \label{fig:Quantitative_Fig_PDA}
   \vskip -0.15in
\end{figure*}

\textbf{Feature Visualization. }
We visualize the target representations learned from different models via t-SNE \cite{maaten2008visualizing} in Figure \ref{fig:Quantitative_Fig_UDA} (f)-(h). We also provide the intra-class scatter values of the 2-D t-SNE data in figures. The representations without adaptation in (f) induce the negative transfer problem, where some samples are misaligned. Besides, some clusters are close to each others which will increase the empirical risk of the classifier, \eg, the orange and purple '$\circ$'. With the manifold embedding, DMP learns the inter-class separable representations in (g). As MUL learns the global discriminative structure in (h), it further improves the intra-class compactness while alleviating the misalignment problem. Therefore, the intra-class scatter values of MUL is significant smaller than DMP, which are consistent with the result on Figure \ref{fig:Quantitative_Fig_UDA} (e). The results validate that MUL, which characterizes conditional discrepancy directly, achieves higher transferability and discriminability than other methods.

\begin{table*}
   \centering
   \setlength{\abovecaptionskip}{0.0cm}
   \setlength{\belowcaptionskip}{-0.01cm}
   \caption{Ablation study on ImageCLEF, Office-Home and VisDA-2017 datasets.}
   \label{tab:Ablation}
    \renewcommand{\tabcolsep}{0.33pc}
    \renewcommand{\arraystretch}{1.0} 
    \begin{tabular}{c|ccc|cccccc|cccccc|cc}
        \toprule[1pt]
        \multirow{2}{*}{\textbf{Metric}} & \multicolumn{3}{c|}{\multirow{2}{*}{\textbf{Objectives}}} & \multicolumn{6}{c|}{\textbf{ImageCLEF}} & \multicolumn{6}{c|}{\textbf{Office-Home}} & \multicolumn{2}{c}{\textbf{VisDA-2017}} \\
        & \multicolumn{3}{c|}{} & \multicolumn{3}{c}{\textbf{UDA}} & \multicolumn{3}{c|}{\textbf{PDA}} & \multicolumn{3}{c}{\textbf{UDA}} & \multicolumn{3}{c|}{\textbf{PDA}} & \textbf{UDA} & \textbf{PDA} \\
        $D(\cdot \| \cdot)$ & $\MBF{w,y}$  & $\MC{J}_{\text{TU}}$ & $\MC{J}_{\text{DU}}$ &I$\rightarrow$P & P$\rightarrow$I & I$\rightarrow$C & I$\rightarrow$P & P$\rightarrow$I & I$\rightarrow$C & Cl$\rightarrow$Ar & Cl$\rightarrow$Pr & Cl$\rightarrow$Rw & Cl$\rightarrow$Ar & Cl$\rightarrow$Pr & Cl$\rightarrow$Rw &
        S$\rightarrow$R & S$\rightarrow$R6 \\
        \hline
        \multirow{5}{*}{MCMD}&               &  $\checkmark$  &                & 81.3 & 92.0 & 97.2 & 79.7 & 87.5 & 91.4 & 59.0 & 72.1 & 72.7 & 62.7 & 68.4 & 68.9 & 63.5 & 62.9 \\
        & $\checkmark$  &  $\checkmark$  &                & 80.1 & 92.1 & 97.2 & 87.0 & 91.2 & 97.0 & 59.9 & 72.3 & 73.1 & 70.5 & 76.2 & 78.7 & 63.9 & 62.0 \\
        & $\checkmark$  &                &  $\checkmark$  & 80.7 & 92.2 & 96.8 & 82.9 & 89.8 & 95.1 & 60.1 & 72.2 & 70.9 &  66.6 & 72.7 & 75.5 & 69.3 & 77.3 \\
        &              &  $\checkmark$  &  $\checkmark$  & \textbf{81.7} & \textbf{93.0} & \textbf{97.4} & 80.1 & 87.5 & 91.7 & 61.4 & 74.6 & 74.1 &  64.1 & 66.6 & 69.3 & 80.9 & 67.2 \\
        & $\checkmark$  &  $\checkmark$  &  $\checkmark$  & 81.4 & 92.9 & \textbf{97.4} & \textbf{87.5} & \textbf{92.2} & \textbf{98.1} & \textbf{62.5} & \textbf{75.4} & \textbf{75.3} & \textbf{70.9} & \textbf{80.4} & \textbf{82.1} & \textbf{82.8} & \textbf{77.5} \\
        \hline
        MMD & $\checkmark$  &  $\checkmark$  &  $\checkmark$ & 80.6 & 90.9 & 96.6 & 80.1 & 87.4 & 90.4 & 56.4 & 67.2 & 68.1 & 70.7 & 78.6 & 79.8 & 65.9 & 67.6 \\
        \bottomrule[1pt]
   \end{tabular}
\vskip -0.10in
\end{table*}

\begin{figure*}[t]
   \begin{minipage}{0.245\linewidth}
       \centering{\includegraphics[width=0.99\linewidth,height=87pt,trim=70 50 50 40,clip]{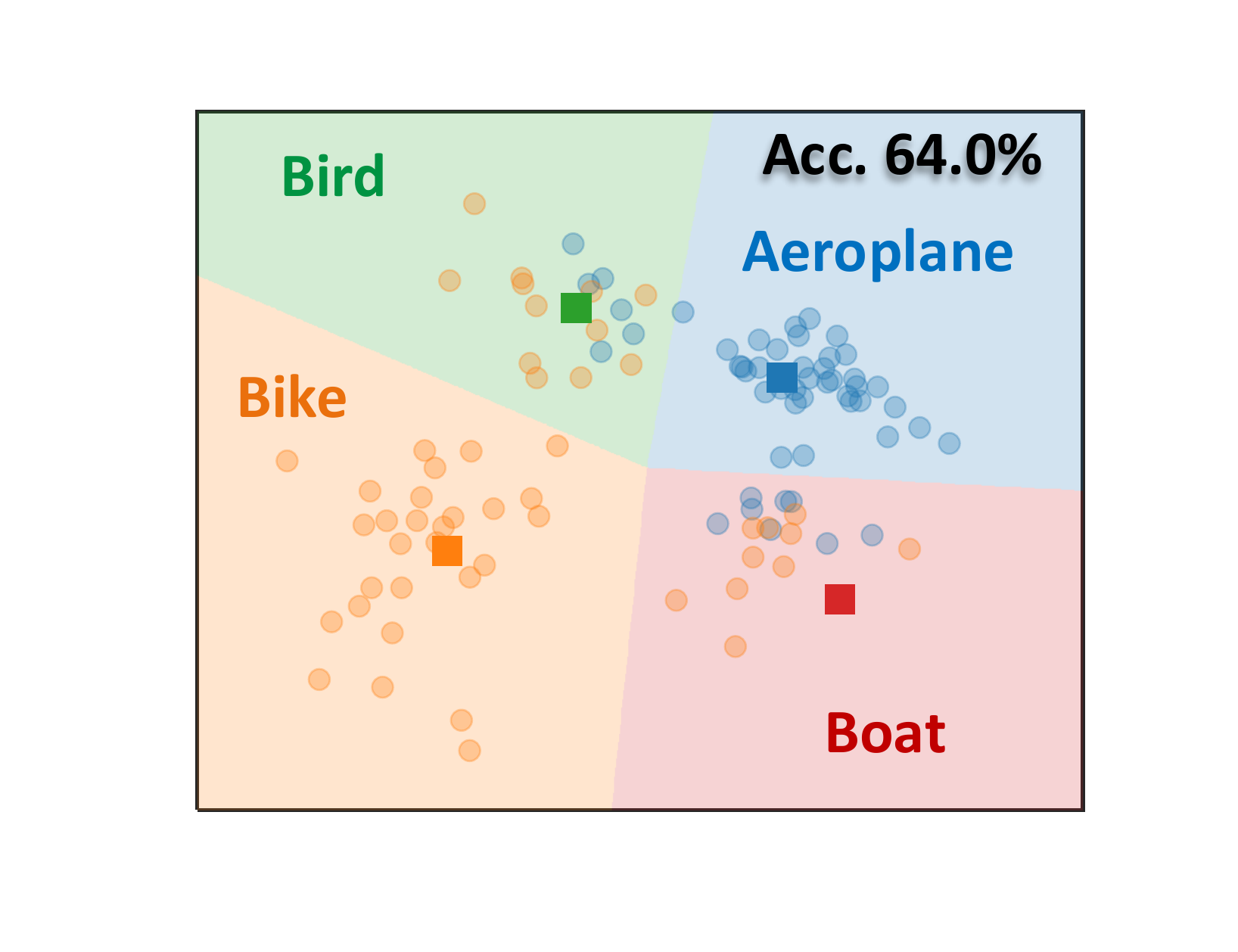}} \\
       (a) MUL (w/o IW)
   \end{minipage}
   \hfill
   \begin{minipage}{0.245\linewidth}
       \centering{\includegraphics[width=0.99\linewidth,height=87pt,trim=70 50 50 40,clip]{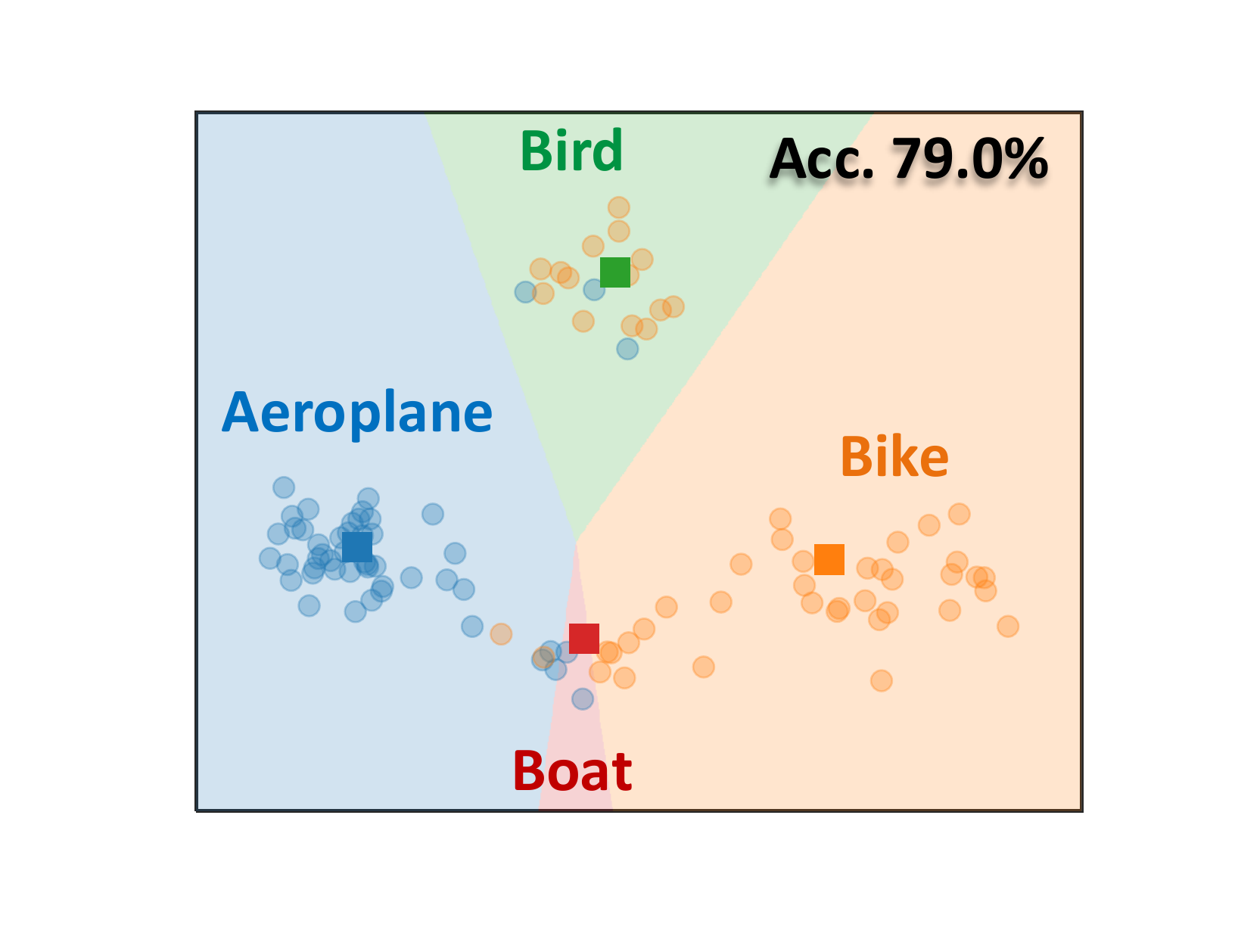}} \\
       (b) MUL (w/ IW)
   \end{minipage}
   \hfill
   \begin{minipage}{0.245\linewidth}
       \centering{\includegraphics[width=0.99\linewidth,height=87pt,trim=70 50 50 40,clip]{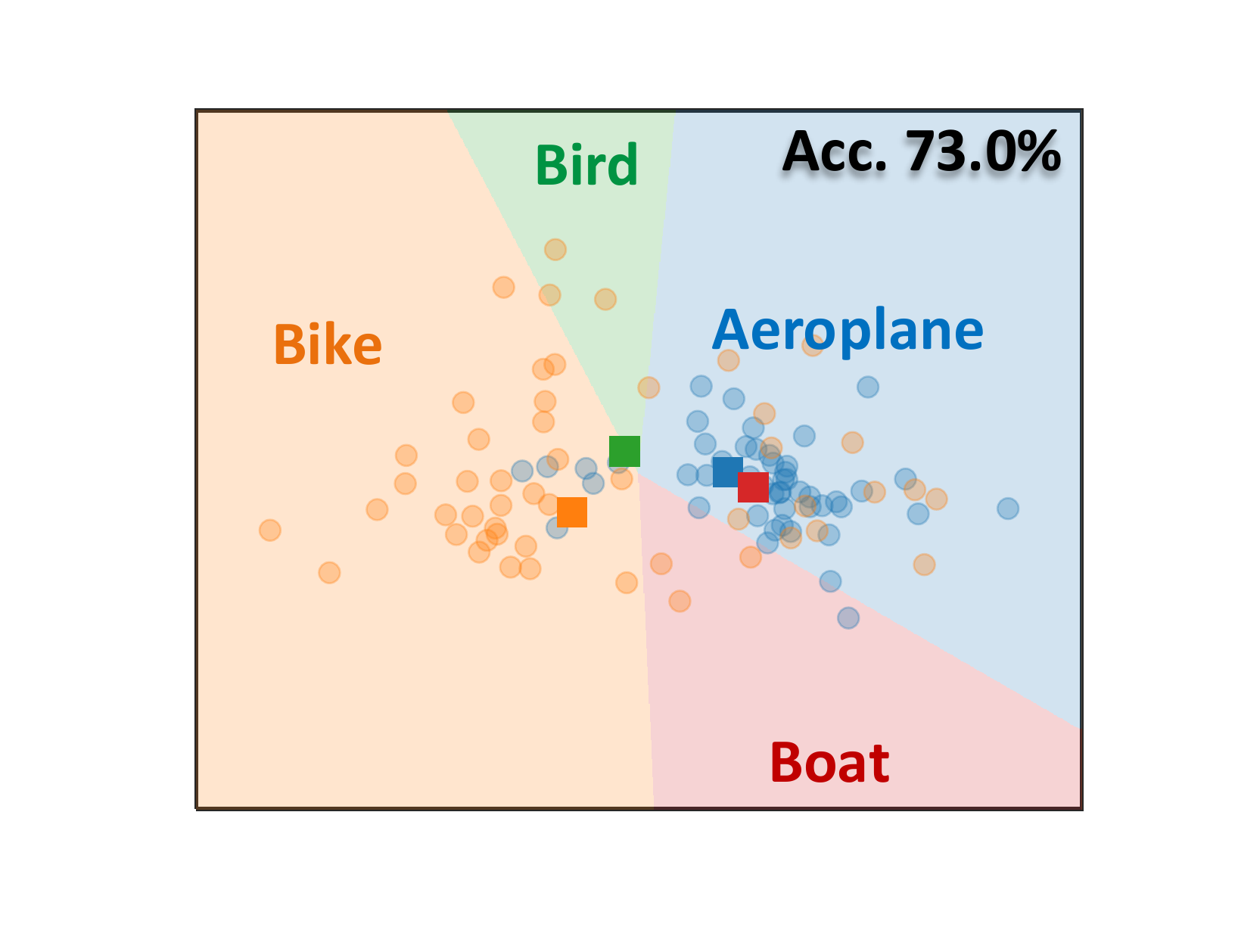}} \\
       (c) MUL (w/o IW)
   \end{minipage}
   \hfill
   \begin{minipage}{0.245\linewidth}
       \centering{\includegraphics[width=0.99\linewidth,height=87pt,trim=70 50 50 40,clip]{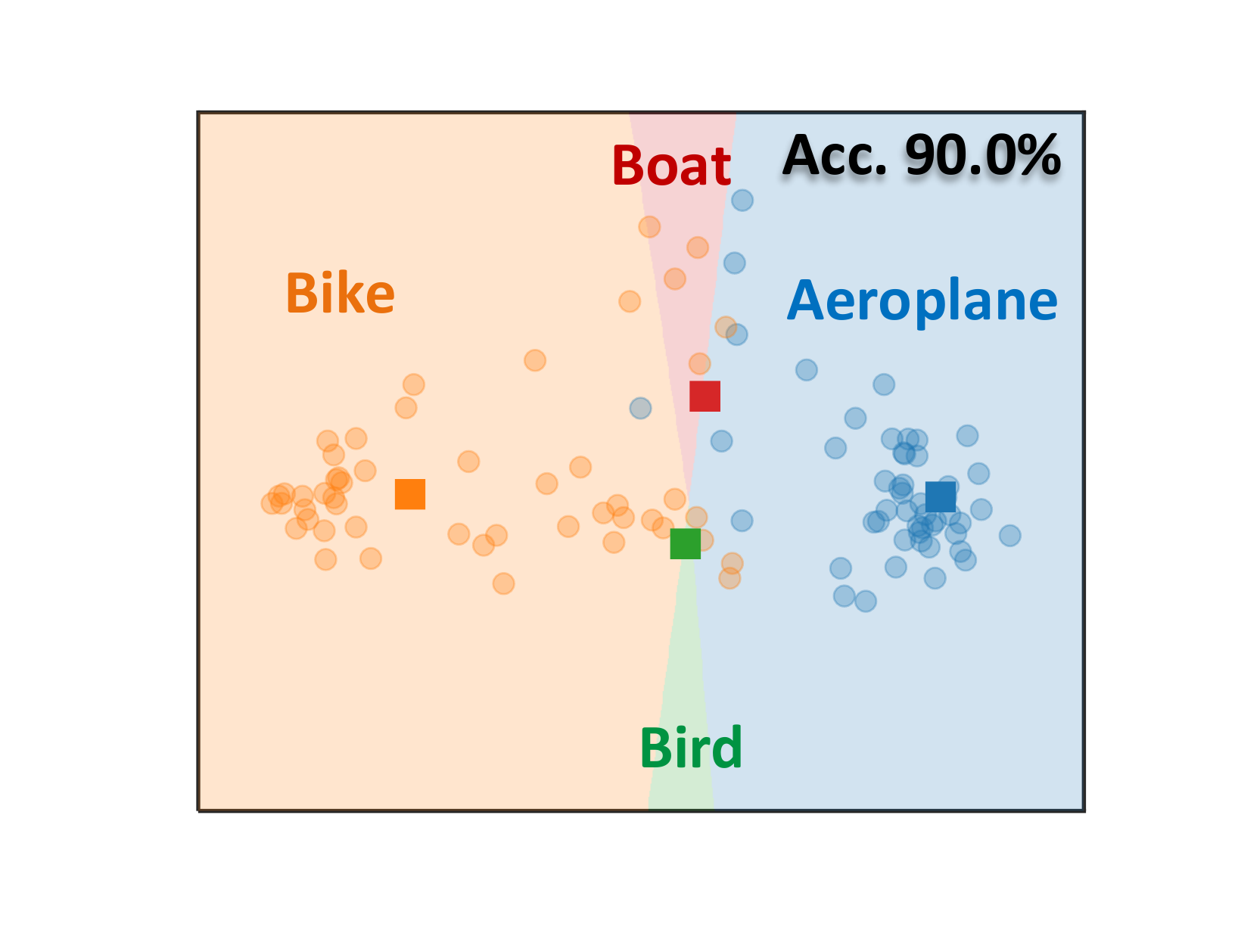}} \\
       (d) MUL (w/ IW)
   \end{minipage}

   \caption{Ablation on the importance weighting strategy via decision boundary visualization on subsampled ImageCLEF dataset, where `$\blacksquare$' represents the prototype of source domain and `$\bullet$' the samples of target domain. The results on \textbf{P}$\rightarrow$\textbf{I} and \textbf{C}$\rightarrow$\textbf{P} tasks are shown in (a)-(b) and (c)-(d), respectively. Best viewed in color.}
   \label{fig:DecisionBoundary_Fig}
   \vskip -0.15in
\end{figure*}

\subsection{Partial Domain Adaptation}
In this section, we conduct experiments under the partial domain adaptation setting. Compared with the vanilla setting, GLS is more severe under partial setting.

\textbf{Comparison. }
MUL is compared with several advanced PDA methods as shown in Table \ref{tab:4Datasets_partial}. On Office-Home dataset, MUL achieves the highest accuracies in 10 out of 12 adaptation tasks and improves the mean accuracy to 77.4\%. Specifically, the accuracy improvements are about 5\% on 6 tasks. On VisDA-2017 dataset, MUL surpasses other methods notably and improves the mean accuracy to 77.5\%. Previous PDA models (\eg, ETN and DRCN) usually focus on the shift of label distributions, which almost correct the biased estimation of empirical risk and achieve higher accuracy than covariate adaptation model (\eg, DANN). MUL attempts to match the conditional distributions across domains to correct the conditional shift, which achieves higher accuracies than the previous PDA methods. Compared with DMP and other conditional shift models, MUL explores the prior weighted matching with minimum transfer uncertainty which guarantees the lower target error.

As shown in the bottom of Table \ref{tab:4Datasets_partial}, the adaptation problem is much easier on the Office-31 and ImageCLEF datasets. MUL improves the accuracies in tasks \textbf{A}$\rightarrow$\textbf{D} and \textbf{W}$\rightarrow$\textbf{A} to 98.5\% and 96.3\% on Office-31 dataset, respectively. On ImageCLEF dataset, the accuracies improved by MUL are more significant. Specifically, MUL improves the accuracies in tasks \textbf{I}$\rightarrow$\textbf{P} and \textbf{C}$\rightarrow$\textbf{P} to 87.5\% and 87.6\%. Those two tasks are difficult since the images on \textbf{P} usually contain large background and clutter. The improvements on hard tasks validate that MUL can transfer the task-related knowledge and ignore the noise induced by the domains effectively.

\textbf{Partial Transfer. }
To evaluate the transferability and discriminability of different models under partial setting, we compute the pairwise distances between the class-wise prototypes (means) across domains as $\left(\MBF{D}^{st} \right)_{ij} = \| \bar{\MBF{z}}^s_i - \bar{\MBF{z}}^t_j \|_2^2$. For a fair comparison, the $\ell_2$-norm of representations are scaled to be the same for different models. The results are shown in Figure \ref{fig:Quantitative_Fig_PDA} (a)-(c), where the distance values are provided. There are two key observations: 1) MUL and DMP can learn transferrable representations $Z$ with small diagonal distances in $\MBF{D}^{st}$; 2) MUL significantly reduces the decision uncertainty by learning the larger inter-class distances for the target (shared) classes. These results demonstrate that MUL can partially transfer the discriminant information via the prior weighted matching.

\textbf{Feature Visualization. }
We also visualize the representations of both domains under partial setting. As the results shown in Figure \ref{fig:Quantitative_Fig_PDA} (e)-(g), ResNet cannot separate the 6 shared classes on the target domains and misaligns most samples. DMP can match some shared classes while the intra-class scatters are still large. MUL correctly matches all shared classes and achieves the smaller intra-class scatter values than DMP on the target domain. Specifically, the mean scatter value of DMP is about 59.2 while MUL is only 23.9. Note Figure \ref{fig:Quantitative_Fig_PDA} (c) also implies that the inter-class scatter values of MUL are also larger than others. Thus, the results demonstrate that MUL can enhance the discriminability with lower decision uncertainty under partial setting.

\textbf{Label Shift Estimation. }
To evaluate the prior probabilities estimated by different methods, we compare the estimation with ``Oracle''. In Figure \ref{fig:Quantitative_Fig_PDA} (d), we observe that the estimation errors of MUL are nearly zeros for the outlier classes, which means MUL successfully detects the label shift. Note that the estimated probabilities of the outlier classes are about $10^{-9}$ which are invisible on the axis. In Figure \ref{fig:Quantitative_Fig_PDA} (h), we also apply BBSE to other methods. We observe that the estimated probabilities of ResNet and DMP are biased, where the weights and errors of outlier classes are large (about $ 10^{-2}$). The estimated probabilities of MUL are more similar to a uniform distribution (``Oracle'') on the shared classes. These results demonstrate that the conditional invariant representations learned by MUL can be better applied to detect the label shift. Also, an accurate estimator of prior probability will encourage model to transfer the shared knowledge while ignoring the outlier classes.

\begin{table*}[t]
   \centering
   \setlength{\abovecaptionskip}{0.0cm}
   \setlength{\belowcaptionskip}{-0.01cm}
   \caption{Classification accuracies (\%) and $\ell_1$ distances between prior probabilities on subsampled Office-Home (ResNet-50).}
   \label{tab:GLS_subsample_Home}

   \renewcommand{\tabcolsep}{0.36pc} 
   \renewcommand{\arraystretch}{1.0} 
   \begin{tabular}{c|cccccccccccc|c}
   \toprule[1pt]
   \textbf{Office-Home} & Ar$\to$Cl & Ar$\to$Pr & Ar$\to$Rw & Cl$\to$Ar & Cl$\to$Pr & Cl$\to$Rw &
   Pr$\to$Ar & Pr$\to$Cl & Pr$\to$Rw & Rw$\to$Ar & Rw$\to$Cl & Rw$\to$Pr & Avg. \\
   $\|p_Y^s-p_Y^t\|_1$ & 0.53 & 0.60 & 0.52 & 0.80 & 0.63 & 0.66 & 0.76 & 0.57 & 0.60 & 0.69 & 0.52 & 0.56 & 0.62  \\
   \hline
   Source \cite{he2016deep} & 35.7 & 54.7 & 62.6 & 43.7 & 52.5 & 56.6 & 44.3 & 33.1 & 65.2 & 57.1 & 40.5 & 70.0 & 51.3   \\
   DANN \cite{ganin2015unsupervised} & 36.1 & 54.2 & 61.7 & 44.3 & 52.6 & 56.4 & 44.6 & 37.1 & 65.2 & 56.7 & 43.2 & 69.9 & 51.8   \\
   IWDANN \cite{combes2020domain} & 39.8 & 63.0 & 68.7 & 47.4 & 61.1 & 60.4 & 50.4 & 41.6 & 72.5 & 61.0 & 49.4 & 76.1 & 57.6 \\
   JAN \cite{Long2017Deep} & 34.5 & 56.9 & 64.5 & 46.2 & 56.8 & 59.1 & 50.6 & 37.2 & 70.0 & 58.7 & 40.6 & 72.0 & 53.9   \\
   IWJAN \cite{combes2020domain} & 36.2 & 61.0 & 66.3 & 48.7 & 59.9 & 61.9 & 52.9 & 37.7 & 70.9 & 60.3 & 41.5 & 73.3 & 55.9   \\
   CDAN \cite{long2018conditional} & 38.9 & 56.8 & 64.8 & 48.0 & 60.1 & 61.2 & 49.7 & 41.4 & 70.2 & 62.4 & 47.0 & 74.7 & 56.3 \\
   IWCDAN \cite{combes2020domain} & 43.0 & 65.0 & 71.3 & \textbf{52.9} & 64.7 & 66.5 & 54.9 & 44.8 & 75.9 & \textbf{67.0} & 50.5 & 78.6 & 61.2  \\
   \hline
   MUL & \textbf{50.0} & \textbf{67.4} & \textbf{72.2} & 52.3 & \textbf{70.1} & \textbf{68.0} & \textbf{59.6} & \textbf{49.9} & \textbf{77.1} & 65.2 & \textbf{54.0} & \textbf{81.5} & \textbf{64.0}  \\
   \bottomrule[1pt]
   \end{tabular}
   \vskip -0.1in
\end{table*}

\subsection{Ablation Study}
In this section, we conduct ablation experiments on ImageCLEF, Office-Home and VisDA-2017 datasets to validate the effectiveness of the individual modules in MUL.

\textbf{MUL Model. }
We evaluate the effectiveness of the weighting strategy, transfer uncertainty $\MC{J}_{\text{TU}}$ and decision uncertainty $\MC{J}_{\text{DU}}$ in Table \ref{tab:Ablation}. There are four key observations: 1) the transfer and decision uncertainties can benefit from each other, and further achieve a higher accuracy; 2) the improvements of transfer uncertainty $\MC{J}_{\text{TU}}$ and decision uncertainty $\MC{J}_{\text{DU}}$ are more significant when the knowledge transfer is harder, \eg, Office-Home and VisDA-2017; 3) the decision uncertainty $\MC{J}_{\text{DU}}$ is more important than the transfer uncertainty $\MC{J}_{\text{TU}}$ on large-scale data since it encourages the model to preserve discriminative structure, \eg, VisDA-2017; 4) the importance weight $\MBF{w}$ and prior weight $\MBF{y}$ are effective, especially when the label shift is severe, \eg, PDA scenario. Note that there is no label shift on ImageCLEF dataset under vanilla setting, thus the model without weights (the $4^{\text{th}}$ row) is just the ``Oracle'' (uniform weights). In conclusion, the combination of $\MC{J}_{\text{TU}}$ and $\MC{J}_{\text{DU}}$ generally guarantee the SOTA performance, and their effects are more significant on more challenging datasets. Meanwhile, the weighting strategy can further boost the performance by correcting the label shift.

\textbf{Conditional Metric. }
We compare the proposed MCMD metric with the class-wise MMD by replacing the MCMD in $\MC{J}_{\text{DU}}$ and $\MC{J}_{\text{TU}}$ with MMD metric. Specifically, the dataset will be split according to (pseudo) labels, and MMD will be applied to the split data class-wisely. The results of class-wise MMD are shown in the last row in Table \ref{tab:Ablation}. The accuracies of MCMD are about 3.9\%, 4.3\% and 13.4\% higher than class-wise MMD on ImageCLEF, Office-Home and VisDA-2017 datasets, respectively. Thus, the superiority of MCMD to the class-wise MMD is more apparent as sample size increases. It indicates that MCMD can integrate more useful information in estimation, and achieve better performance in matching/separating the conditional distributions.

\textbf{Effect of Importance Weighting. }
In recent work \cite{byrd2019effect}, the authors show that Importance Weighting (IW) has almost no effect on the binary decision boundary when there is no label shift. This is consistent with our observations in the ablation experiments on the ImageCLEF dataset (UDA scenario). However, apart from this special setting, IW is generally important and effective in the complex setting, \eg, GLS. It can be validated by the rest of datasets and scenarios in the ablation experiments (Table \ref{tab:Ablation}), where the conclusion in \cite{byrd2019effect} is generally incorrect.

We also conduct visualization experiments of decision boundary on ImageCLEF \textbf{P}$\rightarrow$\textbf{I} and \textbf{C}$\rightarrow$\textbf{P} tasks. We subsample the dataset by selecting the first 4 classes (alphabetical order) for visualization. To simulate label shift scenario, we set the first 2 classes as shared classes and the rest 2 classes as outlier classes. The visualizations of decision boundary (background color) and target samples are presented in Figure \ref{fig:DecisionBoundary_Fig}. In (a) and (c), we observe that the region of outlier classes (\ie, Bird and Boat) are large where many samples are misaligned to the outlier classes. In (b) and (d), there has a significant shrinkage in the decision boundaries of outlier classes after applying IW. Besides, less samples are located in the region of outlier classes and the accuracies of MUL (w/ IW) are higher. These observations demonstrate that IW has a significant impact on model when GLS exists.

\vspace{-2pt}
\subsection{GLS with Subsampled Data}
To exacerbate GLS with enlarged label discrepancy and less training samples, we follow the protocol in \cite{combes2020domain} to subsample the Office-Home dataset. The subsampled data contains 30\% of the samples in the first 32 classes (alphabetical order) and all samples in remaining 33 classes. We only subsample the source domain while keeping the target domain unchanged. We reproduce the subsample process with the same random seeds and code released by the authors \cite{combes2020domain}. Compared with vanilla setting, the subsample setting is more challenging since: 1) GLS is more severe and $\ell_1$ distance is about $2\times$ larger; 2) there are less labeled source data for training.

The results are presented in Table \ref{tab:GLS_subsample_Home}. In the presence of GLS, the importance weight strategy in \cite{combes2020domain} can actually boost the performance of previous methods like DANN, JAN and CDAN. However, this strategy relies on the assumption that the backbone models can match the conditional distributions, which is generally unsatisfied by previous models. Since we propose the conditional metric operator to learn the conditional invariant transformation actively, the proposed model outperforms others and improve the mean accuracy to 64.0\%. Compared with literature \cite{combes2020domain}, MUL employs the same weight estimator (BBSE \cite{lipton2018detecting}) but different weighting strategy and learning criteria. Thus, the improvements validate the effectiveness of MUL in learning conditional invariant representations and correcting GLS.

\section{Conclusion}
In this paper, we propose the \textit{minimum uncertainty principle} for correcting GLS, which bridges the gap between GLS and statistical learning theory. The theoretical results introduce a lower generalization error bound than previous covariate adaptation. To mathematically characterize the uncertainty, we propose the \textit{conditional metric operator} in infinite-dimensional Hilbert space based on the conditional embeddings. Following the principle, a novel conditional adaptation framework is proposed to deal with GLS, which is generally applicable for different domain adaptation scenarios, \eg, UDA and PDA. With the \textit{conditional metric operator}, some appealing properties are ensured for the empirical MUL, \eg, identifiability and consistency. Extensive experiments validate that MUL transfers the knowledge correctly and guarantees the discriminability on both domains.

A tighter upper bound without conditional and label discrepancies will be our future work. For methodology, MUL can be extended to other UDA settings such as multi-source, open set and universal settings.

\ULforem
\bibliographystyle{IEEEtran}
\bibliography{MUL_Ref}
%
\vspace{-15pt}

\begin{IEEEbiography}[{\resizebox{1.0in}{1.3in}{\includegraphics*{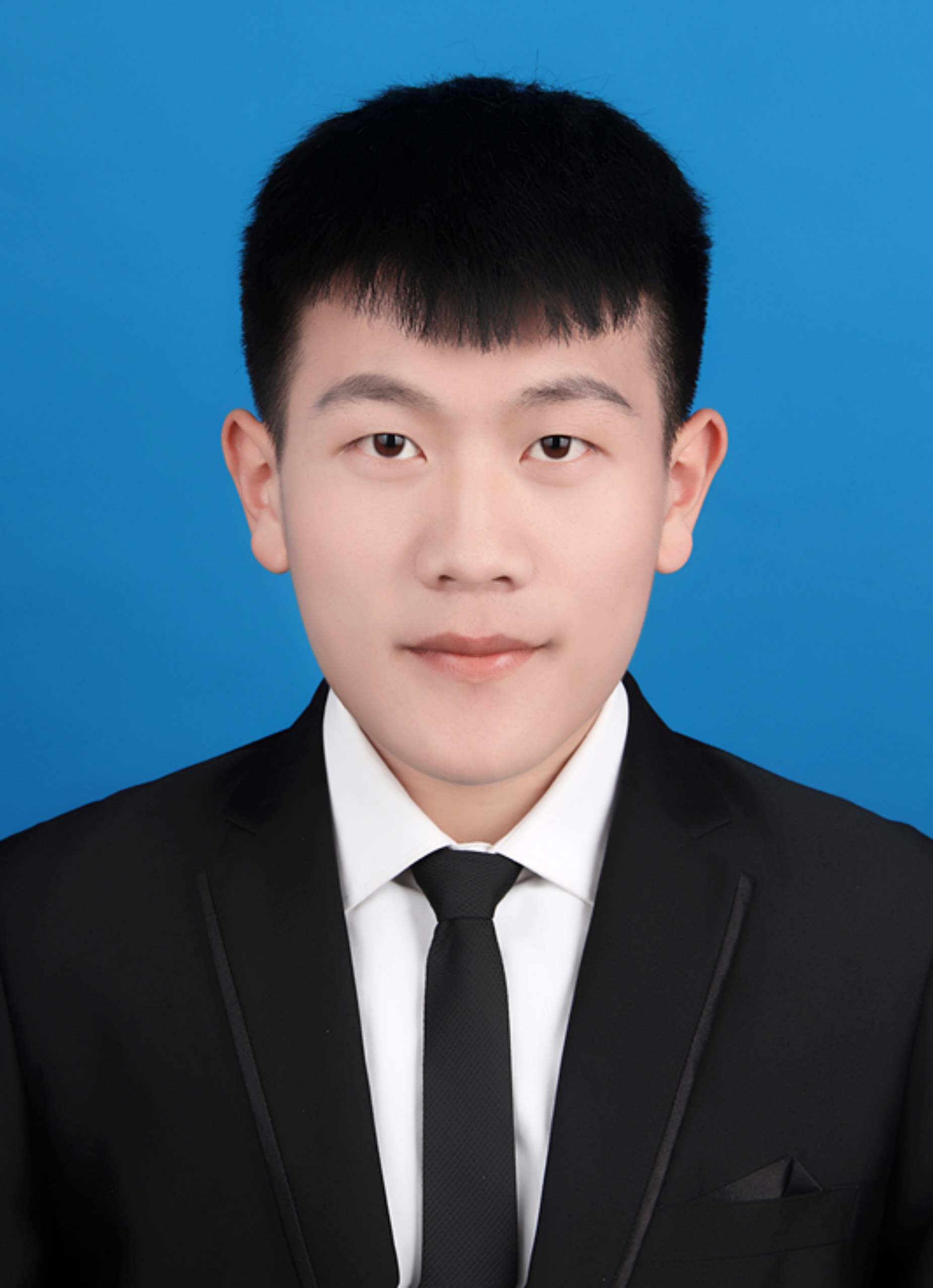}}}]{You-Wei Luo}
received the B.S. degree in statistics from China University of Mining and Technology, Xuzhou, China, in 2018.  He is currently pursuing the Ph.D. degree with  the School of Mathematics, Sun Yat-sen University, Guangzhou, China. His research interests include image processing, manifold learning and machine learning.
\end{IEEEbiography}
\vspace{-15pt}

\begin{IEEEbiography}[{\resizebox{1.0in}{1.3in}{\includegraphics*{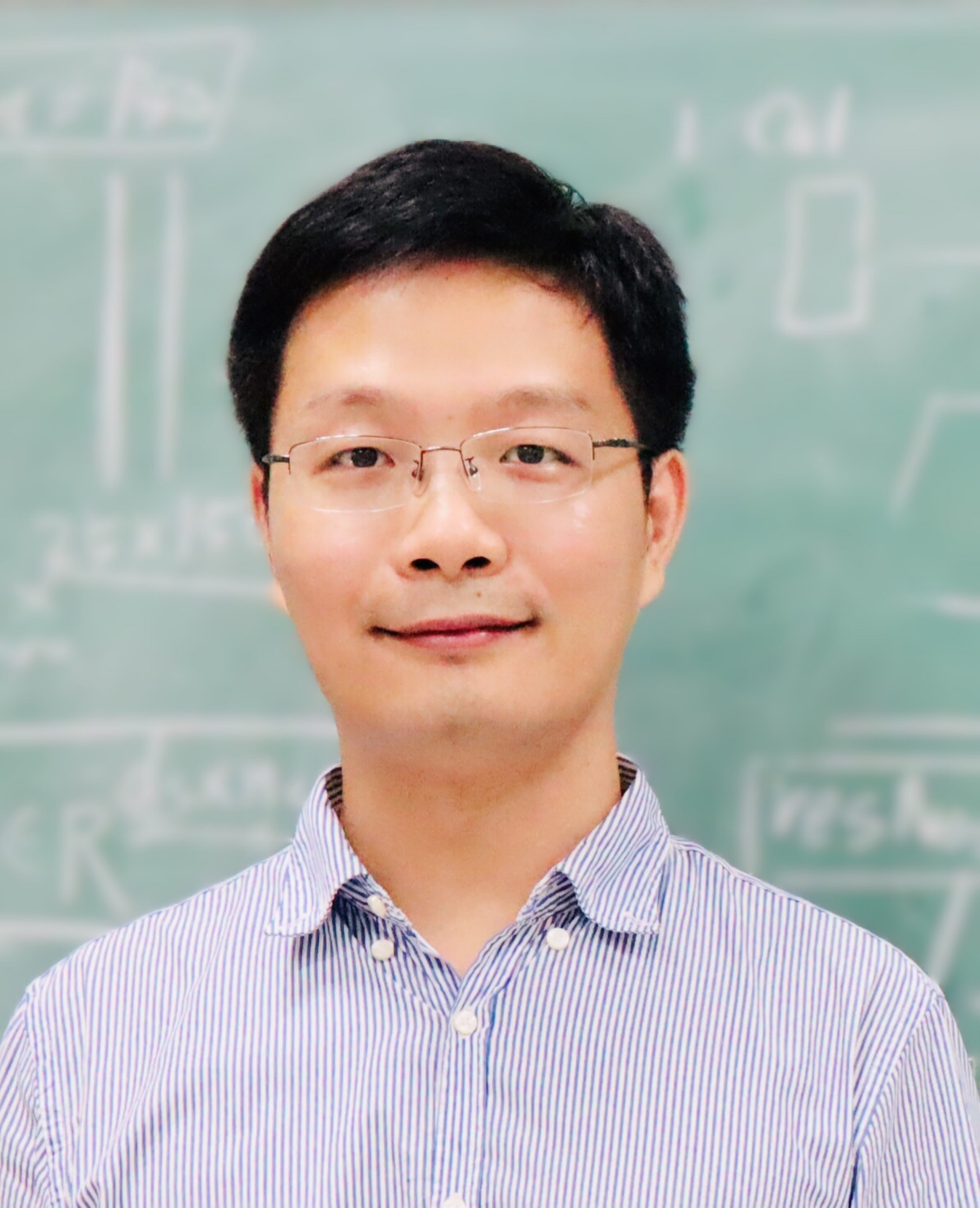}}}]{Chuan-Xian Ren}
received the PhD degree from Sun Yat-Sen University, Guangzhou, China, in 2010. He is currently Associate professor of the School of Mathematics, Sun Yat-Sen University. \par His research interests include image processing, pattern recognition and machine learning.
\end{IEEEbiography}
\vspace{-15pt}






\end{document}